%% file: main.tex
\documentclass[lettersize,journal]{IEEEtran}
\usepackage{amsmath,amsfonts}
\usepackage{algorithmic}
\usepackage{algorithm}
\usepackage{array}
\usepackage[caption=false,font=footnotesize,labelfont=rm,textfont=rm]{subfig}
\usepackage{textcomp}
\usepackage{stfloats}
\usepackage{orcidlink}
\usepackage{url}
\usepackage{verbatim}
\usepackage{graphicx}
\usepackage{cite}
\usepackage{bbding}
\usepackage{threeparttable}
\usepackage[T1]{fontenc}
\usepackage{cleveref}
\usepackage{multirow}
\usepackage{booktabs}
\usepackage{colortbl}
\usepackage{xcolor}
\usepackage{bm}
\usepackage{tikz}
\hypersetup{hidelinks}

\usetikzlibrary{arrows.meta,bending,intersections,quotes,shapes.geometric,shapes, arrows, positioning,trees,backgrounds,shapes.misc}

\tikzstyle{tikzfig}=[baseline=-0.25em,scale=0.5]

\pgfkeys{/tikz/tikzit fill/.initial=0}
\pgfkeys{/tikz/tikzit draw/.initial=0}
\pgfkeys{/tikz/tikzit shape/.initial=0}
\pgfkeys{/tikz/tikzit category/.initial=0}

\pgfdeclarelayer{edgelayer}
\pgfdeclarelayer{nodelayer}
\pgfsetlayers{background,edgelayer,nodelayer,main}

\tikzstyle{none}=[inner sep=0mm]

\newcommand{\tikzfig}[1]{
{\tikzstyle{every picture}=[tikzfig]
\IfFileExists{#1.tikz}
  {\input{#1.tikz}}
  {
    \IfFileExists{./figures/#1.tikz}
      {\input{./figures/#1.tikz}}
      {\tikz[baseline=-0.5em]{\node[draw=red,font=\color{red},fill=red!10!white] {\textit{#1}};}}
  }}
}

\tikzstyle{every loop}=[]

\graphicspath{{figures}}
\DeclareMathOperator*{\argmax}{\arg\!\max}

\def\eg{e.g.,~}

\def\etal{{\em et al.}}
\definecolor{policy}{RGB}{170, 193, 154}
\definecolor{experience}{RGB}{151, 170, 203}
\definecolor{dynamic}{RGB}{253, 194, 128}
\definecolor{reward}{RGB}{255, 186, 199}
\newcolumntype{L}[1]{>{\raggedright\let\newline\\\arraybackslash\hspace{0pt}}m{#1}}
\newcolumntype{C}[1]{>{\centering\let\newline\\\arraybackslash\hspace{0pt}}m{#1}}
\newcolumntype{R}[1]{>{\raggedleft\let\newline\\\arraybackslash\hspace{0pt}}m{#1}}

\makeatletter
\newcommand{\crefnames}[3]{
  \@for\next:=#1\do{
    \expandafter\crefname\expandafter{\next}{#2}{#3}
  }
}
\makeatother

\crefnames{part,chapter,section}{\S}{\S\S}
\crefnames{paragraph,subparagraph}{\P}{\P\P}

\begin{document}

\title{A Survey of Continual Reinforcement Learning}

\author{Chaofan Pan$^{\orcidlink{0000-0001-5345-2746}}$, \IEEEauthorblockN{Xin Yang\thanks{* Xin Yang is corresponding author.}}$^{\orcidlink{0000-0002-0406-6774}*}$, \textit{Member}, \textit{IEEE}, Yanhua Li$^{\orcidlink{0009-0008-7933-1289}}$, Wei Wei$^{\orcidlink{0000-0003-3963-2884}}$, \textit{Member}, \textit{IEEE}, Tianrui Li$^{\orcidlink{0000-0001-7780-104X}}$, \textit{Senior Member}, \textit{IEEE}, Bo An$^{\orcidlink{0000-0002-7064-7438}}$, \textit{Senior Member}, \textit{IEEE}, Jiye Liang$^{\orcidlink{0000-0001-5887-9327}}$, \textit{Fellow}, \textit{IEEE}

\thanks{Chaofan Pan, Xin Yang, and Yanhua Li are with the School of Computing and Artificial Intelligence, Southwestern University of Finance and Economics, Chengdu, 611130, China. E-mail: pan.chaofan@foxmail.com, yangxin@swufe.edu.cn, yyhhliyanhua@163.com.}
\thanks{Wei Wei and Jiye Liang are with the Key Laboratory of Computational Intelligence and Chinese Information Processing of Ministry of Education, School of Computer and Information Technology, Shanxi University, Taiyuan, Shanxi, 030006, China. E-mail: \{weiwei, ljy\}@sxu.edu.cn.}
\thanks{Tianrui Li is with the School of Computing and Artificial Intelligence, Southwest Jiaotong University, Chengdu, 611756, China. E-mail: trli@swjtu.edu.cn.}
\thanks{Bo An is with the College of Data Science and Computing, Nanyang Technological University, 639798, Singapore. E-mail: boan@ntu.edu.sg.}

}

\markboth{IEEE Transactions on Pattern Analysis and Machine Intelligence,~Vol.~xx, No.~xx, xx~2025}
{Shell \MakeLowercase{\textit{et al.}}: A Sample Article Using IEEEtran.cls for IEEE Journals}

\maketitle

\begin{abstract}
  Reinforcement Learning (RL) is an important machine learning paradigm for solving sequential decision-making problems.
  Recent years have witnessed remarkable progress in this field due to the rapid development of deep neural networks.

  However, the success of RL currently relies on extensive training data and computational resources. 
  In addition, RL’s limited ability to generalize across tasks restricts its applicability in dynamic and real-world environments.
  With the arisen of Continual Learning (CL), Continual Reinforcement Learning (CRL) has emerged as a promising research direction to address these limitations by enabling agents to learn continuously, adapt to new tasks, and retain previously acquired knowledge.

  In this survey, we provide a comprehensive examination of CRL, focusing on its core concepts, challenges, and methodologies.
  Firstly, we conduct a detailed review of existing works, organizing and analyzing their metrics, tasks, benchmarks, and scenario settings.
  Secondly, we propose a new taxonomy of CRL methods, categorizing them into four types from the perspective of knowledge storage and/or transfer.
  Finally, our analysis highlights the unique challenges of CRL and provides practical insights into future directions.

\end{abstract}


\section{INTRODUCTION}\label{sec:introduction}
\input{sections/introduction.tex}

\section{BACKGROUND}\label{sec:background}
\input{sections/background.tex}

\section{OVERVIEW}\label{sec:overview}
\input{sections/overview.tex}

\section{METHODS REVIEW}\label{sec:review}
\input{sections/review.tex}

\section{FUTURE WORKS}\label{sec:future}
\input{sections/future.tex}

\section{CONCLUSION}\label{sec:conclusion}
\input{sections/conclusion.tex}

\bibliographystyle{IEEEtran}
\bibliography{IEEEabrv, main_reb}

\section{Biography Section}

\vspace{-25pt}
\begin{IEEEbiography}[{\includegraphics[width=1in,height=1.25in,clip,keepaspectratio]{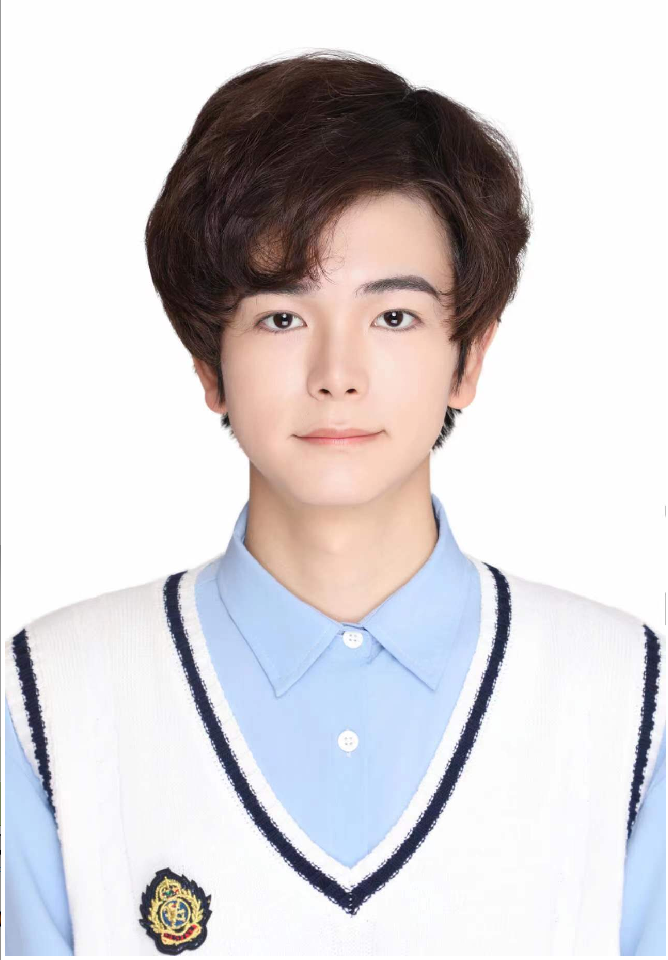}}]{Chaofan Pan}
received the B.S. and M.S. degrees from Southwest Petroleum University, Chengdu, China, in 2020 and 2023, respectively. Now he is pursuing a Ph.D. degree from the School of Computing and Artificial Intelligence, Southwestern University of Finance and Economics. His main research interests include reinforcement learning, self-supervised learning, and continual reinforcement learning.
\end{IEEEbiography}
\vspace{-33pt}

\begin{IEEEbiography}[{\includegraphics[width=1in,height=1.25in,clip,keepaspectratio]{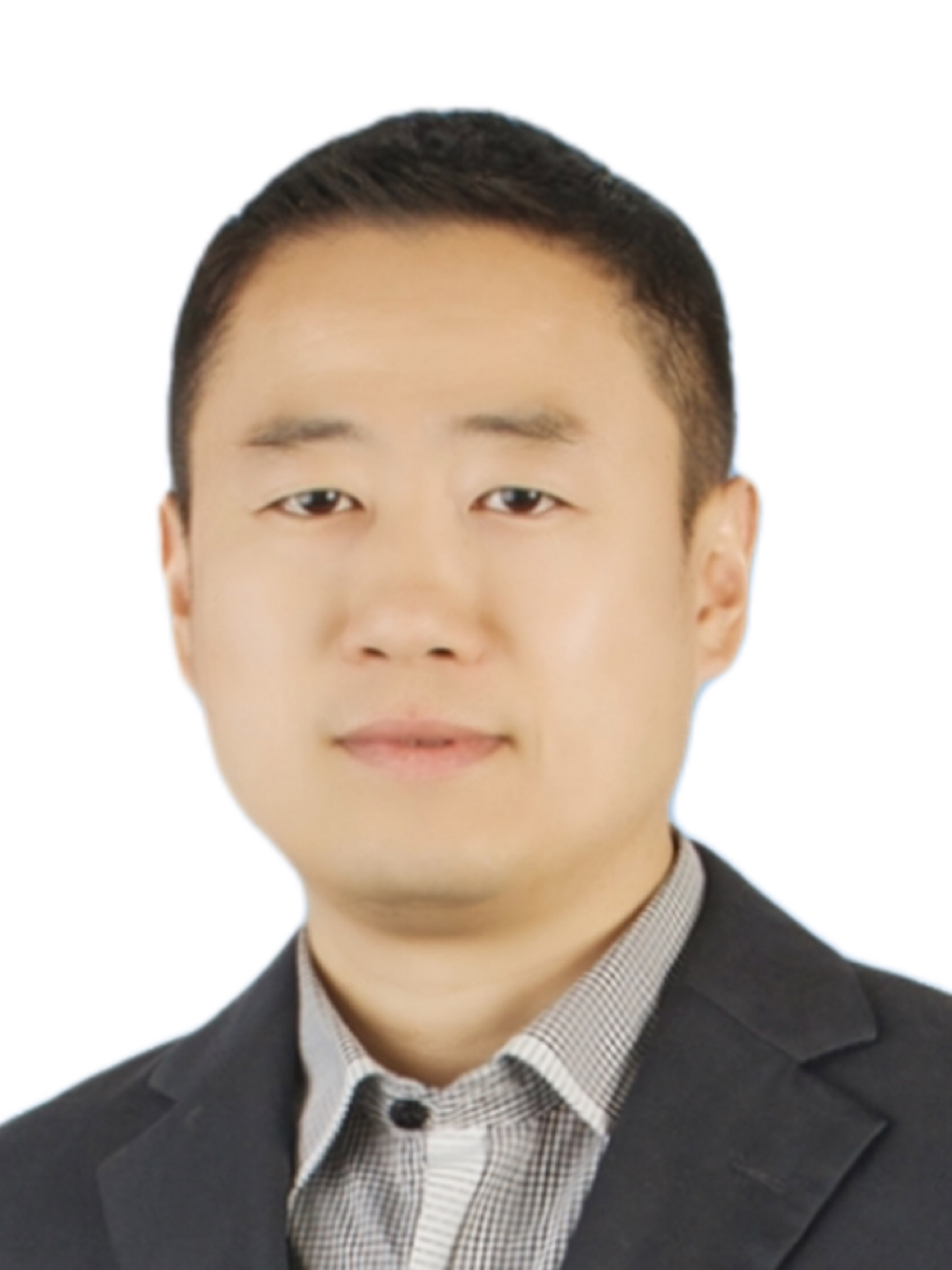}}]{Xin Yang} (Member, IEEE) received the Ph.D. degree in computer science from Southwest Jiaotong University, Chengdu, in 2019. He is currently a Professor at the School of Computing and Artificial Intelligence, Southwestern University of Finance and Economics. He has authored more than 100 research papers in refereed journals and conferences. His research interests include federated learning, continual learning, and multi-granularity learning.
\end{IEEEbiography}
\vspace{-33pt}

\begin{IEEEbiography}[{\includegraphics[width=1in,height=1.25in, clip,keepaspectratio]{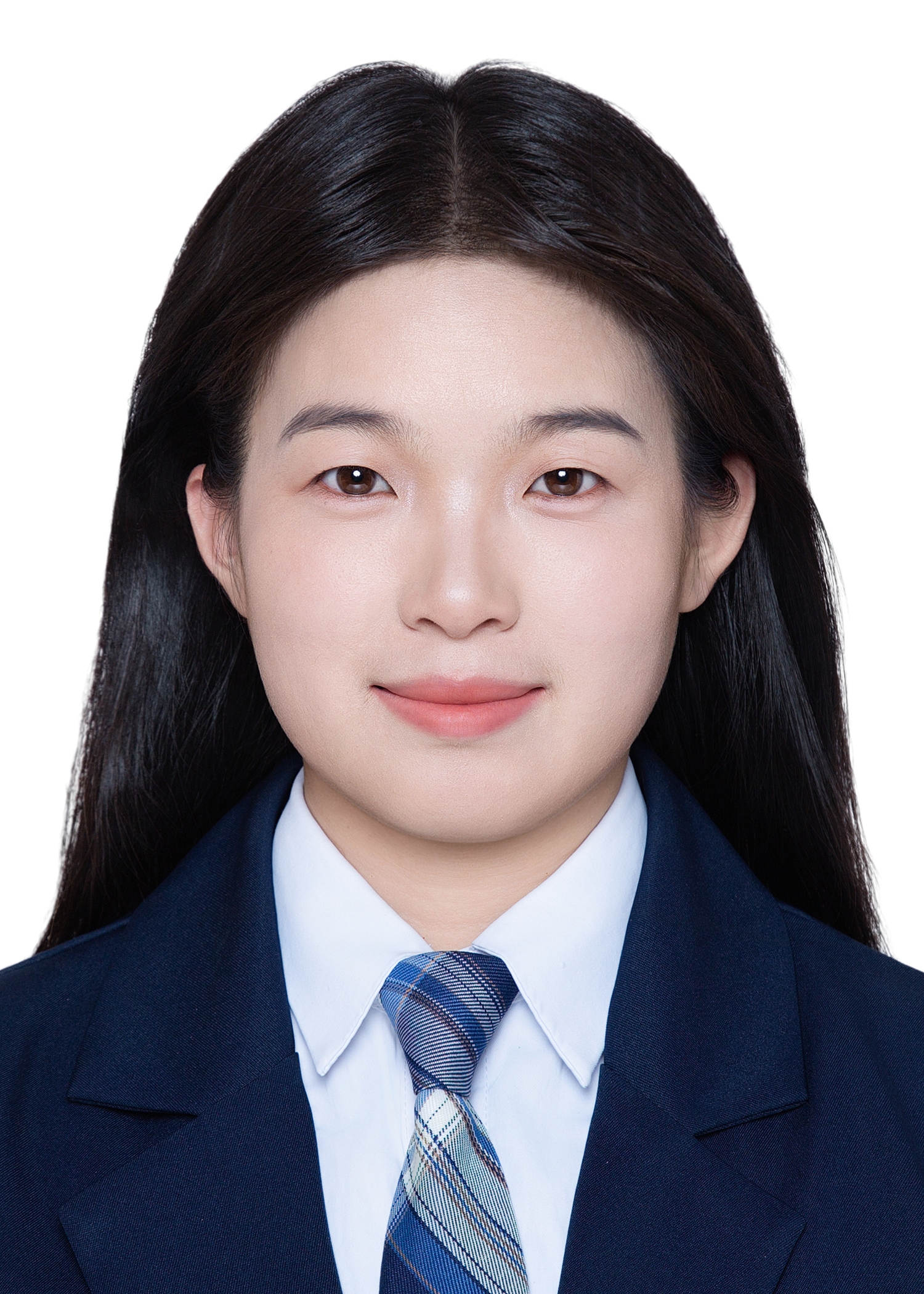}}]{Yanhua Li} received the M.S. degree from Southwest Petroleum University, Chengdu, China, in 2022. She is currently working toward a Ph.D. degree at the School of Computing and Artificial Intelligence, Southwestern University of Finance and Economics, Chengdu, China. Her research interests include continual learning, multi-granularity learning, and open intent classification. She has published several papers in conferences and journals, such as AAAI and Information Fusion.
\end{IEEEbiography} 
\vspace{-33pt}

\begin{IEEEbiography}[{\includegraphics[width=1in,height=1.25in, clip,keepaspectratio]{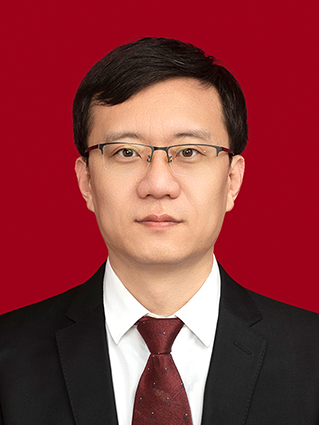}}]{Wei Wei} (Member, IEEE) received the Ph.D. degree in computer science from Shanxi University in 2012. He is currently a professor with the School of Computer and Information Technology, Shanxi University. He has authored or coauthored more than 20 journal papers in his research fields. His research interests include reinforcement learning and granular computing.
\end{IEEEbiography} 
\vspace{-33pt}

\begin{IEEEbiography}[{\includegraphics[width=1in,height=1.25in,clip,keepaspectratio]{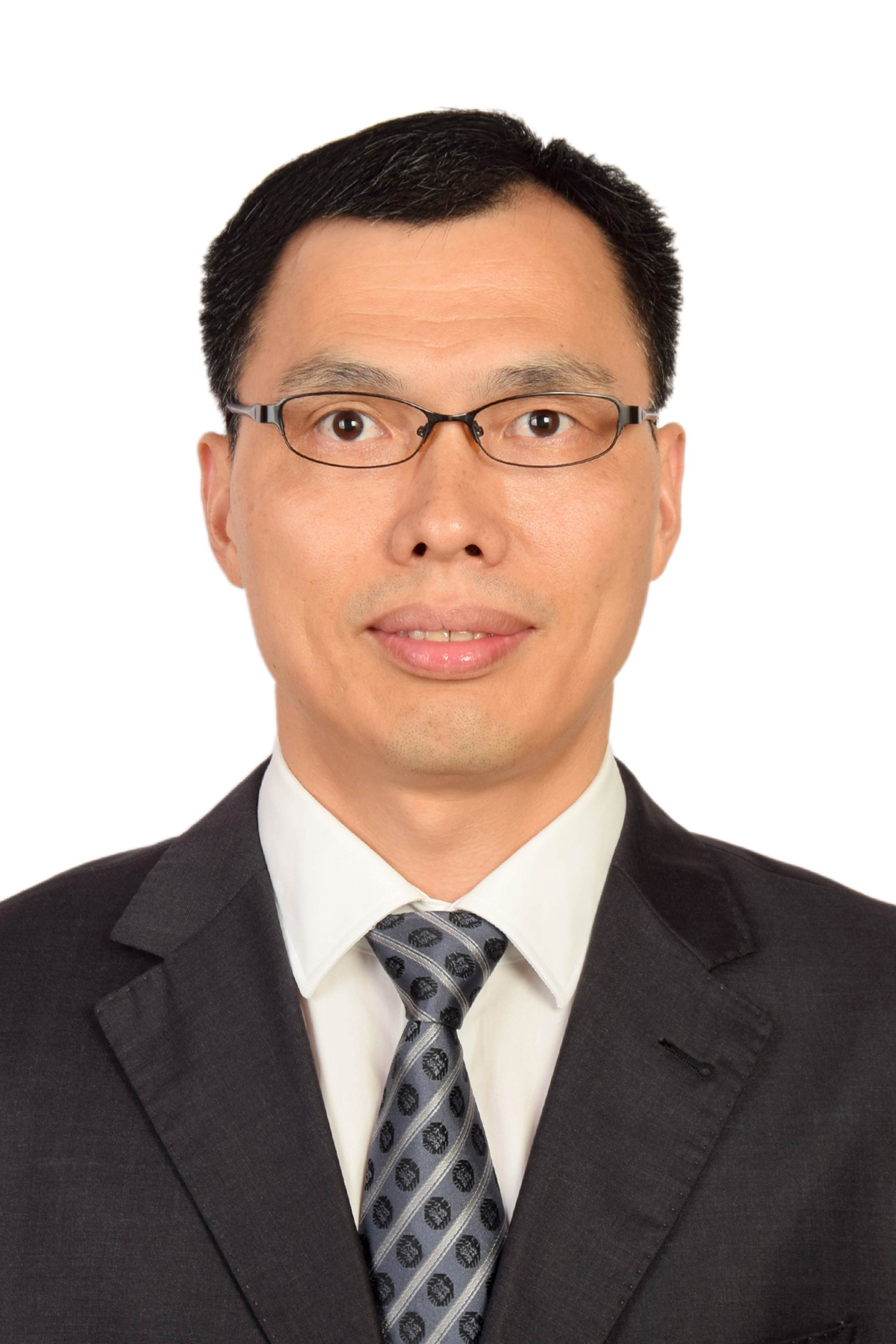}}]{Tiranrui Li} (Senior Member, IEEE) received the Ph.D. degree from Southwest Jiaotong University, Chengdu, China, in 2002. 

He is currently a professor of the Key Laboratory of Cloud Computing and Intelligent Techniques, Southwest Jiaotong University. 
He has published more than 300 research papers in refereed journals and conferences. 
His research interests include big data, machine learning, data mining, granular computing, and rough sets.
\end{IEEEbiography}  
\vspace{-33pt}

\begin{IEEEbiography}[{\includegraphics[width=1in,height=1.25in,clip,keepaspectratio]{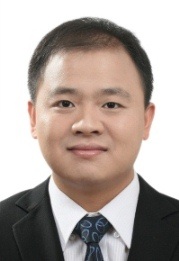}}]{Bo An} (Senior Member, IEEE) received his Ph.D. degree in Computer Science from the University of Massachusetts, Amherst, MA, USA, in 2010. He is a President's Council Chair professor at Nanyang Technological University. His research interests include artificial intelligence, multi-agent systems, reinforcement learning, game theory, and optimization.
\end{IEEEbiography} 
\vspace{-33pt}

\begin{IEEEbiography}[{\includegraphics[width=1in,height=1.25in,clip,keepaspectratio]{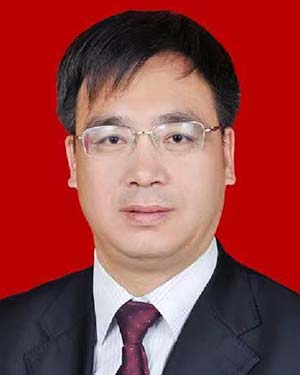}}]{Jiye Liang} (Fellow, IEEE) received the Ph.D. degree from Xi'an Jiaotong University, Xi'an, China, in 2001. 
He is a professor with the School of Computer and Information Technology, Shanxi University, where he is the director of the Key Laboratory of Computational Intelligence and Chinese Information Processing of Ministry of Education. 
He has published more than 200 papers, including the IEEE TPAMI, ICML, AAAI, etc.. 
His research interests include data mining and artificial intelligence.
\end{IEEEbiography}

\input{sections/appendix.tex}
\end{document}

%% file: sections/introduction.tex
\IEEEPARstart{C}{ontinual} Reinforcement Learning (CRL, a.k.a. \textit{Lifelong Reinforcement Learning}, LRL) studies how an agent maintains and extends decision-making competence as it encounters a stream of related, non-stationary tasks without restarting from scratch \cite{ring1997child,khetarpal2022towards}. Real deployments across robotics, autonomy, and agentic software cannot afford to repeatedly reset and retrain. 
Instead, they demand a persistent agent that balances three coupled objectives: acquiring new behavior (plasticity), retaining previously learned skills (stability), and staying within compute/memory budgets as the task stream grows (scalability).

Modern \textit{Deep Reinforcement Learning} (Deep RL) has achieved striking performance in single-task settings, yet practical deployments remain constrained by sample demands, brittle generalization, and the tendency to re-optimize from scratch when the environment or goal specification changes \cite{ding2020deep,dylac2021challenges}. 
In contrast, humans routinely reuse and refine prior experience while avoiding catastrophic forgetting \cite{kudithipudi2022biological}. 
This gap motivates continual learning (CL): learning systems that adapt to new tasks while preserving prior knowledge, navigating the well-known stability--plasticity dilemma \cite{german2019continual,de2020acontinual,wang2024comprehensive}.

\begin{figure}
    \centerline{\includegraphics[width=0.9\linewidth]{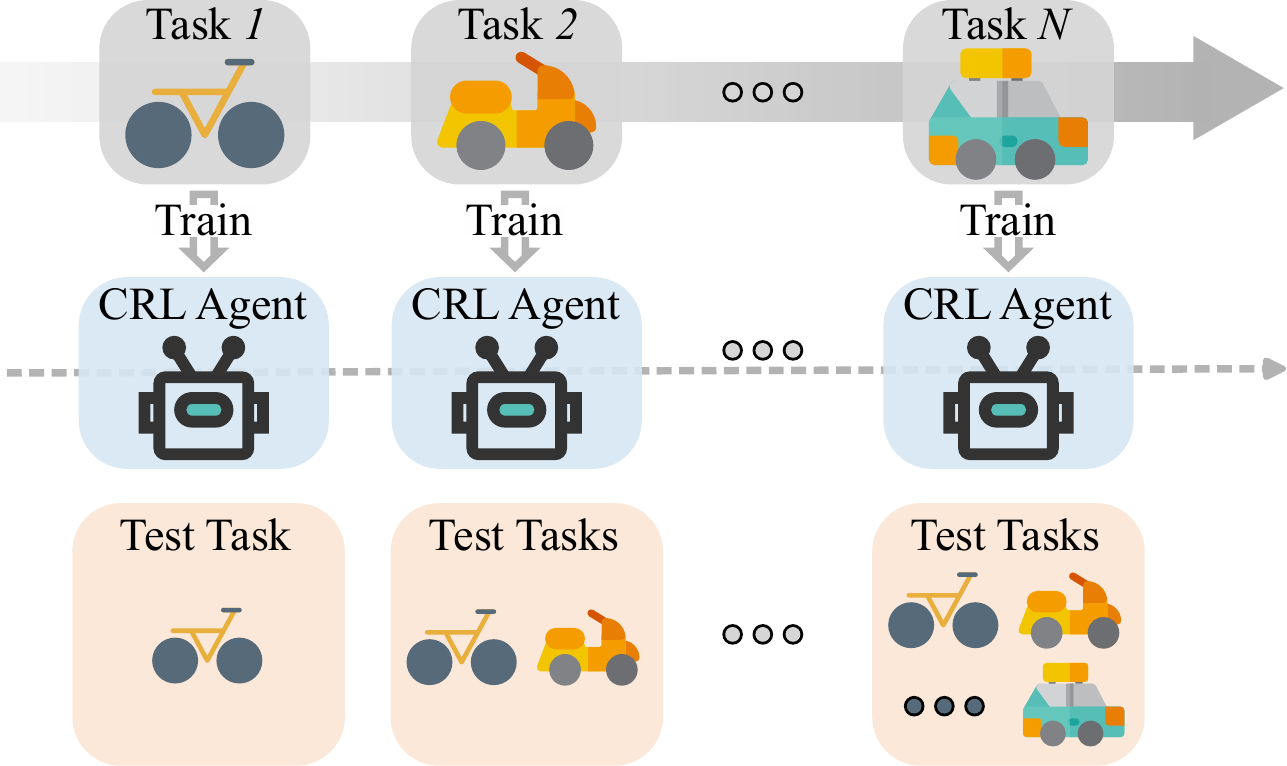}}
    \caption{
        The setting of CRL. Different types of tasks arrive sequentially, and the agent is required to learn to solve the new task incrementally. After learning each task, the agent is evaluated on all previously learned tasks.
        }
    \label{fig: problem}
    \vspace{-0.5cm}
\end{figure}

CRL sits at the intersection of RL and CL \cite{ring1997child,khetarpal2022towards}. 
Beyond the supervised CL template, the sequential decision-making context introduces additional failure modes and resource pressures: the agent must act under partial observability and delayed rewards, while its data distribution shifts as both the policy and the environment evolve. 
Fig.~\ref{fig: problem} illustrates the CRL workflow: tasks arrive sequentially, every new lesson risks catastrophic forgetting, and success is judged on performance over the entire task history rather than only the final task.

It is worth noting that the terms ``lifelong'' and ``continual'' are often used interchangeably in the RL literature, but their usage can vary significantly across studies, potentially leading to confusion \cite{muppidi2024fast}.
In general, most LRL research emphasizes rapid adaptation to new tasks, while CRL research prioritizes avoiding catastrophic forgetting. 
In this survey, we unify the two terms under the umbrella of CRL, reflecting the broader trend in CL research to address both aspects simultaneously.
A CRL agent is expected to achieve two key objectives: 1) minimizing the forgetting of knowledge from previously learned tasks and 2) leveraging prior experiences to learn new tasks more efficiently. 
By fulfilling these objectives, CRL holds the promise of addressing the current limitations of DRL, paving the way for RL techniques to be applied in broader and more complex domains.

Currently, a limited number of works have reviewed the field of CRL.
Some surveys \cite{wang2024comprehensive,wang2025comprehensive} provide a comprehensive overview of CL in general, including both supervised learning and RL.
Most notably, Khetarpal \etal \cite{khetarpal2022towards} published a survey on CRL from the perspective of non-stationary RL.
The survey first formulates the definition of the general CRL problem and provides a taxonomy of different CRL formulations by mathematically characterizing two key properties of non-stationarity.
However, it lacks detailed comparison and discussion of some important parts in CRL, such as the scalability and scenario settings, which are essential for guiding practical research.
Furthermore, the number of CRL methods has been growing rapidly in recent years.
Therefore, it is necessary to provide a new review of the most recent research in CRL.

In this survey, we provide a comprehensive examination of CRL, focusing on its foundational concepts, challenges, and methodologies. 
We explore the intricacies of CRL and identify the key challenges and benchmarks that define the field. 
To achieve this, we systematically review the current state of CRL research and propose a taxonomy that organizes existing methods into distinct categories. 
Our analysis also extends to novel research areas that push the boundaries of CRL, offering insights into how these innovations can be harnessed to develop more sophisticated \textit{Artificial Intelligence} (AI) systems.

Table \ref{tab: structure} shows the structure of this survey. 
The primary contributions of this survey are as follows:
\begin{enumerate}
    \item \textit{Challenges}: We highlight the unique challenges faced by CRL, emphasizing the need for a triangular balance among plasticity, stability, and scalability. 
    \item \textit{Scenario Settings}: We categorize CRL scenarios into lifelong adaptation, non-stationarity learning, task incremental learning, and task-agnostic learning. 
    \item \textit{Taxonomy}: We present a new taxonomy of CRL methods, categorizing them based on the type of knowledge stored and/or transferred. 
    \item \textit{Method Review}: We provide the most recent literature review of CRL methods, including detailed discussions of seminal works, recently published articles, and promising preprints.
    \item \textit{Open Challenges}: We discuss open challenges and future research directions in CRL.
\end{enumerate}

\begin{table}
    \caption{The structure of this survey.}
    \label{tab: structure}
    \renewcommand{\arraystretch}{0.5}

    \centering
    \begin{tabular}{@{}c|c|l}
    \toprule
    \multicolumn{3}{c}{\textcolor{red}{$\S~$}\ref{sec:introduction}:~Introduction}  \\ 
    \cmidrule(l){1-3}
    \multicolumn{1}{c|}{\multirow{2}{*}{\begin{tabular}[c]{@{}c@{}} \textcolor{red}{$\S~$}\ref{sec:background} \\ Background \end{tabular}}} &
    \multicolumn{2}{l}{\textcolor{red}{$\S~$}\ref{sec:rl}:~Reinforcement Learning}  \\ \cmidrule(l){2-3} 
    &
    \multicolumn{2}{l}{\textcolor{red}{$\S~$}\ref{sec:cl}:~Continual Learning}  \\ \cmidrule(l){1-3} 
    \multicolumn{1}{c|}{\multirow{6}{*}{\begin{tabular}[c]{@{}c@{}} \textcolor{red}{$\S~$}\ref{sec:overview} \\ Overview \end{tabular}}} &
    \multicolumn{2}{l}{\textcolor{red}{$\S~$}\ref{sec:definition}:~Definition}  \\ \cmidrule(l){2-3} 
    &
    \multicolumn{2}{l}{\textcolor{red}{$\S~$}\ref{sec:challenges}:~Challenges}  \\ \cmidrule(l){2-3} 
    &
    \multicolumn{2}{l}{\textcolor{red}{$\S~$}\ref{sec:metrics}:~Metrics} \\
    \cmidrule(l){2-3} 
    &
    \multicolumn{2}{l}{\textcolor{red}{$\S~$}\ref{sec:benchmarks}:~Benchmarks} \\
    \cmidrule(l){2-3}
    &
    \multicolumn{2}{l}{\textcolor{red}{$\S~$}\ref{sec: scenarios}:~Scenario Settings} \\
    \cmidrule(l){1-3}
    \multicolumn{1}{c|}{\multirow{6}{*}{\begin{tabular}[c]{@{}c@{}} \textcolor{red}{$\S~$}\ref{sec:review} \\ Methods Review \end{tabular}}} &
    \multicolumn{2}{l}{\textcolor{red}{$\S~$}\ref{sec:taxonomy}:~Taxonomy Methodology} \\ \cmidrule(l){2-3} 
    &
    \multicolumn{2}{l}{\textcolor{red}{$\S~$}\ref{sec: policy-based}:~Policy-Focused Methods} \\ \cmidrule(l){2-3} 
    &
    \multicolumn{2}{l}{\textcolor{red}{$\S~$}\ref{sec: experience-focused}:~Experience-Focused Methods}  \\ \cmidrule(l){2-3} 
    &
    \multicolumn{2}{l}{\textcolor{red}{$\S~$}\ref{sec: dynamic-focused}:~Dynamic-Focused Methods} \\ \cmidrule(l){2-3} 
    &
    \multicolumn{2}{l}{\textcolor{red}{$\S~$}\ref{sec: reward-focused}:~Reward-Focused Methods} \\ 
    \cmidrule(l){1-3}
    \multicolumn{1}{c|}{\multirow{5}{*}{\begin{tabular}[c]{@{}c@{}} \textcolor{red}{$\S~$}\ref{sec:future} \\ Future Works \end{tabular}}} &
    \multicolumn{2}{l}{\textcolor{red}{$\S~$}\ref{sec:task-free}:~Task-Free CRL}  \\ \cmidrule(l){2-3} 
    &
    \multicolumn{2}{l}{\textcolor{red}{$\S~$}\ref{sec:ptms}:~Large-Scale Pre-Trained Model}  \\ \cmidrule(l){2-3} 
    &
    \multicolumn{2}{l}{\textcolor{red}{$\S~$}\ref{sec:adjacent}:~Adjacent Directions} \\ 
    \cmidrule(l){1-3} 
    \multicolumn{3}{c}{\textcolor{red}{$\S~$}\ref{sec:conclusion}:~Conclusion}  \\ 
    \bottomrule
    \end{tabular}
    \vspace{-0.5cm}
\end{table}

%% file: sections/background.tex
\subsection{Reinforcement Learning}\label{sec:rl}

RL is a fundamental paradigm in machine learning, where an agent interacts with an environment to learn optimal behaviors through trial and error.
A common framework used to model the interaction is the \textit{Markov Decision Process} (MDP) \cite{martin1990markov}. 
In many real deployments, however, the interaction process is not strictly stationary, and understanding which forms of drift can be handled (or detected) is central to continual formulations \cite{lyle2025WhatCan}.
An MDP comprises a tuple of shared components: state space $\mathcal{S}$, action space $\mathcal{A}$, transition distribution $T(s^\prime \vert s, a) \in [0,1]$, reward function $R: \mathcal{S} \times \mathcal{A} \to \mathbb{R}$, and initial-state distribution $\rho_0$.
Because RL tasks differ in whether they terminate, we distinguish two standard formulations.

\textbf{Episodic (finite-horizon) MDP.}
An episodic MDP is defined as $M_H = \langle \mathcal{S},\mathcal{A},R,T,\rho_0,H \rangle$, where the horizon $H\in\mathbb{N}$ bounds the maximum number of steps per episode.
The agent generates a trajectory $\tau = (s_0, a_0, s_1, \cdots, s_{H-1}, a_{H-1}, s_{H})$ with probability \cite{sutton2018reinforcement}:
\begin{equation}
    P_{\pi}(\tau) := \rho_{0}\prod_{t=0}^{H-1} \pi\left(a_{t} \vert s_{t}\right) T\left(s_{t+1} \vert s_{t},a_{t}\right),
    \label{eq:pi}
\end{equation}
where $\pi:\mathcal{S} \to \Delta (\mathcal{A})$ maps states to action distributions.
The value of a policy is the expected undiscounted return over the episode:
\begin{equation}
    V_{\pi}^{H} := \mathbb{E}_{\tau \sim P_{\pi}}\!\left[\sum_{t=0}^{H-1} R\left(s_{t}, a_{t}\right)\right].
    \label{eq:value_episodic}
\end{equation}

\textbf{Continuing (infinite-horizon) MDP.}
A continuing MDP is defined as $M_\gamma = \langle \mathcal{S},\mathcal{A},R,T,\rho_0,\gamma \rangle$, where $\gamma\in(0,1)$ is a discount factor that ensures convergence of the infinite sum.
In this setting, the interaction does not terminate, and the policy value is \cite{boucherie2017markov}:
\begin{equation}
    V_{\pi}^{\gamma} := \mathbb{E}_{\pi}\!\left[\sum_{t=0}^{\infty} \gamma^{t}\, R\left(s_{t}, a_{t}\right)\right].
    \label{eq:value_continuing}
\end{equation}

This survey adopts whichever formulation matches the CRL setting at hand: $M_H$ when task sequences are episodic and $M_\gamma$ when the agent interacts in a continuing loop without enforced termination.
For episodic tasks that additionally employ discounting, one may replace $R$ by $\gamma^{t}R$ in Eq.~\ref{eq:value_episodic}.
However, we keep the two base cases separate for clarity.
In both settings, the objective of standard RL is to find an optimal policy $\pi^*$ that maximizes the expected return (i.e., $V_{\pi}^{H}$ or $V_{\pi}^{\gamma}$, respectively):
\begin{equation}
    \pi^{\star} := \argmax_\pi \; V_\pi.
    \label{eq:obj}
\end{equation}

\subsection{Continual Learning}\label{sec:cl}

CL is an emerging paradigm in machine learning that focuses on incrementally updating models to adapt to new tasks while maintaining performance on previous tasks. 
In CL, the learner accumulates knowledge over time to enhance the model's effectiveness while saving computational resources and time through incremental modeling, providing a viable solution for a series of tasks under resource constraints \cite{martin2023wholistic}. 
The learning process in CL can be mathematically represented as follows \cite{timothee2020continual}, This formulation abstracts any continual update as a mapping from the prior model, auxiliary knowledge, and incoming task data to the refreshed model and knowledge:
\begin{equation}
    \left\langle {{h_{k - 1}},{U_{k - 1}},D_{k}} \right\rangle  \to \left\langle {{h_k},{U_k}} \right\rangle,
    \label{eq:cl}
\end{equation}
where $h_{k - 1}$ and $h_{k}$ are the CL models for tasks $k-1$ and $k$, respectively; $U_{k - 1}$ and $U_{k}$ are the auxiliary information extracted from tasks $k-1$ and $k$; $D_{k}$ is the incremental data for task $k$. This equation emphasizes that CL centers on iteratively updating the working hypothesis and any auxiliary components as new task data arrives, laying out a concise abstraction for the subsequent CRL specialization.
As the distribution of data changes, the variety of tasks increases, and the complexity of models deepens, CL faces multiple challenges.

Two of the most pressing challenges are \textit{catastrophic forgetting} and \textit{knowledge transfer} \cite{german2019continual}, which have garnered significant attention.
Catastrophic forgetting refers to the degradation of model performance on previous tasks upon learning new ones, a phenomenon attributed to the inherent limitations of current neural network technologies, which diverge from the continuous learning mode of the human brain \cite{kudithipudi2022biological}.
Knowledge transfer, on the other hand, involves leveraging knowledge from previous tasks to facilitate learning on new tasks.
To mitigate catastrophic forgetting and facilitate knowledge transfer, CL approaches strive for a balance between stability, to protect acquired knowledge, and plasticity, to adapt to new tasks efficiently. 
This balance, often referred to as the \textit{stability-plasticity dilemma}, is critical for the success of CL systems.

The diversity of tasks, ranging from classification to robot control, introduces variability in goals, data distributions, and input formats, further complicating the CL landscape \cite{Hsu2018ReevaluatingCL,wang2024comprehensive,vandeVen2022}. 
In embodied and robotic settings, continual adaptation further intertwines perception, semantics, and safety requirements, making the stability--plasticity trade-off especially consequential \cite{rubavicius2025SecureSemanticsaware}.
CRL is a special instance within this diversity, focusing on RL tasks and providing a unique perspective on the challenges and solutions in the CL field.
For broader CRL perspectives and terminology alignment across RL and CL, we refer to recent CRL surveys \cite{liu2025ContinualReinforcement}.

%% file: sections/overview.tex
This section provides an overview of the research on CRL \cite{abel2024definition, abbas2023loss}, focusing on its definition, evaluation, and relevant scenario settings. 
The existing tasks within CRL are described in Appendix C.

\begin{figure}[htbp]
    \vspace{-0.5cm}
    \subfloat[Traditional RL]{\includegraphics[width = 0.23\textwidth]{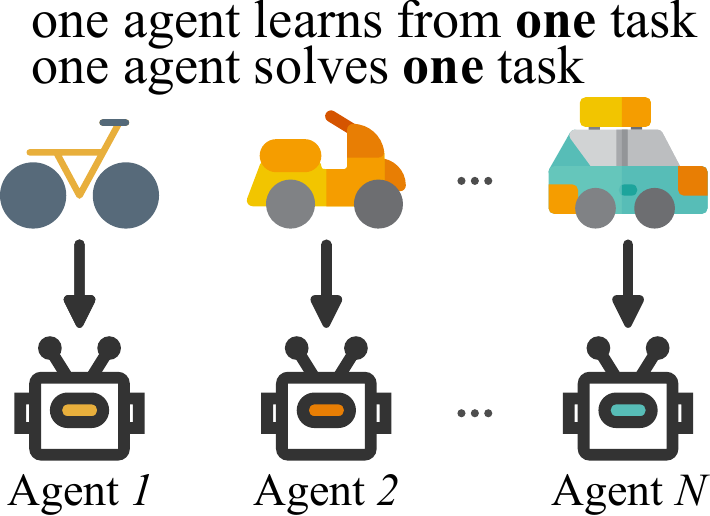}}
    \hfill
    \subfloat[Multi-Task RL]{\includegraphics[width = 0.23\textwidth]{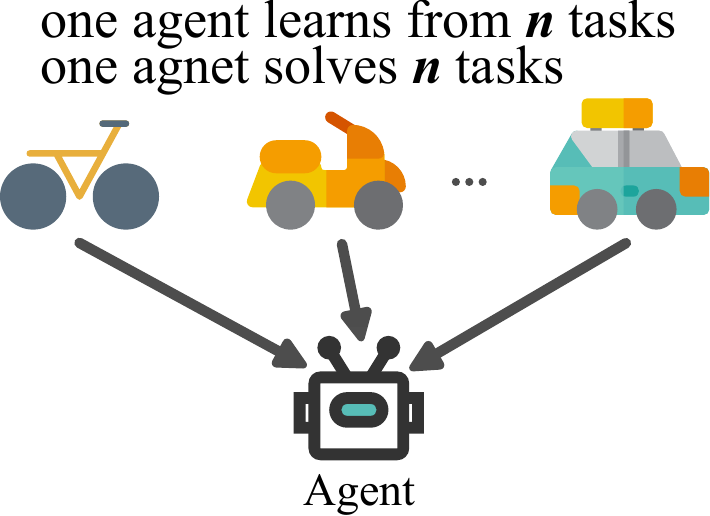}}
    \hfill
    \subfloat[Transfer RL]{\includegraphics[width = 0.23\textwidth]{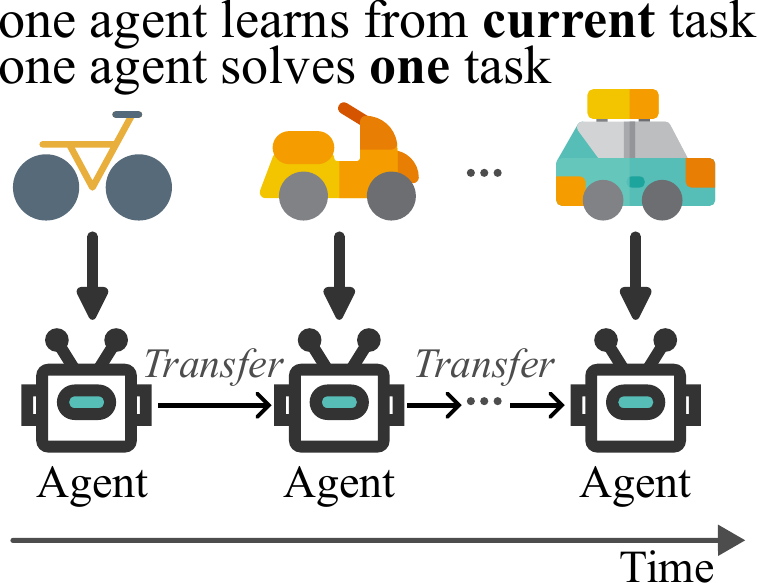}}
    \hfill
    \subfloat[Continual RL]{\includegraphics[width = 0.23\textwidth]{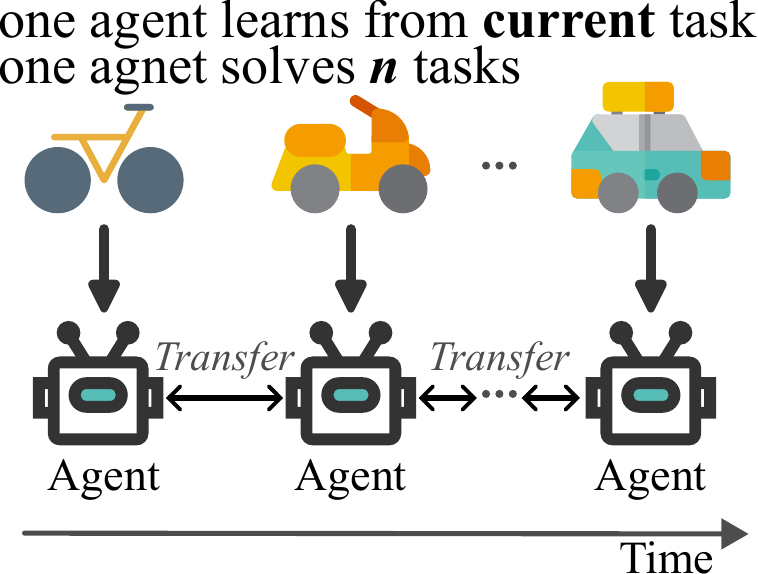}}
    \caption{A comparison of four RL paradigms.}
    \label{fig:4rl}
    \vspace{-0.7cm}
\end{figure}

\subsection{Definition}\label{sec:definition}
The term ``\textit{Continual Reinforcement Learning}'' can be broken down into two main components: ``\textit{continual}'' and ``\textit{reinforcement learning}''. 
While ``\textit{reinforcement learning}'' remains the core subject of study, the term ``\textit{continual}'' emphasizes the extension of traditional RL to a dynamic, multi-task framework, where agents continuously learn, adapt, and retain knowledge across various tasks.
Recent discussions further revisit how to formalize ``continuality'' in RL,
including the implicit assumptions behind task boundaries and non-stationarity abstractions and how these choices affect what should be counted as CRL \cite{elelimy2025rethinking}.

 In CRL, the learning process is typically modeled using MDPs, similar to traditional RL. The general structure includes a state space $\mathcal{S}$, an action space $\mathcal{A}$, an observation space $\mathcal{O}$, a reward function $R: \mathcal{S} \times \mathcal{A} \to \mathbb{R}$, a transition function $T: \mathcal{S} \times \mathcal{A} \to \mathcal{S}$, and an observation function $\Omega: \mathcal{S} \to \mathcal{O}$. The most general form of CRL can be defined as \cite{khetarpal2022towards}:
Eq.~\ref{eq:cl} in Sec.~\ref{sec:cl} frames CL as an abstract update step: prior hypothesis/auxiliary knowledge plus incoming task data produce an updated hypothesis and knowledge.
CRL specializes this abstraction by treating each update step as an RL task (often instantiated as an MDP or POMDP) and interpreting the task data $D_k$ as interaction experience, such as transitions or trajectories collected for that task, depending on whether the setting is online, offline, or a mixture.
\begin{equation}
   M_{\text{CRL}} := \langle \mathcal{S}_{t} ,\mathcal{A}_{t} ,R_{t} ,T_{t} ,\Omega_{t} ,\mathcal{O}_{t}   \rangle .
    \label{eq:crl}
\end{equation}

Eq.~\ref{eq:crl} expresses the most general time-indexed CRL environment abstraction, where every component may depend on the current interaction index $t$, capturing arbitrary non-stationarity through time-varying spaces and functions.
In practical CRL benchmarks, this abstraction is often restricted to piecewise-stationary task sequences: the interaction stream is segmented into tasks indexed by $k$, only a subset of tuple elements varies across tasks (often the reward and transition dynamics $(R,T)$, and sometimes the observation components $(\Omega,\mathcal{O})$), and the remaining components (states, actions) stay fixed.
This task segmentation provides the concrete RL instantiation of Eq.~\ref{eq:cl}: for the $k$-th task (environment segment), $D_k$ corresponds to the experience used for learning (collected online or provided offline), and $U_k$ can represent any carried-over auxiliary state such as replay buffers, learned representations, or models.

Several related learning paradigms, such as \textit{Multi-Task Reinforcement Learning} (MTRL) and \textit{Transfer Reinforcement Learning} (TRL)
, also aim to address multiple RL tasks. 
A comparison between traditional RL, MTRL, TRL, and CRL is presented in Fig. \ref{fig:4rl}. 
The goal of MTRL is to train an agent that can handle multiple tasks simultaneously, where both source and target tasks belong to a fixed and known set \cite{varghese2020multitask}. 
TRL focuses on transferring knowledge from source tasks to target tasks, facilitating faster learning on the target tasks \cite{zhu2023transfer}. 
In contrast, CRL is designed for environments that change continuously, with tasks often arriving sequentially over time. 
The primary objective is to enable an agent to accumulate knowledge over extended periods and quickly adapt to new tasks as they arise. 
CRL shares similarities with both MTRL and TRL. 
MTRL typically employs a cross-task shared structure that allows the agent to handle multiple tasks simultaneously and can be viewed as a timeless version of CRL. 
TRL, which focuses on knowledge transfer between tasks, can be regarded as a subset of CRL, where forgetting is not considered.
Thus, CRL can be considered a more generalized learning paradigm that encompasses the domains of MTRL and TRL. 
Even \textit{Continual Supervised Learning} (CSL)  and traditional RL can be viewed as special cases within the broader framework of CRL \cite{abel2024definition}.

\subsection{Challenges}\label{sec:challenges}
\begin{figure}[htbp]
    \centerline{\includegraphics[width=0.8\linewidth]{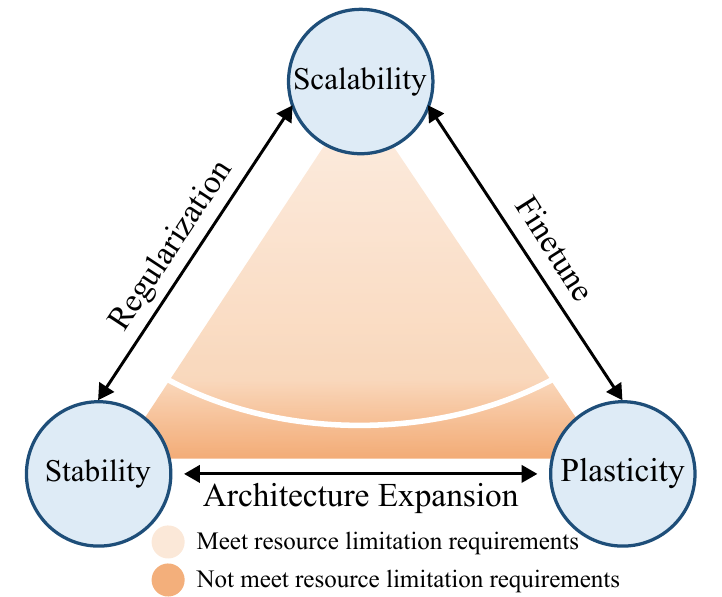}}
    \caption{
        The triangular balance of plasticity, stability, and scalability in CRL.
        Scalability determines the usability of CRL methods, while low scalability fails to meet resource constraints.
        }
    \label{fig: balance}
    \vspace{-0.2cm}
\end{figure}

Research on CRL faces several challenges that distinguish it from traditional RL. 
Inspired by the CL challenges in supervised learning \cite{wang2024comprehensive}, the primary challenge in CRL can be described as achieving a triangular balance among three key aspects: plasticity, stability, and scalability. 
Fig.~\ref{fig: balance} illustrates the relationship between these three aspects:
\begin{enumerate}
    \item \textbf{Stability} refers to an agent's ability to maintain performance on previously learned tasks while simultaneously learning new tasks. 
    Stability is closely related to the problem of catastrophic forgetting, where learning new tasks causes a significant decline in performance on previously learned tasks.
    Addressing catastrophic forgetting is a key focus in CRL research to ensure that the agent retains knowledge over time.
    \item \textbf{Plasticity} refers to an agent's ability to learn new tasks after being trained on previous tasks.
    A critical component of plasticity is the agent's transfer ability, which enables it to leverage knowledge from previously learned tasks to enhance performance on new tasks (forward transfer) or on earlier tasks (backward transfer).
    \item \textbf{Scalability} refers to the ability of an agent to learn many tasks using limited resources. 
    This aspect involves the efficient use of memory and computation, as well as the agent's capacity to handle increasingly complex and diverse task distributions. 
    Scalability is particularly critical in real-world applications, where agents must efficiently adapt to a wide range of tasks.
\end{enumerate}

The balance between these three aspects is crucial for the success of CRL algorithms.
In the early stages of CL research, the primary focus was on stability, with significant efforts directed toward mitigating catastrophic forgetting \cite{james2017overcoming, kaplanis2018continual}. 
In recent years, more CL studies have recognized the importance of plasticity, emphasizing the need for effective knowledge transfer and adaptation to new tasks \cite{abbas2023loss,dohare2024loss}.
Currently, the issue of plasticity loss has even become a crucial concern in DRL research \cite{klein2024plasticity, juliani2024study}.

In the broader CL literature, the interplay between plasticity and stability is often referred to as the \textit{stability-plasticity dilemma} \cite{wang2024comprehensive}, which underscores the inherent trade-off between these two aspects.
Moreover, in the context of CRL, scalability emerges as an equally critical factor. 
Unlike supervised learning, RL algorithms typically require substantial computational resources and memory to learn complex tasks. 
In addition, if resource constraints are ignored, agents could theoretically store all data and train a separate model for each task. 
However, such an approach contradicts the principles of continual learning, which aim to develop resource-efficient algorithms that generalize across tasks \cite{wang2024comprehensive}.
Therefore, CRL must address the challenge of balancing plasticity, stability, and scalability to enable agents to learn effectively in dynamic environments.

\subsection{Metrics}\label{sec:metrics}
 The longstanding tradition in RL is to evaluate a policy on each task by episodic return, success rate, or other reward-based signals.
 In CRL benchmarks, these per-task scores remain the base measurement, but their aggregation and interpretation are defined by the benchmark protocol (task order, evaluation frequency, normalization, and whether task boundaries are observable), which makes results comparable across methods \cite{mesbahi2025position}.
 Following benchmark-native protocols (\eg, Continual World and later suites), we summarize a task sequence by a small set of standardized statistics that capture performance, retention, transfer, and resource constraints \cite{maciej2021advances,tristan2023coom,sam2022cora}.
 To that end, $p_{i,j} \in [0,1]$ denotes the normalized episodic return or success rate on task $j$ after the agent has trained sequentially through tasks $1$ to $i$.

 \textbf{Performance (average performance):} A common benchmark summary is the mean performance over the tasks that have been encountered so far \cite{NIPS2017_f8752278,maciej2021advances,tristan2023coom,sam2022cora}:
\begin{equation}
    A_i := \frac{1}{i} \sum_{j=1}^{i} p_{i,j}.
    \label{eq:acc}
\end{equation}
 The final value $A_N$ is widely reported as an overall performance summary for a length-$N$ task sequence.

 \textbf{Forgetting and retention:} To quantify stability, benchmarks typically report how much performance on an earlier task decays after training on later tasks \cite{james2017overcoming,li2018learning,maciej2021advances,sam2022cora}.
 A benchmark-native choice is to compare the score on task $i$ at the end of training on that task with its final score after completing the full task stream:
\begin{equation}
    FG_i := \max\left(p_{i,i} - p_{N,i},\,0\right).
    \label{eq:fg}
\end{equation}
  The average of $FG_i$ over $i=1,\dots,N-1$ yields the forgetting metric $FG$.
  \emph{Related variants:} The specific choice of reference point is benchmark-dependent \cite{maciej2021advances,sam2022cora}. Examples include the best-so-far performance, results from intermediate checkpoints, performance at the end of training on task $i$, pre-training baselines, or signed differences without a non-negative floor. 

 \textbf{Transfer (forward and backward):} Transfer metrics describe how learning some tasks changes learning speed or asymptotic performance on others \cite{NIPS2017_f8752278,maciej2021advances,tristan2023coom}.
 We start from the benchmark-native definition used in common CRL suites, which treats forward transfer as learning efficiency during each task and measures it via an AUC-based comparison to a single-task baseline \cite{maciej2021advances,tristan2023coom,sam2022cora}.
 \textbf{Forward transfer (FT):} Let $p_i(t)\in[0,1]$ denote the (periodically evaluated) normalized return or success rate on task $i$ at training step $t$.
 For a per-task budget of $\Delta$ steps, define $\mathrm{AUC}_i := \frac{1}{\Delta}\int_{(i-1)\Delta}^{i\Delta} p_i(t)\,\mathrm{d}t$ and $\mathrm{AUC}_i^{b}$ as the corresponding AUC from a reference single-task run.
 The per-task forward transfer is
\begin{equation}
    FT_i := \frac{\mathrm{AUC}_i - \mathrm{AUC}_i^{b}}{1-\mathrm{AUC}_i^{b}},
    \qquad FT := \frac{1}{N-1}\sum_{i=2}^{N} FT_i.
    \label{eq:fwt}
\end{equation}
 Positive $FT$ indicates faster learning than training from scratch on each task, while negative values indicate slowed learning.
 \emph{Related variants:} Some works report zero-shot transfer, which measures performance on task $i$ before any training on it \cite{sam2022cora}, and some CL-derived formulations compute pre-post differences such as $p_{i,i}-p_{i-1,i}$ from checkpointed evaluations. These variants are useful diagnostics, but they should not be conflated with the AUC-based benchmark metric.
 \textbf{Backward transfer (BWT):} As a related, auxiliary notion, backward transfer compares the final performance on earlier tasks with their performance when they were first learned \cite{NIPS2017_f8752278}:
\begin{equation}
    BWT := \frac{1}{N-1} \sum_{i=1}^{N-1} \left(p_{N,i} - p_{i,i}\right).
    \label{eq:bwt}
\end{equation}
  Positive $BWT$ means later training improves earlier tasks. This is sometimes described as \emph{negative forgetting} when forgetting is defined as a signed difference (without a non-negative floor).
  In contrast, our $FG$ in Eq.~\ref{eq:fg} is explicitly floored at zero, so $BWT$ should be treated as a complementary diagnostic rather than a universally co-equal primary metric.

 \textbf{Efficiency and scalability:} Beyond outcome metrics, CRL benchmarks increasingly report resource-related numbers to approximate scalability.
 Importantly, scalability evaluation in CRL still lacks a single standardized quantitative metric, so existing papers usually present practical quantitative proxies that each capture one evaluation dimension rather than serving as a universal standard.
 Common proxies include \textbf{model size} after learning a full task stream \cite{gaya2022building}, \textbf{memory footprint} (e.g., replay buffer size or auxiliary models stored for retention) \cite{maciej2021advances,sam2022cora}, \textbf{training and inference cost} (environment interactions, wall-clock compute, or per-step overhead) \cite{mesbahi2025position}, and \textbf{sample efficiency} (interactions needed to reach a target return or success threshold) \cite{fu2022modelbased}.
 While none of these proxies alone defines scalability, together they help contextualize performance and retention claims under realistic compute and memory constraints.

 \textbf{Additional Metrics:}
 The above metrics provide a benchmark-first evaluation lens and are widely used in CRL suites.
 Furthermore, some benchmarks introduce task-specific summaries such as \textbf{generalization improvement score} \cite{hadi2021continuous} or \textbf{performance relative to a single-task expert} \cite{erik2022l2explorer} to capture aspects that are not well-reflected by a single sequence-level statistic.

\begin{table*}
    \centering
    \caption{Comparison of continual reinforcement learning benchmarks.}
    \label{tab: benchmarks}
    \renewcommand{\arraystretch}{0.8}
    \begin{threeparttable}
    \begin{tabular}{m{3.1cm}|m{1cm}<{\centering}m{1.4cm}<{\centering}m{1.4cm}<{\centering}m{1.5cm}<{\centering}cm{1.7cm}<{\centering}cp{2cm}}
    \hline
    Benchmark & 3D & Number of Sequences  & Length of Sequences &  Partially Observable & Multi-Agent & Image Observation & Metrics\tnote{1}  \\
    \hline
    CRL Maze\cite{lomonaco2020continual} &\CheckmarkBold & 4 & 3 & \CheckmarkBold & \XSolidBrush & \CheckmarkBold & $A_N$ \\
    Lifelong Hanabi\cite{hadi2021continuous} & \XSolidBrush & \XSolidBrush & \XSolidBrush & \CheckmarkBold & \CheckmarkBold & \XSolidBrush & $A_N, FG, FT, GIS$ \\
    Continual World\cite{maciej2021advances} & \CheckmarkBold & 2 & 10,20 & \XSolidBrush & \XSolidBrush & \XSolidBrush & $A_N, FG, FT$ \\
    L2Explorer\cite{erik2022l2explorer} & \CheckmarkBold & \XSolidBrush & \XSolidBrush & \CheckmarkBold & \XSolidBrush & \CheckmarkBold & $FG, FT, BT, PR, SE$ \\
    CORA\cite{sam2022cora}\tnote{2} & \CheckmarkBold \& \XSolidBrush & 4 & 4,6,15 & \CheckmarkBold \& \XSolidBrush &  \XSolidBrush & \CheckmarkBold \& \XSolidBrush & $A_N, FG, FT$ \\
    Lifelong Manipulation\cite{yang2022evaluations} & \CheckmarkBold & \XSolidBrush & 10 & \XSolidBrush & \XSolidBrush & \XSolidBrush & $A_N$\\
    COOM\cite{tristan2023coom} & \CheckmarkBold & 7 & 4,8,16 & \CheckmarkBold & \XSolidBrush & \CheckmarkBold & $A_N, FG, FT$ \\
    \hline
    \end{tabular}
    \begin{tablenotes}
        \footnotesize
        \item[1] ``$A_N$'' stands for the average performance. ``$FG$'' stands for the forgetting. ``$FT$'' stands for the forward transfer. ``$BT$'' stands for the backward transfer. ``$GIS$'' stands for the generalization improvement score. ``$PR$'' stands for the performance relative to a single-task expert. ``$SE$'' stands for the sample efficiency.
        \item[2] CORA is based on four environments with different features.
    \end{tablenotes}
\end{threeparttable}
\vspace{-0.3cm}
\end{table*}

\subsection{Benchmarks}\label{sec:benchmarks}
Although CRL has gained increasing attention in recent years, its growth has been relatively slow compared to CSL.
One of the reasons for this slow development is the difficulty of reproduction and the large amount of computation required for experiments.
Another reason is the lack of standardized benchmarks and metrics for evaluation \cite{sam2022cora}.

Recently, several benchmarks have been proposed for CRL, including dedicated suites that focus on standardized task streams and reproducible protocols \cite{libero2023benchmarking,kobanda2025benchmark}. 
Table \ref{tab: benchmarks} provides a comparison of these benchmarks \cite{tristan2023coom}. 
These benchmarks vary in characteristics such as the number of tasks, the length of task sequences, and the type of observations. 
Below, we briefly describe the notable features of these benchmarks:
\begin{itemize}
    \item \textbf{CRL Maze} \cite{lomonaco2020continual} \footnote{\url{https://github.com/Pervasive-AI-Lab/crlmaze}}: A 3D environment based on ViZDoom, featuring non-stationary object-picking tasks with modified attributes such as light, textures, and objects.
    \item \textbf{Lifelong Hanabi} \cite{hadi2021continuous} \footnote{\url{https://github.com/chandar-lab/Lifelong-Hanabi}}: A partially observable multi-agent environment based on the card game Hanabi \cite{bard2020hanabi}. It challenges agents to cooperate and adapt in a dynamic, multi-agent setting.
    \item \textbf{Continual World} \cite{maciej2021advances} \footnote{\url{https://github.com/awarelab/continual_world}}: This benchmark comprises a set of robotic manipulation tasks derived from Meta-World \cite{yu2020metaworld}, evaluating agents across a diverse set of these operations.
    \item \textbf{L2Explorer} \cite{erik2022l2explorer} \footnote{\url{https://github.com/lifelong-learning-systems/l2explorer}}: A 3D Procedural Content Generation (PCG) world that includes five tasks within a single environment, providing a highly configurable and diverse set of challenges for CRL algorithms.
    \item \textbf{CORA} \cite{sam2022cora} \footnote{\url{https://github.com/AGI-Labs/continual_rl}}: Based on four different environments, each with unique features, CORA offers a comprehensive evaluation platform for CRL algorithms.
    \item \textbf{Lifelong Manipulation} \cite{yang2022evaluations}: This benchmark includes ten manipulation tasks designed to evaluate agents at different levels of difficulty. It is easier to train compared to the Continual World.
    \item \textbf{COOM} \cite{tristan2023coom} \footnote{\url{https://github.com/hyintell/COOM}}: Another benchmark based on various ViZDoom environments, COOM focuses on embodied perception tasks, providing a robust platform for evaluating CRL algorithms.
\end{itemize}

\begin{table*}
    \centering
    \caption{A formal comparison of typical continual reinforcement learning scenarios.
    }
    \label{tab: scenarios}
    \begin{threeparttable}
    \renewcommand{\arraystretch}{1.2} 
    \begin{tabular}{m{4cm}|m{6cm}<{\centering}|m{6cm}<{\centering}}
    \hline
    Scenario & Learning\tnote{1} & Evaluation \\
    \hline
    Lifelong Adaptation & $\{M_k\}_{k=1}^{K}$ & $M_K$\\
    Non-Stationarity Learning & $\{M_k\}_{k=1}^{K}, R_i \neq R_j \text{ or } T_i \neq T_j \text{ for } i\neq j$  & $\{M_k\}_{k=1}^{K}$, $k$ is available \\
    Task Incremental Learning & $\{M_k\}_{k=1}^{K}, R_i \neq R_j \text{ and } T_i \neq T_j \text{ for } i\neq j$ & $\{M_k\}_{k=1}^{K}$, $k$ is available \\
    Task-Agnostic Learning & $\{M_k\}_{k=1}^{K}$ & $\{M_k\}_{k=1}^{K}$, $k$ is unavailable \\
    \hline
    \end{tabular}
    \begin{tablenotes}
        \item[1] $M_k$ is the MDP of task $k$, $K$ is the identifier of the latest task, $R_i$ is the reward function of task $i$, and $T_i$ is the transition function of task $i$.
    \end{tablenotes}
    \end{threeparttable}
    \vspace{-0.5cm}
\end{table*}

The primary challenge in creating benchmarks for CRL lies in the design of task sequences. 
This is a complex engineering task that requires careful consideration of various factors, such as task difficulty, task order, and the number of tasks. 
Currently, there is no single ideal benchmark, and different benchmarks focus on different aspects of CRL \cite{sam2022cora}. 
Therefore, building a comprehensive and standardized benchmark for CRL remains an ongoing challenge.

\subsection{Scenario Settings}\label{sec: scenarios}
CRL scenarios can be categorized into four main types based on the non-stationarity property and the availability of task identities. 
A formal comparison of these scenarios is provided in Table \ref{tab: scenarios}, which outlines the learning and evaluation processes for each scenario type.
We summarize the key scenario types as follows:

\input{figures/timeline.tex}

\textbf{Lifelong Adaptation:} The agent is trained on a sequence of tasks, and its performance is evaluated only on new tasks. 
This scenario was prevalent in the early stages of CRL research \cite{abel2018policy, garcia2019advances} and shares similarities with TRL, albeit with continual tasks. Lifelong adaptation can be viewed as a subproblem within the broader CRL framework, as its focus is on adapting to new tasks without addressing the full range of CRL challenges.

\textbf{Non-Stationarity Learning:} The tasks in the sequence differ in terms of their reward functions \cite{kessler2022same,gaya2022building} or transition functions \cite{powers2022self,zhang2024dynamics}, but they share the same underlying logic.
The agent is evaluated on all tasks in the sequence. 
While some studies have explored non-stationarity in action space or state space within lifelong adaptation settings \cite{actionset2020chandak, ding2023incremental}, this specific issue has not been thoroughly investigated within the broader context of CRL.

\textbf{Task Incremental Learning:} The tasks in the sequence differ significantly from one another in terms of both reward and transition functions \cite{riemer2022continual, sam2022cora, tristan2023coom}.
These tasks are more distinct compared to those in non-stationary learning.
Some tasks may even have different state and action spaces \cite{Hihn2023}. 
Moreover, a few studies have extended this scenario to include tasks from different domains \cite{ammar2015autonomous, qian2021zeroshot}, increasing the diversity of tasks the agent must learn.

\textbf{Task-Agnostic Learning:} The agent is trained on a sequence of tasks without full knowledge of task labels or identities \cite{powers2022self,jacobson2022task}.
The agent might not even be aware of task boundaries, requiring it to infer task changes from the data itself \cite{xu2020task, caccia2023task, anand2023prediction}.
This scenario is particularly relevant to real-world applications, where agents often do not have explicit knowledge of the tasks they are solving or the states they are in. 

In most CRL research, it is assumed that each task provides a sufficient number of steps for the agent to learn. 
Typically, the task sequence is fixed. However, some studies have explored dynamically generated task sequences \cite{powers2022self, wang2023dirichlet}. 
Recent research has also increasingly focused on scenarios where agents are unable to observe task boundaries, which adds a level of realism and complexity to the problem.
Therefore, some researchers consider this scenario as CRL and refer to scenarios that do not fully satisfy these criteria as semi-continual RL \cite{anand2023prediction}.
While the above categories offer a structured view of CRL scenarios, it is important to note that the boundaries between these scenarios are not always clear-cut. 
Many studies integrate multiple scenarios to address more complex, general CRL problems \cite{gaya2022building, fu2022modelbased}.

To connect scenario settings, challenge emphasis, taxonomy categories, and scalability considerations in a single view, Table~\ref{tab:integration} summarizes representative CRL families discussed in this survey.
The ``Typical Scenario'' column follows the scenario categories in Table~\ref{tab: scenarios}.
For scalability, we summarize qualitative resource-growth trends (memory, compute, and task-scaling) alongside brief bottleneck notes.

\begin{table*}
    \centering
    \begin{threeparttable}
    \caption{Unified view linking CRL methods to scenarios, challenge emphasis, and scalability profiles.\tnote{3}}
    \label{tab:integration}
    \renewcommand{\arraystretch}{1.1}
    \begin{tabular}{m{3.6cm}|m{2.3cm}|m{2.9cm}|m{3.1cm}|m{4cm}}
    \hline
    \textbf{Method Family (examples)} & \textbf{Taxonomy Category} & \textbf{Typical Scenario} & \textbf{Challenge Focus (P/S/Sc)}\tnote{1} & \textbf{Scalability Profile}\tnote{2} \\
    \hline
    Policy reuse: init. (MAXQINIT \cite{abel2018policy}, CSP \cite{gaya2022building}) & policy-focused & lifelong adaptation & P-high / S-low / Sc-med & memory growth: med; compute growth: med; task-scaling: med \\
    Policy decomposition (OWL \cite{kessler2022same}, DaCoRL \cite{zhang2024dynamics}) & policy-focused & task-agnostic learning & P-high / S-med / Sc-high & memory growth: low; compute growth: high; task-scaling: high \\
    Policy merging: Regularization (EWC \cite{james2017overcoming}) & policy-focused & task incremental learning & P-med / S-high / Sc-low & memory growth: low; compute growth: med; task-scaling: low \\
    Policy merging: Masks (MASK \cite{beniwhiwhu2023lifelong}) & policy-focused & task incremental learning & P-low / S-high / Sc-low & memory growth: med; compute growth: low; task-scaling: low \\
    Policy merging: Distillation (DisCoRL \cite{traore2019discorl}) & policy-focused & task incremental learning & P-med / S-high / Sc-med & memory growth: low; compute growth: high; task-scaling: med \\
    Policy merging: Hypernets (HN-PPO \cite{pfilemon2022hypernetworkppo}) & policy-focused & task incremental learning & P-high / S-med / Sc-med & memory growth: low; compute growth: high; task-scaling: med \\
    Direct replay (CLEAR \cite{rolnick2019experience}, Selective \cite{isele2018selective}) & experience-focused & non-stationarity learning & P-med / S-high / Sc-med & memory growth: high; compute growth: high; task-scaling: med \\
    Generative replay (RePR \cite{craig2021pseudo}, SLER \cite{li2021sler}) & experience-focused & task incremental learning & P-med / S-high / Sc-med & memory growth: low; compute growth: med; task-scaling: med \\
    Direct modeling (MOLe \cite{nagabandi2018deep}, HyperCRL \cite{huang2021continual}) & dynamic-focused & non-stationarity learning & P-med / S-high / Sc-low & memory growth: high; compute growth: high; task-scaling: med \\
    Indirect modeling (LILAC \cite{xie2020DeepReinforcement}, Dreamer \cite{kessler2023effectiveness}) & dynamic-focused & task-agnostic learning & P-high / S-med / Sc-high & memory growth: low; compute growth: med; task-scaling: high \\
    Reward-focused (ELIRL \cite{mendez2018lifelong}, IML \cite{bougie2021intrinsically}) & reward-focused & task incremental learning & P-high / S-med / Sc-med & memory growth: low; compute growth: med; task-scaling: med \\
    \hline
    \end{tabular}
    \begin{tablenotes}
        \item[1] P: Plasticity; S: Stability; Sc: Scalability. 
        \item[2] For growth metrics (memory/compute), \textit{low} is preferable; for \textit{task-scaling}, \textit{high} indicates better ability to handle long task streams. \item[3] Note that the properties described in this table are representative general trends for each family. Individual methods may exhibit different characteristics depending on specific algorithmic choices and implementations.
    \end{tablenotes}
    \end{threeparttable}
    \vspace{-0.5cm}
\end{table*}

%% file: figures/timeline.tex
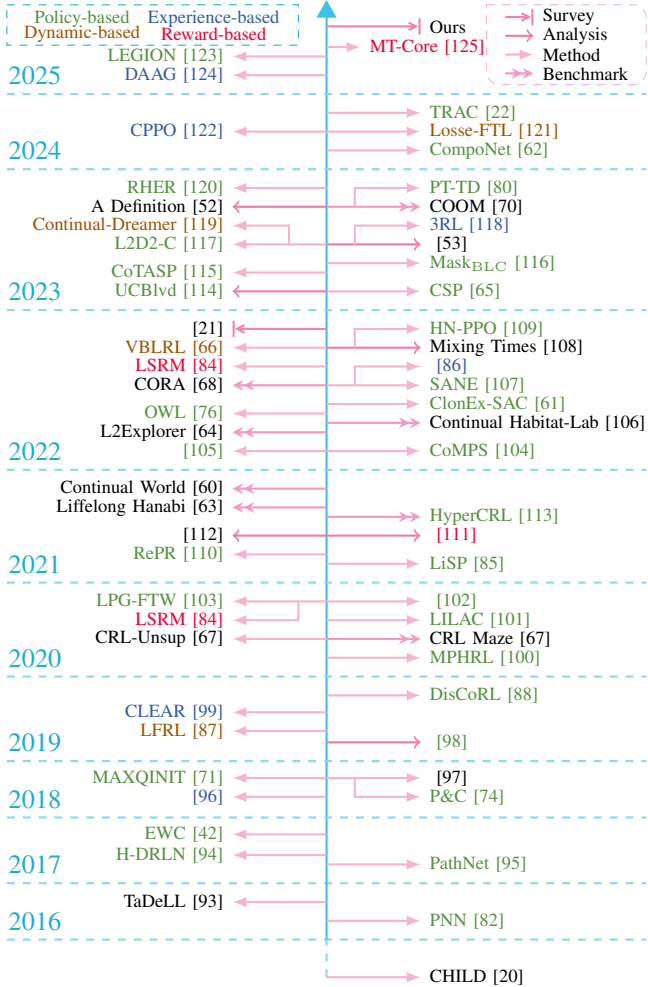
\begin{figure}
    \definecolor{yearblue}{RGB}{30, 173, 211}
    \definecolor{policygreen}{RGB}{84, 145, 64}
    \definecolor{dynamicyellow}{RGB}{158, 84, 2}
    \definecolor{experienceblue}{RGB}{52, 91, 168}
    \definecolor{rewardpink}{RGB}{241, 0, 48}
    \tikzstyle{cic}=[fill=none, draw=black, shape=circle]
    \tikzstyle{new style 0}=[fill=black, draw=none, shape=circle, minimum size=0.5cm]
    \tikzstyle{new style 1}=[fill=none, draw=black, shape=rectangle]
    \tikzstyle{w_text}=[fill=none, draw=none, anchor=west, font={\scriptsize}]
    \tikzstyle{e_test}=[fill=none, draw=none, anchor=east, font={\scriptsize}]
    \tikzstyle{policy}=[text=policygreen]
    \tikzstyle{experience}=[text=experienceblue]
    \tikzstyle{reward}=[text=rewardpink]
    \tikzstyle{dynamic}=[text=dynamicyellow]
    \tikzstyle{year node}=[fill=none, draw=none, text=yearblue]

    \tikzstyle{arrow}=[->, >=latex]
    \tikzstyle{new edge style 0}=[dashed, ->]
    \tikzstyle{area}=[-, fill={gray!30}, draw=none]
    \tikzstyle{blodline}=[-, line width=3pt, line cap=round, line join=round]
    \tikzstyle{method line}=[->,>=latex, draw={rgb,255: red,245; green,179; blue,207}, thick]
    \tikzstyle{benchmark line}=[draw={rgb,255: red,247; green,150; blue,192}, ->>, thick, >=stealth]
    \tikzstyle{analysis line}=[draw={rgb,255: red,238; green,123; blue,172}, thick, ->, >=to]
    \tikzstyle{survey line}=[draw={rgb,255: red,248; green,129; blue,167}, ->|, >={Straight Barb[length=1mm,width=1mm]}, thick]
    \tikzstyle{time line}=[draw={rgb,255: red,60; green,194; blue,229}, ->, >={triangle 60}, thick]
    \tikzstyle{split line}=[-, draw={rgb,255: red,136; green,217; blue,240}, dashed, thick]

    \centering
    \begin{tikzpicture}[scale=0.5]
        \begin{pgfonlayer}{nodelayer}
            \node [style=none] (0) at (-8.5, 0) {};
            \node [style=none] (1) at (8.5, 0) {};
            \node [style=year node] (2) at (-7.75, 0.5) {2016};
            \node [style={w_text}] (4) at (2.5, -1) {CHILD\cite{ring1997child}};
            \node [style=none] (5) at (0, -1) {};
            \node [style=none] (6) at (0, 0) {};
            \node [style=none] (7) at (-8.5, 1.5) {};
            \node [style=none] (8) at (8.5, 1.5) {};
		\node [style=none] (9) at (0, 27.5) {};
            \node [style=none] (10) at (0, 1) {};
            \node [style=none] (11) at (0, 0.5) {};
            \node [style={e_test}] (12) at (-2.5, 1) {TaDeLL\cite{isele2016UsingTask}};
            \node [style={w_text,policygreen}] (13) at (2.5, 0.5) {PNN\cite{rusu2016progressive}};
            \node [style=none] (14) at (-8.5, 3.25) {};
            \node [style=none] (15) at (8.5, 3.25) {};
            \node [style=none] (16) at (-8.5, 4.75) {};
            \node [style=none] (17) at (8.5, 4.75) {};
            \node [style=none] (18) at (-8.5, 7) {};
            \node [style=none] (19) at (8.5, 7) {};
            \node [style=none] (20) at (-8.5, 9.5) {};
            \node [style=none] (21) at (8.5, 9.5) {};
            \node [style=none] (24) at (-8.5, 12.5) {};
            \node [style=none] (25) at (8.5, 12.5) {};
            \node [style=year node] (26) at (-7.75, 2) {2017};
            \node [style=year node] (27) at (-7.75, 3.75) {2018};
            \node [style=year node] (28) at (-7.75, 5.25) {2019};
            \node [style=year node] (29) at (-7.75, 7.5) {2020};
            \node [style=year node] (30) at (-7.75, 10) {2021};
            \node [style=year node] (32) at (-7.75, 21) {2024};
            \node [style=year node] (33) at (-7.75, 13) {2022};
            \node [style=year node] (34) at (-7.75, 23) {2025};
            \node [style=none] (35) at (-8.5, 20.5) {};
            \node [style=none] (36) at (8.5, 20.5) {};
            \node [style=none] (37) at (-8.5, 22.5) {};
            \node [style=none] (38) at (8.5, 22.5) {};
            \node [style=none] (39) at (0, 2.25) {};
            \node [style=none] (40) at (0, 2) {};
            \node [style={e_test,policygreen}] (41) at (-2.5, 2.25) {H-DRLN\cite{tessler2017DeepHierarchical}};
            \node [style={w_text,policygreen}] (42) at (2.5, 2) {PathNet\cite{fernando2017pathnet}};
            \node [style=none] (43) at (0, 2.8) {};
            \node [style={e_test,policygreen}] (44) at (-2.5, 2.8) {EWC\cite{james2017overcoming}};
            \node [style=none] (45) at (0, 3.8) {};
            \node [style=none] (46) at (0, 4.3) {};
            \node [style={e_test,experienceblue}] (47) at (-2.5, 3.8) {\cite{isele2018selective}};
            \node [style={w_text}] (48) at (2.5, 4.3) {\cite{abel2018state}};
            \node [style=none] (49) at (0, 4.3) {};
            \node [style={e_test,policygreen}] (50) at (-2.5, 4.3) {MAXQINIT\cite{abel2018policy}};
            \node [style=none] (51) at (0.75, 3.8) {};
            \node [style={w_text,policygreen}] (52) at (2.5, 3.8) {P\&C\cite{schwarz2018progress}};
            \node [style=none] (53) at (0.75, 4.3) {};
            \node [style=none] (54) at (0, 5.25) {};
            \node [style={w_text,policygreen}] (55) at (2.5, 5.25) {\cite{van2019composing}};
            \node [style=none] (56) at (0, 6.5) {};
            \node [style={w_text,policygreen
            }] (57) at (2.5, 6.5) {DisCoRL\cite{traore2019discorl}};
            \node [style=none] (58) at (0, 5.55) {};
            \node [style=none] (59) at (0, 5.55) {};
            \node [style={e_test, dynamicyellow}] (60) at (-2.5, 5.55) {LFRL\cite{liu2019lifelong}};
            \node [style=none] (61) at (0, 6.05) {};
            \node [style=none] (62) at (0, 6.05) {};
            \node [style={e_test, experienceblue}] (63) at (-2.5, 6.05) {CLEAR\cite{rolnick2019experience}};
            \node [style=none] (64) at (0, 7.5) {};
            \node [style={w_text, policygreen}] (65) at (2.5, 7.5) {MPHRL\cite{wu2020ModelPrimitives}};
            \node [style=none] (66) at (0, 8) {};
            \node [style={w_text}] (67) at (2.5, 8) {CRL Maze \cite{lomonaco2020continual}};
            \node [style=none] (69) at (0, 8) {};
            \node [style={e_test}] (70) at (-2.5, 8) {CRL-Unsup \cite{lomonaco2020continual}};
            \node [style=none] (71) at (0, 8.5) {};
            \node [style={w_text, policygreen}] (72) at (2.5, 8.5) {LILAC \cite{xie2020DeepReinforcement}};
            \node [style=none] (73) at (0, 8.5) {};
            \node [style=none] (74) at (0, 9) {};
            \node [style={w_text, policygreen}] (75) at (2.5, 9) {\cite{nangue2020boolean}};
            \node [style=none] (76) at (0, 9) {};
            \node [style=none] (77) at (0, 9) {};
            \node [style={e_test, policygreen}] (78) at (-2.5, 9) {LPG-FTW \cite{mendez2020lifelong}};
            \node [style=none] (79) at (-0.75, 9) {};
            \node [style={e_test, rewardpink}] (80) at (-2.5, 8.5) {LSRM \cite{zheng2022lifelong}};
            \node [style=none] (81) at (-0.75, 8.5) {};
            \node [style=none] (82) at (-8.5, 16.75) {};
            \node [style=none] (83) at (8.5, 16.75) {};
            \node [style=year node] (84) at (-7.75, 17.25) {2023};
            \node [style=none] (85) at (0, 13) {};
            \node [style={w_text, policygreen}] (86) at (2.5, 13) {CoMPS \cite{berseth2022comps}};
            \node [style={e_test, policygreen}] (87) at (-2.5, 13) {\cite{mendez2022modular}};
            \node [style=none] (88) at (0, 13.5) {};
            \node [style={e_test}] (89) at (-2.5, 13.5) {L2Explorer \cite{erik2022l2explorer}};
            \node [style=none] (90) at (0, 13.75) {};
            \node [style={w_text}] (91) at (2.5, 13.75) {Continual Habitat-Lab \cite{lucchesi2022avalanche}};
            \node [style=none] (92) at (0, 14) {};
            \node [style={e_test, policygreen}] (93) at (-2.5, 14) {OWL\cite{kessler2022same}};
            \node [style=none] (94) at (0, 14.25) {};
            \node [style={w_text, policygreen}] (95) at (2.5, 14.25) {ClonEx-SAC\cite{wolczyk2022disentangling}};
            \node [style=none] (96) at (0, 14.75) {};
            \node [style={w_text, policygreen}] (97) at (2.5, 14.75) {SANE\cite{powers2022self}};
            \node [style={e_test}] (98) at (-2.5, 14.75) {CORA\cite{sam2022cora}};
            \node [style=none] (99) at (0.75, 14.75) {};
            \node [style=none] (100) at (0.75, 15.25) {};
            \node [style={w_text, experienceblue}] (101) at (2.5, 15.25) {\cite{daniels2022modelfree}};
            \node [style=none] (102) at (0, 15.25) {};
            \node [style={e_test, rewardpink}] (103) at (-2.5, 15.25) {LSRM\cite{zheng2022lifelong}};
            \node [style=none] (104) at (0, 15.75) {};
            \node [style={w_text}] (105) at (2.5, 15.75) {Mixing Times\cite{riemer2022continual}};
            \node [style=none] (106) at (0.75, 15.75) {};
            \node [style=none] (107) at (0.75, 16.25) {};
            \node [style={w_text, policygreen}] (108) at (2.5, 16.25) {HN-PPO\cite{pfilemon2022hypernetworkppo}};
            \node [style=none] (109) at (0, 15.75) {};
            \node [style={e_test, dynamicyellow}] (110) at (-2.5, 15.75) {VBLRL\cite{fu2022modelbased}};
            \node [style=none] (111) at (0, 16.25) {};
            \node [style={e_test}] (112) at (-2.5, 16.25) {\cite{khetarpal2022towards}};
            \node [style=none] (113) at (0, 10) {};
            \node [style={w_text, policygreen}] (114) at (2.5, 10) {LiSP\cite{lu2021resetfree}};
            \node [style=none] (115) at (0, 10.25) {};
            \node [style={e_test, policygreen}] (116) at (-2.5, 10.25) {RePR\cite{craig2021pseudo}};
            \node [style=none] (117) at (0, 10.75) {};
            \node [style={w_text, rewardpink}] (118) at (2.5, 10.75) {\cite{jiang2021temporal}};
            \node [style=none] (119) at (0, 10.75) {};
            \node [style={e_test}] (120) at (-2.5, 10.75) {\cite{lecarpentier2021lipschitz}};
            \node [style=none] (121) at (0, 11.25) {};
            \node [style={w_text, policygreen}] (122) at (2.5, 11.25) {HyperCRL\cite{huang2021continual}};
            \node [style=none] (123) at (0, 11.5) {};
            \node [style={e_test}] (124) at (-2.5, 11.5) {Liffelong Hanabi\cite{hadi2021continuous}};
            \node [style=none] (125) at (0, 12) {};
            \node [style={e_test}] (126) at (-2.5, 12) {Continual World\cite{maciej2021advances}};
            \node [style=none] (127) at (0, 17.25) {};
            \node [style={w_text, policygreen}] (128) at (2.5, 17.25) {CSP \cite{gaya2022building}};
            \node [style=none] (129) at (0, 17.25) {};
            \node [style={e_test, policygreen}] (130) at (-2.5, 17.25) {UCBlvd \cite{amani2023provably}};
            \node [style=none] (132) at (0, 17.75) {};
            \node [style={e_test, policygreen}] (133) at (-2.5, 17.75) {CoTASP \cite{yang2023ContinualTask}};
            \node [style={w_text, policygreen}] (134) at (2.5, 18) {Mask$_{\rm{BLC}}$ \cite{beniwhiwhu2023lifelong}};
            \node [style=none] (135) at (0, 18) {};
            \node [style={w_text}] (136) at (2.5, 18.5) {\cite{abbas2023loss}};
            \node [style=none] (137) at (0, 18.5) {};
            \node [style={e_test, policygreen}] (138) at (-2.5, 18.5) {L2D2-C\cite{nath2023sharing}};
            \node [style=none] (139) at (0.75, 18.5) {};
            \node [style=none] (140) at (0.75, 19) {};
            \node [style={w_text, experienceblue}] (141) at (2.5, 19) {3RL \cite{caccia2023task}};
            \node [style=none] (142) at (-1, 18.5) {};
            \node [style=none] (143) at (-1, 19) {};
            \node [style={e_test, dynamicyellow}] (144) at (-2.5, 19) {Continual-Dreamer\cite{kessler2023effectiveness}};
            \node [style={w_text}] (145) at (2.5, 19.5) {COOM\cite{tristan2023coom}};
            \node [style=none] (146) at (0, 19.5) {};
            \node [style={e_test}] (147) at (-2.5, 19.5) {A Definition\cite{abel2024definition}};
            \node [style=none] (148) at (0.75, 19.5) {};
            \node [style=none] (149) at (0.75, 20) {};
            \node [style={w_text, policygreen}] (150) at (2.5, 20) {PT-TD\cite{anand2023prediction}};
            \node [style=none] (152) at (0, 20) {};
            \node [style={e_test, policygreen}] (153) at (-2.5, 20) {RHER\cite{luo2023relay}};
            \node [style={w_text, policygreen}] (154) at (2.5, 21) {CompoNet\cite{malagon2024self}};
            \node [style=none] (155) at (0, 21) {};
            \node [style={w_text, dynamicyellow}] (156) at (2.5, 21.5) {Losse-FTL \cite{liu2024locality}};
            \node [style=none] (157) at (0, 21.5) {};
            \node [style={e_test, experienceblue}] (158) at (-2.5, 21.5) {CPPO \cite{zhang2024cppo}};
            \node [style={w_text, policygreen}] (159) at (2.5, 22) {TRAC \cite{muppidi2024fast}};
		\node [style=none] (197) at (0, 25) {};
		\node [style={e_test}, policygreen] (198) at (-2.5, 25) {RPG\cite{yao2025GeneralRelation}};
		\node [style={w_text, rewardpink}] (199) at (2.5, 25) {MT-Core\cite{pan2025multigranularity}};
		\node [style=none] (200) at (0, 23.7) {};
		\node [style={e_test}, policygreen] (202) at (-2.5, 23.7) {C-CHAIN\cite{tang2025mitigating}};
		\node [style={w_text}] (201) at (2.5, 23.7) {Grokking\cite{lyle2025WhatCan}};
		\node [style={e_test, policygreen}] (163) at (-2.5, 22.75) {LEGION \cite{meng2025preserving}};
		\node [style={w_text, experienceblue}] (185) at (2.5, 23) {DAAG \cite{palo2025diffusion}};
		\node [style=none] (186) at (-8.25, 26.55) {};
		\node [style=none] (187) at (-0.5, 26.55) {};
		\node [style=none] (188) at (-0.5, 27.625) {};
		\node [style=none] (189) at (-8.25, 27.625) {};
		\node [style=policygreen] (191) at (-6.25, 27.25) {\scriptsize Policy-based};
		\node [style=experienceblue] (192) at (-2.75, 27.25) {\scriptsize Experience-based};
		\node [style=dynamicyellow] (193) at (-6.25, 26.85) {\scriptsize Dynamic-based};
		\node [style=rewardpink] (194) at (-2.75, 26.85) {\scriptsize Reward-based};
		\node [style=none] (164) at (0, 22.75) {};
		\node [style={w_text}] (165) at (2.5, 26) {Ours};
		\node [style=none] (166) at (0, 26) {};
		\node [style={w_text}] (195) at (2.4, 24.5) {Continual NavBench\cite{kobanda2025benchmark}};
		\node [style={e_test, policygreen}] (207) at (-2.5, 25.75) {FAME\cite{sun2026principled}};
		\node [style=none] (196) at (0, 24.5) {};
		\node [style=none] (167) at (4, 25.75) {};
		\node [style=none] (168) at (8.25, 25.75) {};
		\node [style=none] (169) at (8.25, 27.875) {};
		\node [style=none] (170) at (4, 27.875) {};
		\node [style={w_text}] (171) at (5.25, 26.55) {Method};
		\node [style={w_text}] (173) at (5.25, 26.05) {Benchmark};
		\node [style={w_text}] (174) at (5.25, 27.05) {Analysis};
		\node [style={w_text}] (175) at (5.25, 27.55) {Survey};
		\node [style=none] (176) at (4.5, 26.55) {};
		\node [style=none] (177) at (5.25, 26.55) {};
		\node [style=none] (178) at (4.5, 26.05) {};
		\node [style=none] (179) at (5.25, 26.05) {};
		\node [style=none] (180) at (4.5, 27.55) {};
		\node [style=none] (181) at (5.25, 27.55) {};
		\node [style=none] (182) at (4.5, 27.05) {};
		\node [style=none] (183) at (5.25, 27.05) {};
		\node [style=none] (184) at (0, 23) {};
		\node [style=none] (205) at (0, 25) {};
		\node [style=year node] (206) at (-7.75, 26) {2026};
		\node [style=none] (208) at (0, 25.75) {};
		\node [style=none] (209) at (-8.5, 25.5) {};
		\node [style=none] (210) at (8.5, 25.5) {};
		\node [style={e_test, policygreen}] (211) at (-2.5, 23.25) {MacPro\cite{yuan2025MultiagentContinual}};
		\node [style=none] (212) at (0, 23.25) {};
		\node [style=none] (213) at (-1, 23.7) {};
		\node [style=none] (214) at (-1, 24.25) {};
		\node [style={e_test}, dynamicyellow] (215) at (-2.5, 24.25) {DRAGO\cite{fu2025knowledge}};
        \end{pgfonlayer}
        \begin{pgfonlayer}{edgelayer}
            \draw [style=split line] (0.center) to (1.center);
            \draw [style=split line] (5.center) to (6.center);
            \draw [style=method line] (5.center) to (4);
            \draw [style=split line] (7.center) to (8.center);
            \draw [style=time line] (6.center) to (9.center);
            \draw [style=method line] (10.center) to (12);
            \draw [style=method line] (11.center) to (13);
            \draw [style=split line] (14.center) to (15.center);
            \draw [style=split line] (16.center) to (17.center);
            \draw [style=split line] (18.center) to (19.center);
            \draw [style=split line] (20.center) to (21.center);
            \draw [style=split line] (24.center) to (25.center);
            \draw [style=split line] (35.center) to (36.center);
            \draw [style=split line] (37.center) to (38.center);
            \draw [style=method line] (39.center) to (41);
            \draw [style=method line] (40.center) to (42);
            \draw [style=method line] (43.center) to (44);
            \draw [style=method line] (45.center) to (47);
            \draw [style=method line] (46.center) to (48);
            \draw [style=method line] (49.center) to (50);
            \draw [style=method line] (53.center)
                 to (51.center)
                 to (52);
            \draw [style=analysis line] (54.center) to (55);
            \draw [style=method line] (56.center) to (57);
            \draw [style=method line] (59.center) to (60);
            \draw [style=method line] (62.center) to (63);
            \draw [style=method line] (64.center) to (65);
            \draw [style=benchmark line] (66.center) to (67);
            \draw [style=method line] (69.center) to (70);
            \draw [style=method line] (71.center) to (72);
            \draw [style=method line] (74.center) to (75);
            \draw [style=method line] (77.center) to (78);
            \draw [style=method line] (79.center)
                 to (81.center)
                 to (80);
            \draw [style=split line] (82.center) to (83.center);
            \draw [style=method line] (85.center) to (86);
            \draw [style=method line] (85.center) to (87);
            \draw [style=benchmark line] (88.center) to (89);
            \draw [style=benchmark line] (90.center) to (91);
            \draw [style=method line] (92.center) to (93);
            \draw [style=method line] (94.center) to (95);
            \draw [style=method line] (96.center) to (97);
            \draw [style=benchmark line] (96.center) to (98);
            \draw [style=method line] (99.center)
                 to (100.center)
                 to (101);
            \draw [style=method line] (102.center) to (103);
            \draw [style=analysis line] (104.center) to (105);
            \draw [style=method line] (106.center)
                 to (107.center)
                 to (108);
            \draw [style=method line] (109.center) to (110);
            \draw [style=survey line] (111.center) to (112);
            \draw [style=method line] (113.center) to (114);
            \draw [style=method line] (115.center) to (116);
            \draw [style=analysis line] (117.center) to (118);
            \draw [style=analysis line] (119.center) to (120);
            \draw [style=benchmark line] (121.center) to (122);
            \draw [style=benchmark line] (123.center) to (124);
            \draw [style=benchmark line] (125.center) to (126);
            \draw [style=method line] (127.center) to (128);
            \draw [style=analysis line] (129.center) to (130);
            \draw [style=method line] (132.center) to (133);
            \draw [style=method line] (135.center) to (134);
            \draw [style=analysis line] (137.center) to (136);
            \draw [style=method line] (137.center) to (138);
            \draw [style=method line] (139.center)
                 to (140.center)
                 to (141);
            \draw [style=method line] (142.center)
                 to (143.center)
                 to (144);
            \draw [style=benchmark line] (146.center) to (145);
            \draw [style=analysis line] (146.center) to (147);
            \draw [style=method line] (148.center)
                 to (149.center)
                 to (150);
            \draw [style=method line] (152.center) to (153);
            \draw [style=method line] (155.center) to (154);
            \draw [style=method line] (157.center) to (156);
            \draw [style=method line] (157.center) to (158);
            \draw [style=method line] (197.center) to (198);
            \draw [style=method line] (197.center) to (199);
            \draw [style=method line] (200.center) to (202);
            \draw [style=analysis line] (200.center) to (201);
            \draw [style=method line] (164.center) to (163);
            \draw [style=survey line] (166.center) to (165);
            \draw [style=benchmark line] (196.center) to (195);
		\draw [draw={rgb,255: red,245; green,179; blue,207}, dashed, line cap=round, line join=round, rounded corners=5pt] (170.center)
                 to (169.center)
                 to (168.center)
                 to (167.center)
			 to cycle;
            \draw [style=method line] (176.center) to (177.center);
            \draw [style=benchmark line] (178.center) to (179.center);
            \draw [style=survey line] (180.center) to (181.center);
            \draw [style=analysis line] (182.center) to (183.center);
            \draw [style=method line] (184.center) to (185);
            \draw [draw=yearblue, dashed] (189.center)
            to (188.center)
            to (187.center)
            to (186.center)
            to cycle;
		\draw [style=method line] (208.center) to (207);
		\draw [style=split line] (209.center) to (210.center);
		\draw [style=method line] (212.center) to (211);
		\draw [style=method line] (213.center) to (214.center)
          to (215);
        \end{pgfonlayer}
    \end{tikzpicture}   
    \caption{
        Timeline illustrating the key developments, by order and interval, in the field of CRL from the end of 2016 to the present day (2026).
        The timeline includes methods, benchmarks, surveys, and analysis papers published in the field.
        At the start of the timeline, we highlight CHILD, which is described the first CL agent and pioneered the field of CRL.
        We then jump toward the middle of 2016, highlighting PNN, a stepping stone towards a full CRL agent.
        The dates shown in the timeline are the publication dates of the papers.
        }
    \label{fig: timeline}     
    \vspace{-0.6cm}                    
\end{figure}

%% file: sections/review.tex
In this section, we present our taxonomy of CRL methods.
Khetarpal \etal \cite{khetarpal2022towards} proposed a taxonomy for CRL, classifying approaches into three categories: explicit knowledge retention, leveraging shared structures, and learning to learn. 
While this taxonomy provides valuable insights, it does not adequately capture the unique characteristics of CRL, and it falls short of encompassing the breadth of recent advancements in the field.
To address these limitations, we propose a new taxonomy that focuses on the unique aspects of CRL, distinguishing it from traditional CL methods. 
Our taxonomy is grounded in the key components of RL and organizes CRL methods based on the type of knowledge they store and transfer.
In addition, we provide the most updated and comprehensive review of CRL methods, including the latest advancements in the field.
The applications and more related works of CRL are described in Appendix D.
Fig. \ref{fig: timeline} presents a timeline with the representative methods in CRL, allowing one to evaluate the novelty and popularity of each class of methods.

\begin{figure}
    \centerline{\includegraphics[width=.98\linewidth]{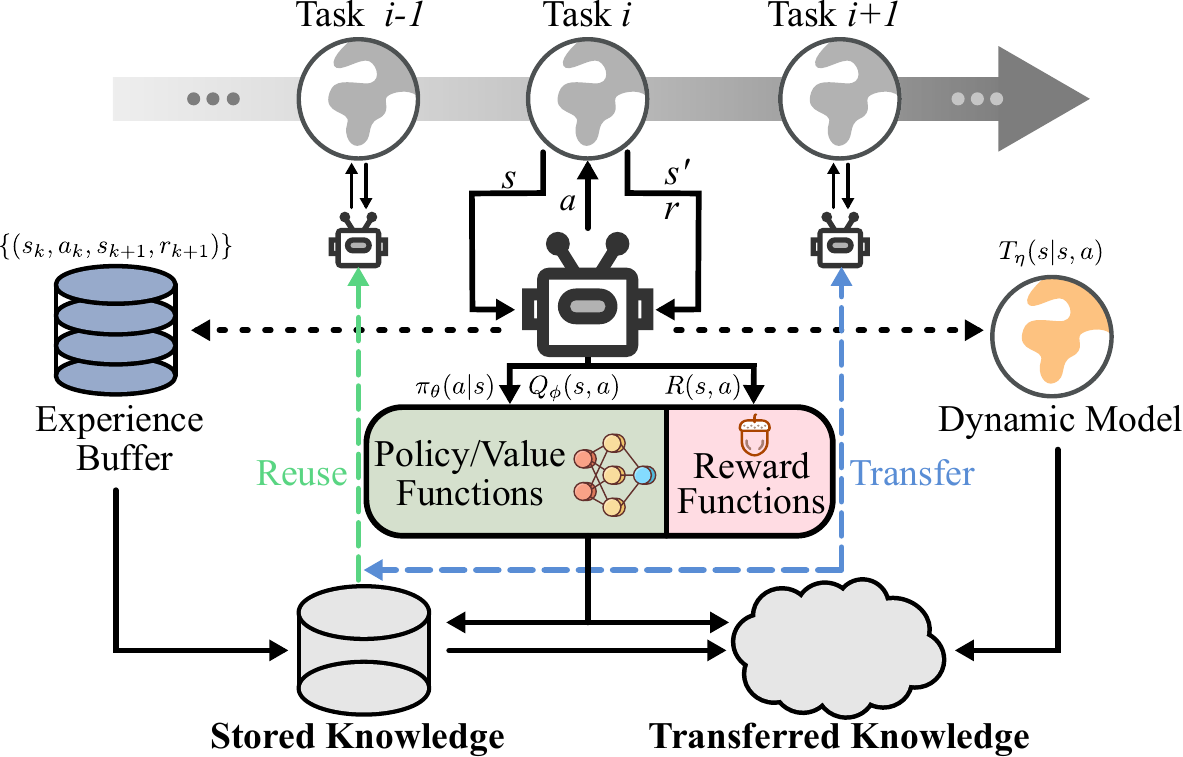}}
    \caption{
        Illustration of the general structure of a CRL method, organized by the knowledge that is stored and/or transferred.
        }
    \label{fig: crl structure}
    \vspace{-0.7cm}
\end{figure}

\subsection{Taxonomy Methodology}\label{sec:taxonomy}
Fig. \ref{fig: crl structure} illustrates the general structure of CRL methods. 
In this framework, an agent's knowledge can be broadly categorized into four main types: \textit{policy}, \textit{experience}, \textit{dynamics}, and \textit{reward}.
While other elements in RL, such as action space and state space, can also be considered forms of knowledge, they are often overlooked in existing CRL methods.
Therefore, our taxonomy primarily focuses on four categories, which are central to the design and implementation of CRL systems.

To systematically organize CRL methods, we address the following key question: \textbf{``What knowledge is stored and/or transferred?''}
Based on this guiding question, we classify CRL methods into four main categories: policy-focused, experience-focused, dynamic-focused, and reward-focused.
We further divide some categories into sub-categories based on how the knowledge is utilized. 
It is important to note that this taxonomy is not exhaustive, and many methods may span multiple categories. 
To facilitate a comprehensive overview of the development of CRL methods, we list representative approaches in Table \ref{tab: methods}, organized chronologically. Before detailing each category, we refer readers to Table~\ref{tab:integration}, which serves as a compact guide linking representative CRL families to their typical scenarios, challenge emphasis, and scalability profiles.

\subsection{Policy-Focused Methods}\label{sec: policy-based}
We start by introducing the policy-focused methods, which are the most common and fundamental in CRL.
This is because the policy function or value function constitutes the core knowledge in RL, directly determining the agent's decision-making process.
Among these methods, the fine-tuning strategy, where the agent inherits the policy or value function from a previous task, is widely adopted as a naive mechanism for knowledge transfer. 
This setup also naturally arises when agents start from large pretrained backbones and must update them continually across task streams \cite{piccoli2025CombiningPretrained}.
This strategy often overlaps with other CRL approaches discussed later \cite{bengio2012deep,gaya2022building}.
The policy-focused methods can be further divided into three sub-categories: \textit{policy reuse, policy decomposition, and policy merging}.
Below, we provide a detailed discussion.

\begin{table*}
    \caption{The taxonomy of continual reinforcement learning methods.}
    \label{tab: methods}

    \centering
    \begin{tabular}{@{}c|c|l|p{0.6\textwidth}<{\centering}@{}}
    \toprule
    \multicolumn{3}{c|}{\textbf{Category}} & \textbf{Methods} \\ \midrule
    \midrule
    \multicolumn{1}{c|}{\multirow{4}{*}{\begin{tabular}[c]{@{}c@{}} \textcolor{red}{$\S~$}\ref{sec: policy-based} \\ Policy-focused \end{tabular}}} &
    \multicolumn{2}{l|}{\textcolor{red}{$\S~$}\ref{sec: policy_reuse}:~Policy Reuse} & \cellcolor{policy!60} MAXQINIT\cite{abel2018policy}, LFRL\cite{liu2019lifelong}, \cite{garcia2019advances}, \cite{grbic2019towards}, \cite{nangue2020boolean}, \cite{tasse2020logical},  CSP\cite{gaya2022building}, SOPGOL\cite{tasse2022generalisation}, ClonEx-SAC\cite{wolczyk2022disentangling}, UCOI\cite{soumia2023value}, SWOKS\cite{dick2024StatisticalContext}, \cite{piccoli2025CombiningPretrained}, LEGION \cite{meng2025preserving} \\ \cmidrule(l){2-4} 
    &
    \multicolumn{2}{l|}{\textcolor{red}{$\S~$}\ref{sec: policy_decomposition}:~Policy Decomposition} & \cellcolor{policy!60} PG-ELLA\cite{ammar2014online}, \cite{ammar2015autonomous}, \cite{ammar2015safe}, TaDeLL\cite{isele2016UsingTask}, PNN\cite{rusu2016progressive}, ePG-ELLA\cite{zhan2017scalabe}, MPHRL\cite{wu2020ModelPrimitives}, LPT-FTW\cite{mendez2020lifelong}, \cite{mowakeaa2021KernelbasedLifelong}, CDLRL-ZPG\cite{qian2021zeroshot}, OWL\cite{kessler2022same}, SANE\cite{powers2022self}, \cite{mendez2022modular}, HLifeRL\cite{ding2022hliferl}, PT-TD\cite{anand2023prediction}, COVERS\cite{liu2023continual},  HVCL\cite{Hihn2023},  RHER\cite{luo2023relay}, CompoNet\cite{malagon2024self},  DaCoRL\cite{zhang2024dynamics}, MacPro\cite{yuan2025MultiagentContinual}\\ \cmidrule(l){2-4} 
    &
    \multicolumn{2}{l|}{\textcolor{red}{$\S~$}\ref{sec: policy_merge}:~Policy Merging} & \cellcolor{policy!60} EWC\cite{james2017overcoming}, PathNet\cite{fernando2017pathnet},  Benna-Fusi\cite{kaplanis2018continual}, P\&C\cite{schwarz2018progress}, Online-EWC\cite{schwarz2018progress}, PC\cite{kaplanis2019policy}, DisCoRL\cite{traore2019discorl},  \cite{traore2019ContinualReinforcement}, VPC\cite{doyle2019VariationalPolicy}, BLIP\cite{shi2021ContinualLearning}, HN-PPO\cite{pfilemon2022hypernetworkppo}, CoTASP\cite{yang2023ContinualTask}, MASK$_\text{BLC}$\cite{beniwhiwhu2023lifelong}, UCBlvd\cite{amani2023provably}, \cite{zhao2024ExperienceConsistency}, RPT\cite{yao2025GeneralRelation}, C-CHAIN\cite{tang2025mitigating}, CR\cite{surdej2025BalancingExpressivity}, VQ-CD\cite{hu2025TacklingContinual}, CKA-RL\cite{hu2025ContinualKnowledge}, CDE\cite{jaziri2025MitigatingStabilityplasticity}, FAME\cite{sun2026principled} \\
     \cmidrule(l){1-4} 
    \multicolumn{1}{c|}{\multirow{3}{*}{\begin{tabular}[c]{@{}c@{}} \textcolor{red}{$\S~$}\ref{sec: experience-focused} \\ Experience-focused \end{tabular}}} &
    \multicolumn{2}{l|}{\textcolor{red}{$\S~$}\ref{sec: direct_replay}:~Direct Replay} & \cellcolor{experience!60} \cite{isele2018selective}, CLEAR\cite{rolnick2019experience}, CoMPS\cite{berseth2022comps}, \cite{zhou2022forgetting}, \cite{xie2022lifelong}, 3RL\cite{caccia2023task}, CPPO\cite{zhang2024cppo}, \cite{xu2024PolicyCorrection}, DAAG \cite{palo2025diffusion}, \cite{zhang2023ReplayenhancedContinual} \\ \cmidrule(l){2-4} 
    &
    \multicolumn{2}{l|}{\textcolor{red}{$\S~$}\ref{sec: generative_experience}:~Generative Replay} & \cellcolor{experience!60} SLER\cite{li2021sler}, RePR\cite{craig2021pseudo}, S-TRIGGER\cite{caselles2021strigger}, \cite{daniels2022modelfree}, \cite{yue2025TDGRTrajectoryBased} \\ \cmidrule(l){1-4} 
    \multicolumn{1}{c|}{\multirow{2}{*}{\begin{tabular}[c]{@{}c@{}} \textcolor{red}{$\S~$}\ref{sec: dynamic-focused} \\ Dynamic-focused \end{tabular}}} &
    \multicolumn{2}{l|}{\textcolor{red}{$\S~$}\ref{sec: direct_modeling}:~Direct Modeling} & \cellcolor{dynamic!60} MOLe\cite{nagabandi2018deep}, \cite{xu2020task}, HyperCRL\cite{huang2021continual}, VBLRL\cite{fu2022modelbased}, LLIRL\cite{wang2022lifelong},   \cite{wang2023dirichlet}, Losse-FTL\cite{liu2024locality}, DRAGO\cite{fu2025knowledge} \\ \cmidrule(l){2-4} 
    &
    \multicolumn{2}{l|}{\textcolor{red}{$\S~$}\ref{sec: indirect_modeling}:~Indirect Modeling} & \cellcolor{dynamic!60} LILAC\cite{xie2020DeepReinforcement},  LiSP\cite{lu2021resetfree}, 3RL\cite{caccia2023task}, Continual-Dreamer\cite{kessler2023effectiveness} \\ \cmidrule(l){1-4} 
    \multicolumn{3}{c|}{\textcolor{red}{$\S~$}\ref{sec: reward-focused}:~Reward-focused} & \cellcolor{reward!60} ELIRL\cite{mendez2018lifelong}, \cite{jiang2021temporal}, IML\cite{bougie2021intrinsically}, SR-LLRL\cite{kun2021accelerating},  LSRM\cite{zheng2022lifelong}, \cite{steinparz2022reactive}, MT-Core\cite{pan2025multigranularity} \\ 
    \bottomrule
    \end{tabular}
    \vspace{-0.5cm}
\end{table*}

\subsubsection{Policy Reuse}\label{sec: policy_reuse}
\begin{figure}
    \centerline{\includegraphics[width=.6\linewidth]{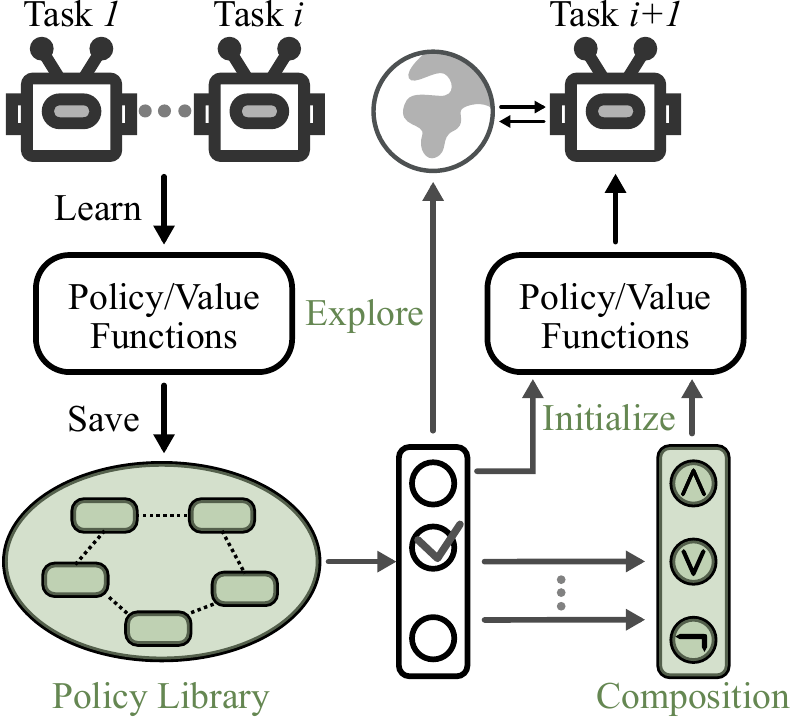}}
    \caption{
        The framework of policy reuse in CRL methods.
        Stored policies are reused to initialize new policies, enhance exploration, and improve generalization by leveraging task composition frameworks.
        }
    \label{fig: policy-reuse}
    \vspace{-0.5cm}
\end{figure}

Policy reuse is a widely used strategy in CRL, where the agent retains and reuses complete policies from previous tasks. 
As illustrated in Fig. \ref{fig: policy-reuse}, the simplest approach involves storing all previously learned policies to prevent catastrophic forgetting and using them as a foundation for developing new policies.
While most methods of policy reuse have limited scalability, it remains a common approach for implementing CRL agents, as it primarily focuses on knowledge transfer and adaptation.

One key form of policy reuse involves \textbf{initializing} new policies with prior knowledge before fine-tuning. 
To evaluate initial target-task performance following knowledge transfer, the ``\textit{jumpstart}'' metric was introduced \cite{abel2018policy}. 
The jumpstart objective \cite{traore2019discorl,abel2024definition,anand2023prediction} seeks to maximize the expected value function at the initial state:
\begin{equation}
    J(\pi) = \argmax_{\pi} \mathbb{E}_{M\in\mathcal{M}} [V^{\pi}_M(s_0)],
    \label{eq:jumpstart}
\end{equation}
where $V^{\pi}_M(s_0)$ is the value function of policy $\pi$ at initial state $s_0$ for task $M$ in the task set $\mathcal{M}$.
Building on this, MAXQINIT optimally initializes the $Q$-function using maximum estimated $Q$-values from the empirical task distribution \cite{abel2018policy}:
\begin{equation}
    \hat{Q}_{max}(s, a) = \max_{M \in \mathcal{\hat{M}}} Q_M(s, a),
    \label{eq:maxqinit}
\end{equation}
with $\mathcal{\hat{M}}$ denoting sampled tasks and $Q_M$ the learned task-specific $Q$-values.
Additionally, Mehimeh \etal \cite{soumia2023value} proposed \textit{Uncertainty and Confidence aware Optimistic Initialization} (UCOI). UCOI selectively applies optimistic initialization based on state-action uncertainty (derived from past outcome variability) and PAC-MDP confidence bounds, thereby minimizing unnecessary exploration and improving efficiency.

Policy reuse can also improve \textbf{exploration} by leveraging past policies. 
Evaluating various exploration strategies for SAC, Wolczyk \etal \cite{wolczyk2022disentangling} proposed ClonEx-SAC, which utilizes the best-return past policy to excel in knowledge transfer, especially for recurring tasks. 
Furthermore, Meta-MDP \cite{garcia2019advances} formulates the search for an optimal exploration strategy as a meta-level MDP. 
By decoupling exploration from exploitation, it enables cross-task optimization of exploratory behavior, which boosts efficiency despite the difficulty of determining optimal exploration durations.

While effective, these methods often depend heavily on task similarity for generalization. 
To overcome this, some CRL approaches achieve zero-shot generalization via \textbf{task composition} frameworks like Boolean algebra and logical composition \cite{nangue2020boolean,tasse2020logical,tasse2022generalisation}, enabling the direct reuse of composed policies for novel tasks without retraining. 

Task composition via Boolean algebra facilitates combining tasks through negation ($\neg$), disjunction ($\lor$), and conjunction ($\land$) over a task set $\mathcal{M}$ \cite{nangue2020boolean}. 
Specifically:
\begin{enumerate}
    \item Negation $\neg$ yields a task with reward $R_{\neg M}(s,a) = (R_{M_{\text{MAX}}}(s,a) + R_{M_{\text{MIN}}}(s,a)) - R_M(s,a)$, where $R_{M_{\text{MAX}}}(s,a) = \max_{M \in \mathcal{M}} R_M(s,a)$ and $R_{M_{\text{MIN}}}(s,a) = \min_{M \in \mathcal{M}} R_M(s,a)$. 
    \item Disjunction $\lor$ merges $M_1$ and $M_2$ via $R_{M_1 \lor M_2}(s,a) = \max\{R_{M_1}(s,a), R_{M_2}(s,a)\}$.
    \item Conjunction $\land$ combines them via $R_{M_1 \land M_2}(s,a) = \min\{R_{M_1}(s,a), R_{M_2}(s,a)\}$.
\end{enumerate}

For goal-based settings, this framework extends to computing optimal value functions for composed tasks directly. 
The extended $Q$-value function $\bar{Q}$ is defined as:
\begin{equation}
    \bar{Q}(s,g,a) = \bar{R}(s,g,a) + \gamma \sum_{s'} P(s'|s,a) \bar{V}^{\bar{\pi}}(s',g),
    \label{eq:extended_q}
\end{equation}
where $g$ is a goal, $\bar{R}$ penalizes undesired goals, and $\bar{V}^{\bar{\pi}}$ is the value under $\bar{\pi}$. 
Logical operations apply analogously to these functions \cite{tasse2022generalisation}:
\begin{equation}
    \begin{aligned}
        & \neg(\bar{Q}^{\star})(s,g,a) \\ &= (\bar{Q}_{M_{\text{MIN}}}^{\star}(s,g,a) + \bar{Q}_{M_{\text{MAX}}}^{\star}(s,g,a)) - \bar{Q}^{\star}(s,g,a), \\
        & \lor(\bar{Q}_1^{\star}, \bar{Q}_2^{\star})(s,g,a) = \max\{\bar{Q}_1^{\star}(s,g,a), \bar{Q}_2^{\star}(s,g,a)\}, \\
        & \land(\bar{Q}_1^{\star}, \bar{Q}_2^{\star})(s,g,a) = \min\{\bar{Q}_1^{\star}(s,g,a), \bar{Q}_2^{\star}(s,g,a)\}.
        \label{eq:logical_q}
    \end{aligned}
\end{equation}
This zero-shot composition lets agents immediately solve novel tasks using prior skills. 
Furthermore, \textit{Sum Of Products with Goal-Oriented Learning} (SOPGOL) accommodates stochastic and discounted tasks, helping agents decide between reusing skills or learning anew \cite{tasse2020logical}, significantly boosting sample efficiency \cite{nangue2020boolean,tasse2020logical}.

\begin{figure}
    \centerline{\includegraphics[width=.98\linewidth]{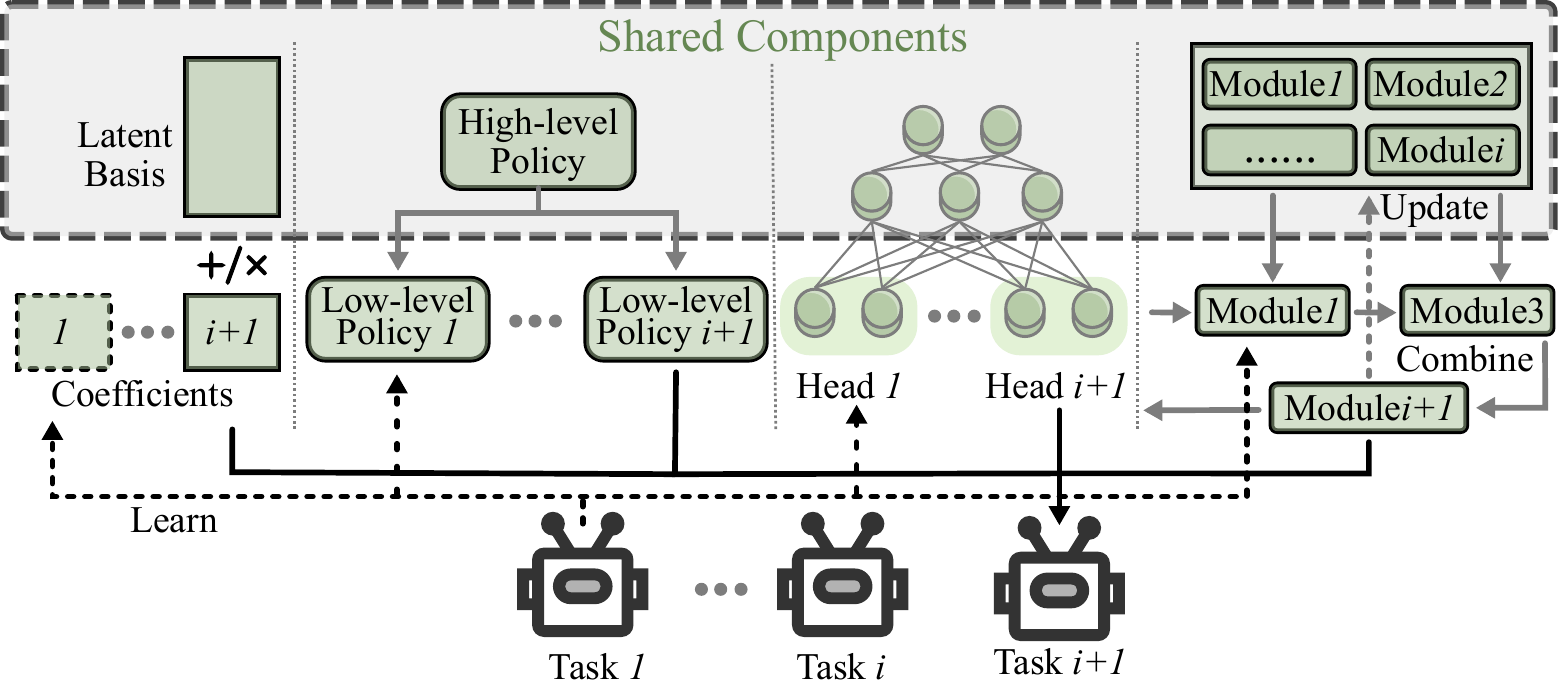}}
    \caption{
        The framework of policy decomposition in CRL methods.
        Factor decomposition, multi-head network, hierarchical decomposition, and modular architecture are used to decompose the policy into a shared base and task-specific components.
        }
    \label{fig: policy-decomposition}
    \vspace{-0.7cm}
\end{figure}

To scale policy reuse, \textit{Continual Subspace of Policies} (CSP) maintains a policy \textbf{subspace} rather than discrete models \cite{gaya2022building}.
Represented as a convex hull in parameter space, each vertex (anchor) encapsulates a policy. 
New policies emerge as convex combinations of these anchors, while novel learning adds new vertices, yielding sublinear model growth and a better performance-scalability tradeoff.

\subsubsection{Policy Decomposition}\label{sec: policy_decomposition}
Policy decomposition is another widely used strategy in CRL, where the agent decomposes the policy into multiple components and reuses them in various ways.
The primary challenge in policy decomposition lies in determining how to effectively decompose the policy.
As shown in Fig. \ref{fig: policy-decomposition}, this can be achieved through four main approaches: \textit{factor decomposition}, \textit{multi-head decomposition}, \textit{modular decomposition}, and \textit{hierarchical decomposition}.

\textbf{Factor decomposition} is mainly used in early CRL methods without deep learning, where the policy is decomposed into a shared base and task-specific components.
This approach is is derived from multi-task supervised learning, and has been successfully applied in CRL by \textit{Policy Gradient Efficient Lifelong Learning Algorithm} (PG-ELLA) \cite{ammar2014online,zhan2017scalabe}.
PG-ELLA introduces a latent basis representation to model each task's parameters as a linear combination of components from a shared knowledge base.
Specifically, the policy parameters for task $k$ are represented as:
\begin{equation}
    \bm{\theta}_{k} = \bm{L} \bm{s}_{k},
    \label{eq:pg-ella}
\end{equation}
where $\bm{L}$ is the shared latent basis and $\bm{s}_{k}$ are the task-specific coefficients. 
However, PG-ELLA trains individual policies first, which may lead to incompatibility with the shared base.
To address this, \textit{Lifelong PG: Faster Training Without forgetting} (LPG-FTW) \cite{mendez2020lifelong} optimizes task-specific coefficients directly using policy gradients, ensuring compatibility with the shared base while leveraging shared knowledge to accelerate learning.
This modification enables LPG-FTW to handle more complex dynamical systems. 

Building on the foundations of PG-ELLA, subsequent research has improved the efficiency and scalability of the factor decomposition method using kernel methods \cite{mowakeaa2021KernelbasedLifelong} or the multiple processing units assumption \cite{zhan2017scalabe}.
Furthermore, the introduction of cross-domain lifelong RL frameworks \cite{ammar2015autonomous} enables agents to efficiently learn and generalize across multiple task domains.
This is achieved by partitioning the series of tasks into task groups, such that all tasks within a particular group $\mathcal{G}$ share a common state and action space.
Then, the policy parameters for task $k$ in group $\mathcal{G}$ are represented as:
\begin{equation}
    \bm{\theta}_{k} = \bm{\Psi}^{(\mathcal{G})} \bm{L} \bm{s}_{k},
    \label{eq:cross-domain}
\end{equation}
where $\bm{\Psi}^{(\mathcal{G})}$ is a group-specific projection matrix that maps the shared latent basis $\bm{L}$ to the task-specific coefficients $\bm{s}_{k}$.

Further advancements include the integration of task descriptors for zero-shot knowledge transfer.
\textit{Task Descriptors for Lifelong Learning} (TaDeLL) assumes that task descriptors $\phi(\mathbf{m}_{k})$ can be linearly factorized using a latent basis $\mathbf{D}$ over the descriptor space, coupled with the policy basis $\mathbf{L}$ to share the same coefficient vectors $\mathbf{s}_{k}$ \cite{isele2016UsingTask}:
\begin{equation}
    \phi(\mathbf{m}_{k}) = \mathbf{D} \mathbf{s}_{k}.
    \label{eq:tadell}
\end{equation}
This allows for consistent task embeddings across policies and descriptors, enhancing learning efficiency and enabling zero-shot transfer.
\textit{Cross-Domain Lifelong Reinforcement Learning algorithm with Zero-shot Policy Generation ability} (CDLRL-ZPG) \cite{qian2021zeroshot} further extends the zero-shot ability by constructing a linear mapping from environmental coefficients $\mathbf{q}^{\mathcal{D}}_{k}$ to task-specific coefficients $\mathbf{s}^{\mathcal{D}}_{k}$ using a matrix $W^{\mathcal{D}}$ in learned task domain $\mathcal{D}$:
\begin{equation}
    \mathbf{s}^{\mathcal{D}}_{k} = W^{\mathcal{D}} \mathbf{q}^{\mathcal{D}}_{k}.
    \label{eq:cdlrl-zpg}
\end{equation}
This mapping allows the generation of approximate optimal policy parameters for new tasks directly from environmental information, significantly improving generalization across different task domains without additional learning.

Finally, although PT-TD learning is not a factor decomposition method, it also uses a similar idea to decompose the value function, and it is also agnostic to the nature of the function approximator used \cite{anand2023prediction}.
PT-TD learning decomposes the value function into two components that update at different timescales: a permanent value function for preserving general knowledge and a transient value function for learning task-specific knowledge.
Then the overall value function is:
\begin{equation}
    V^{\rm(PT)}(s) = V_\theta^{\rm(P)}(s) + V_\phi^{\rm(T)}(s),
    \label{eq:pt-td}
\end{equation}
where $\theta$ and $\phi$ are the parameters of the permanent function $V^{\rm(P)}(s)$ and transient value function $V^{\rm(T)}(s)$, respectively.
The parameters of the transient value function are updated by TD learning during learning on tasks, while the parameters of the permanent value function are updated using all stored states of the task after learning.

Due to the advances in DRL and the increasing complexity of tasks, many CRL methods have evolved to incorporate deep neural networks.
Policy and value function networks can be decomposed into multiple parts, such as multiple heads and multiple modules.
By combining these parts, agents can learn and generalize across tasks more effectively, enhancing scalability and performance in complex environments.

\textbf{Multi-head decomposition} is a common strategy in multi-task learning, where the network consists of a shared backbone and multiple heads, each responsible for a different task.
In CRL, Wolczyk \etal \cite{wolczyk2022disentangling} empirically investigated the impact of multi-head networks on the continual learning performance of SAC.
By assigning separate output heads for each task, the agent facilitates the transfer of knowledge across tasks.
However, freezing the backbone of the critic can hinder forward knowledge transfer, which is against the understanding of transfer in supervised learning.
Additionally, the agent's performance can benefit from the resetting of the head for the critic while being damaged by the resetting of the head for the actor.

Furthermore, \textit{cOntinual RL Without confLict} (OWL) finds that a multi-head network is suitable for dealing with the problem of interference, in which tasks have different goals (reward functions) \cite{kessler2022same}.
In these cases, tasks may be fundamentally incompatible with each other and thus cannot be learned by a single policy.
By using dedicated heads for each task, OWL allows the network to learn task-specific policies without overwriting previously acquired knowledge.
Furthermore, OWL extends the multi-head network to the sequence with unknown task identifiers by modeling the head selection as a multi-armed bandit problem. 

Expanding the application of the multi-head network to dynamic and non-stationary environments, \textit{Dynamics-adaptive Continual RL} (DaCoRL) incorporates a context-conditioned multi-head design to detect and adapt to environmental changes \cite{zhang2024dynamics}.
Each head corresponds to a specific context, defined by a set of tasks with similar dynamics, allowing the network to specialize in context-specific policies.
The framework dynamically expands its architecture by adding new heads when novel contexts are detected, ensuring scalability and adaptability to previously unseen scenarios.

Multi-head decomposition divides a policy or value network into two components, which may not fully address the complexity of relationships among tasks.
A more granular partitioning strategy will enable finer control over the transfer and retention of knowledge.
\textbf{Modular decomposition} is an efficiency strategy in MTL \cite{meyerson2018beyond,chang2019automatically,pfeiffer2023modular}, which leverages the composition of specialized modular deep architectures to capture compositional structures that arise in complex tasks.
It has similarities with the inner workings of the human brain and has been provides evidence of their biological plausibility \cite{stocco2010conditional,kell2018task}.
\textit{Progressive Neural Networks} (PNN) has made early explorations in this direction \cite{rusu2016progressive}, although it does not introduce the concept of modularity.
PNN trains a new column of network parameters for each task, while lateral connections are established between corresponding layers of all previously trained columns.
This design achieves strong forward transfer without overwriting previously learned information. 
However, the scalability of PNNs is a notable limitation, as the number of parameters grows quadratically with the number of tasks \cite{malagon2024self}.
The subsequent methods achieve better scalability through explicit modularity.

An early effort in lifelong learning introduced a framework with a two-stage learning process, separating the reuse of existing knowledge (assimilation) from the improvement of old components or creation of new components (accommodation) \cite{mendez2021lifelong}. 
It emphasizes compositionality, where tasks are represented as combinations of reusable components. 
Mathematically, if a task $k$ can be solved by reusing existing modules $\{m_i\}_{i=1}^n$ from a module set $\bm{M}$, the solution function $f_{k}$ can be represented as a composition of these modules:
\begin{equation}
f_{k}(s) = m_{1} \circ m_{2} \circ \cdots \circ m_{k}(s), \quad m_i \in \bm{M},
\label{eq:modular}
\end{equation}
where $\circ$ denotes the compositional operation (\eg functional composition or linear combination).
Then, they extended this idea to CRL by formalizing lifelong compositional RL problems as a compositional problem graph, where each node represents a module to solve the corresponding subproblem \cite{mendez2022modular}.
Each module is a small neural network that takes as input the module-specific state component along with the output of the module of the previous module.
The goal of a task is to find a path between nodes corresponding to a policy that maximizes the expected return.
Although this method accurately captures the relations across tasks, it requires attempting all possible combinations of modules, which has low scalability.

Subsequent works have advanced the modular paradigm by focusing on dynamic and autonomous module management.
\textit{Self-Activating Neural Ensembles} (SANE) introduced a task-id-free method that automatically detects and responds to drift in the setting by maintaining an ensemble of modules \cite{powers2022self}.
It does this by activating, merging, and creating modules automatically.
Each module $m_i$ contains an actor and a critic.
It is associated with an activation score $u_i(s)$, which determines its relevance to the current state $s$:
\begin{equation}
u_i(s) = |G_t - V_i(s)|, 
\label{eq:sane}
\end{equation}
where $V_i(s)$ is the value estimate of module $i$ at state $s$ and return $G_t$.
The module with the highest activation score is selected for inference.
The \textit{Upper Confidence Bound} (UCB) for each module is used to balance exploration and exploitation:
\begin{equation}
V_i^{\rm{UCB}}(s) = V_i(s) + \alpha_u \cdot u_i(s),
\label{eq:sane_ucb}
\end{equation}
where $\alpha_u$ is a hyperparameter representing how wide a margin around the expected value to allow.
The activated module is:
\begin{equation}
m_{\text{a}} = \arg\max_{i} V_i^{\rm{UCB}}(s).
\label{eq:sane_activation}
\end{equation}
SANE was later adapted for home robotics as SANER, which tailored the modular framework to low-data settings using attention-based interaction policies \cite{powers2023evaluating}. 

To further improve the scalability of the modular framework and avoid interference, \textit{self-Composing policies Network architecture} (CompoNet) introduced attention mechanisms for selective knowledge transfer \cite{malagon2024self}.
Each task corresponds to a module, and the modules of multiple tasks form a cascaded structure.
Each module can access and compose the outputs of the previous modules by two attention heads and an internal policy to solve the current task.
By allowing policies to compose themselves autonomously, CompoNet significantly reduces the memory and computational cost and ensures linear growth of parameters with the number of tasks.

Inspired by the work related to hierarchical RL \cite{pateria2021hierarchical,barto2003recent}, \textbf{hierarchical decomposition} is a prominent approach in policy-focused CRL that leverages the natural structure of tasks to organize policies into a hierarchy of reusable components.
This approach is particularly effective in addressing complex tasks with multiple steps, as it allows agents to decompose tasks into simpler sub-policies or skills that can be reused across tasks. 
In hierarchical decomposition, the policy is typically structured into high-level controllers and low-level sub-policies, enabling efficient task execution and scalability in multi-task or lifelong learning settings. 

A variety of approaches have been proposed to implement hierarchical decomposition in CRL, each emphasizing different mechanisms for skill discovery, knowledge storage, and transfer. 
For instance, \textit{Hierarchical Deep Reinforcement Learning Network} (H-DRLN) integrates reusable \textit{Deep Skill Networks} (DSNs) into a hierarchical framework \cite{tessler2017DeepHierarchical}. 
This approach enables the agent to decompose tasks into reusable skills, reducing sample complexity and improving performance in high-dimensional environments like Minecraft. 
Similarly, the \textit{Hierarchical Lifelong Reinforcement Learning framework} (HLifeRL) employs an option framework to automatically discover and store low-level skills in an option library \cite{ding2022hliferl}. 
The master policy in HLifeRL selects these options to execute tasks, allowing for efficient skill reuse and combating catastrophic forgetting. 
Another notable method is the \textit{Model Primitive Hierarchical Reinforcement Learning} (MPHRL) framework, which utilizes model primitives (suboptimal world models) to decompose tasks into modular sub-policies \cite{wu2020ModelPrimitives}. 
This bottom-up approach enables the agent to learn sub-policies and a gating controller concurrently, enhancing knowledge transfer and scalability.
HLifeRL and MPHRL both show that the cost of learning the first task is amortized over subsequent tasks, resulting in substantial long-term gains.
Additionally, Hihn \etal \cite{Hihn2023} introduced a hierarchical information-theoretic optimality principle with the \textit{Hierarchical Variational Continual Learning} (HVCL) framework. 
It uses a \textit{Mixture-of-Variational-Experts} (MoVE) layer to create multiple information processing paths governed by a gating policy, facilitating specialized learning and mitigating forgetting without requiring task-specific knowledge.

These methods share commonalities in their hierarchical structuring of knowledge but differ in their specific techniques for skill discovery and reuse. 
H-DRLN relies on skill distillation to encapsulate multiple skills into a single network, while HLifeRL uses an explicit option framework to separate high-level decision-making from low-level execution. 
On the other hand, MPHRL emphasizes the use of model primitives to guide task decomposition, with a probabilistic gating mechanism to activate the appropriate sub-policy. 
HVCL introduces diversity objectives to enhance expert allocation and uses Wasserstein distance as a kernel for measuring expert diversity, ensuring distinct expert parameters are maintained.

However, hierarchical decomposition methods face several challenges that remain open for future research. 
One key issue is the increasing complexity of the action space as the number of tasks and skills grows, which can strain the scalability of the hierarchical framework \cite{ding2022hliferl}. 
Additionally, the performance of these methods often depends on the quality and diversity of the discovered skills or model primitives \cite{wu2020ModelPrimitives}. 
Future work could explore more adaptive mechanisms for dynamic skill discovery and online refinement, as well as strategies to manage the complexity of growing skill libraries. 

\begin{figure}
    \centerline{\includegraphics[width=.9\linewidth]{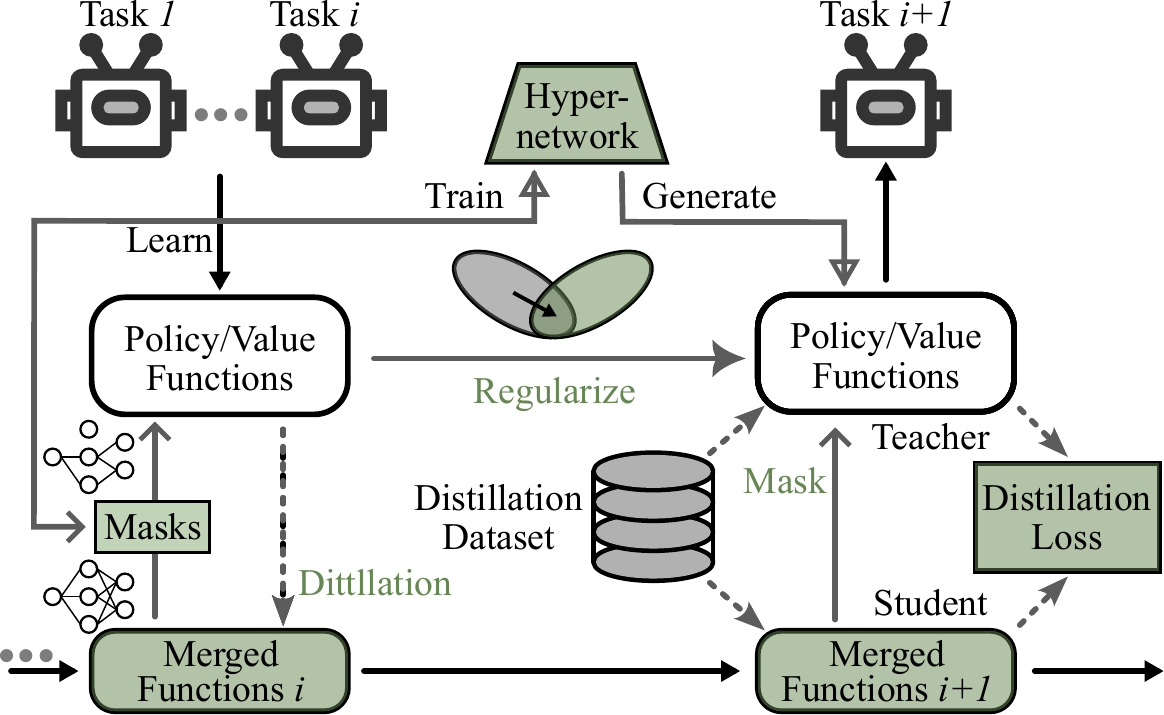}}
    \caption{
        The framework of policy merging in CRL methods.
        Distillation, hypernetworks, masks, and regularization are used to merge multiple policies into a single policy.
        }
    \label{fig: policy-merge}
    \vspace{-0.7cm}
\end{figure}

\subsubsection{Policy Merging}\label{sec: policy_merge}
Policy merging is a storage-sensitive strategy in CRL that focuses on merging the model of policies from multiple tasks into a single model, rather than retaining individual models for each task.
This approach is particularly useful in scenarios where memory constraints are a concern, as it allows agents to compress knowledge from multiple tasks into a more compact representation.
By merging policies, agents can reduce the memory footprint and computational cost of storing and executing multiple policies.
As illustrated in Fig. \ref{fig: policy-merge}, these methods typically involve distillation, hypernetworks, masks, or regularization to combine policies.

\textbf{Distillation} is a common technique in supervised learning that has been adapted for CRL to merge policies and facilitate knowledge retention.
This technique usually involves training a student policy on a new task to mimic the output of a teacher policy learned from previous tasks, effectively transferring knowledge from the old policies to the new ones.
The \textit{Progress \& Compress} (P\&C) framework \cite{schwarz2018progress} exemplifies this approach by integrating a knowledge base and an active column to sequentially learn tasks. 
After training on a new task, the active column's policy is distilled into the knowledge base using a cross-entropy loss. 
The framework also employs a modified version of \textit{Elastic Weight Consolidation} (EWC) \cite{james2017overcoming} to safeguard against catastrophic forgetting during the distillation process.
Similarly, \textit{Distillation for Continual Reinforcement Learning} (DisCoRL) \cite{traore2019discorl} employs policy distillation to merge multiple task-specific policies into a single student policy. 
By generating distillation datasets from each task and using Kullback-Leibler divergence with temperature smoothing as a loss function, it ensures effective knowledge transfer while eliminating the need for explicit task indicators at test time. 

Recent advancements have introduced novel techniques to enhance the distillation process. 
For example, an experience consistency distillation method for robotic manipulation tasks \cite{zhao2024ExperienceConsistency} combines policy distillation with experience distillation, leveraging \textit{Fréchet Inception Distance} (FID) loss to maintain distribution consistency between original and distilled experiences. 
This approach not only mitigates forgetting but also optimizes memory usage by compressing experiences into a compact representation, which is then replayed during training. 
Furthermore, \textit{UCB lifelong value distillation} (UCBlvd) \cite{amani2023provably} incorporates theoretical guarantees into the distillation process, ensuring sublinear regret and computational efficiency in CRL. 
By leveraging linear representations and \textit{Quadratically Constrained Quadratic Programming} (QCQP), UCBlvd minimizes computational complexity while achieving effective merging.

These advancements also collectively highlight a trend toward leveraging distillation not only as a tool for policy merging but also as a means to improve data efficiency and scalability in CRL \cite{tessler2017DeepHierarchical,craig2021pseudo,zhou2022forgetting}. 
Relatedly, knowledge-centric retention objectives have been explored to preserve transferable representations across tasks \cite{fu2025knowledge}.
Despite their successes, challenges remain, such as optimizing the distillation process for diverse task distributions and ensuring the scalability to a larger number of tasks or more complex environments.

\textbf{Hypernetworks} are neural networks that generate the weights of another neural network, allowing for the dynamic generation of task-specific policies \cite{chauhan2024brief}.
The use of hypernetworks in CL has shown promise in merging policies while maintaining task-specific adaptability \cite{Oswald2020Continual}.
In CRL, \textit{HyperNetwork-based implementation of PPO} (HN-PPO) \cite{pfilemon2022hypernetworkppo} employs a hypernetwork to generate policy weights conditioned on task embeddings, enabling the agent to adapt to new tasks without discarding previously acquired knowledge. 
By regularizing the output of the hypernetwork, the method mitigates catastrophic forgetting and ensures stability across tasks. 
Similarly, Xu \etal \cite{xupolicy2024} integrated a hypernetwork for state-conditioned action evaluation, which dynamically generates evaluators to adapt policies based on current states. 
This not only facilitates knowledge transfer but also enhances few-shot generalization.
These methods demonstrate the versatility of hypernetworks in dynamically encoding task-specific policies within a shared architecture, reducing memory overhead while ensuring efficient knowledge reuse.
Recent work also studies how to balance model expressivity with robustness when shared architectures must represent diverse tasks under continual drift \cite{surdej2025BalancingExpressivity}.
However, the computational complexity of hypernetwork training and the need for a refined training pipeline remain challenges for future research.

\textbf{Masks} offer another compelling avenue for policy merging by leveraging task-specific modulating masks to isolate and reuse knowledge. 
While the application of masks has been tested extensively in CSL for classification \cite{mallya2018packnet,wortsman2020supermasks}, very little is known about their effectiveness in CRL \cite{fernando2017pathnet}.
One possible reason is that the previous mask methods lack the ability to transfer knowledge \cite{beniwhiwhu2023lifelong}.
In order to address this limitation, a recent study has explored combinations of previously learned masks to exploit previous knowledge when learning new tasks \cite{beniwhiwhu2023lifelong}.
It introduces a fixed backbone network modulated by learned binary masks that selectively activate relevant parts of the network for each task. 
This approach not only preserves knowledge from prior tasks but also facilitates forward transfer by combining previously learned masks through linear or balanced linear compositions. 
Extending this method, the distributed system \textit{Lifelong Learning Distributed Decentralized Collective} (L2D2-C) \cite{nath2023sharing} enables agents to share task-specific masks in a decentralized manner, enhancing collective learning while maintaining robustness to connection drops. 
The results demonstrate the advantages of masks in continual multi-agent RL, which is a promising but underexplored area.
More broadly, activation/routing-style mechanisms are being adapted to offline continual RL settings, where only logged interaction data are available \cite{hu2025TacklingContinual}.

\textbf{Regularization} is another effective strategy for policy merging, as it allows agents to retain knowledge from previous tasks while adapting to new ones.
In CSL, regularization methods have been widely used to prevent catastrophic forgetting.
They do so by penalizing changes in the parameters that are important for previous tasks.
Although this method is often treated as a single category in many reviews \cite{masana2022class,wang2024comprehensive,wang2025comprehensive}, we consider it as a subcategory of policy merging because it usually combines with other methods \cite{ammar2015safe,pfilemon2022hypernetworkppo,zhang2024dynamics}.
Furthermore, many regularization methods in CSL can be directly applied to CRL.
The most common regularization method is EWC \cite{james2017overcoming}, which has been successfully applied in CRL \cite{lomonaco2020continual,wang2020LearningNavigate}.
EWC employs the Fisher information matrix to identify and constrain critical parameters, effectively merging knowledge from sequential tasks without significant loss of previously acquired knowledge.
This method introduces a regularization term that penalizes changes in the parameters that are important for previous tasks, thereby preserving knowledge while allowing the model to adapt to new tasks.
Formally, the EWC is represented by the loss function:
\begin{equation}
    \mathcal{L}_{\text{EWC}} = \sum_{i} \frac{\lambda}{2} \bm{F}_i (\theta_i - \theta_i^{\star})^2,
    \label{eq:ewc}
\end{equation}
where $\bm{F}_i$ is the Fisher information matrix for parameter $\theta_i$, $\theta_i^{\star}$ is the parameter value after training on the previous task, and $\lambda$ is the regularization strength.

Building on EWC, P\&C introduces an online version of EWC to address the computational challenges associated with traditional EWC's linear growth in complexity \cite{schwarz2018progress}.
The online-EWC method modifies the EWC approach by updating the Fisher information matrix incrementally, allowing for efficient knowledge preservation while enabling the model to gracefully forget older tasks if needed.
Instead of recalculating the Fisher information matrix for each task, online-EWC updates it incrementally. The update rule incorporates a decay factor $\gamma$, which gradually reduces the influence of older tasks:
\begin{equation}
    \bm{F}_i^{(t)} = \gamma \bm{F}_i^{(t-1)} + \bm{F}_i^{\text{new}},
    \label{eq:online-ewc}
\end{equation}
where $\bm{F}_i^{(t)}$ is the Fisher information matrix at time $t$, $\bm{F}_i^{(t-1)}$ is the matrix from the previous time step, and $\bm{F}_i^{\text{new}}$ is the matrix calculated for the current task.
Then, the loss function incorporates the updated Fisher information matrix:
\begin{equation}
    \mathcal{L}_{\text{Online-EWC}} = \sum_{i} \frac{\lambda}{2} \bm{F}_i^{(t)} (\theta_i - \theta_i^*)^2.
    \label{eq:online-ewc-loss}
\end{equation}

In terms of broader regularization, Kaplanis \etal \cite{kaplanis2018continual,kaplanis2019policy} proposed models that leverage multiple timescales of learning. 
Their earlier work \cite{kaplanis2018continual} introduces a synaptic model inspired by biological synapses, which incorporates dynamic variables to mitigate catastrophic forgetting without relying on task boundaries. 
The model's ability to retain information is encapsulated in the dynamics of the hidden variables, which help to regularize the learning process. 
Following this, the \textit{Policy Consolidation} (PC) model builds on these ideas by using a cascade of hidden networks to regularize the current policy based on its historical performance, thereby preventing performance degradation in non-stationary environments \cite{kaplanis2019policy}.
In addition to these foundational methods, recent advancements such as \textit{adapTive RegularizAtion in Continual environments} (TRAC) and \textit{Composing Value Functions} (CFV) highlight the evolving landscape of regularization in CRL. 
TRAC, a parameter-free optimizer, dynamically adjusts regularization to prevent the loss of plasticity while enabling rapid adaptation to new tasks \cite{muppidi2024fast}. 
Beyond these baselines, recent CRL studies propose targeted objectives to mitigate the stability--plasticity tension via more explicit knowledge alignment and consolidation principles \cite{hu2025ContinualKnowledge,tang2025mitigating,jaziri2025MitigatingStabilityplasticity,sun2026principled}. 
CFV, on the other hand, focuses on the theoretical underpinnings of value function composition, providing a framework for leveraging entropy-regularization to solve new tasks without further learning \cite{van2019composing}. 

Overall, policy merging in CRL, particularly through regularization methods like EWC and its variants, provides powerful mechanisms for stability. 
Therefore, native regularization methods are still used as baselines for comparison in many CRL studies \cite{gaya2022building,kessler2023effectiveness}. 
The combination of these techniques with transfer learning methods also provides a combinatorial direction for CRL.

\textbf{Scalability considerations.} As summarized in Table~\ref{tab:integration}, the main bottleneck of policy-focused methods is the overhead of storing, selecting, or merging policies as the task stream grows. Their \textit{memory growth} is often driven by extra policy copies, heads, modules, or consolidation statistics, while \textit{compute growth} increases with routing, distillation, or multi-policy optimization. Accordingly, their \textit{task-scaling} is strongest in reusable or task-structured settings where policies can be composed, specialized, or merged rather than relearned from scratch. These methods therefore fit best when tasks admit recurring structure, explicit reuse, or clear decomposition, but they become less attractive when policy storage and merging overhead dominate the continual budget.

\subsection{Experience-Focused Methods}\label{sec: experience-focused}
\begin{figure}[htbp]
    \centerline{\includegraphics[width=0.9\linewidth]{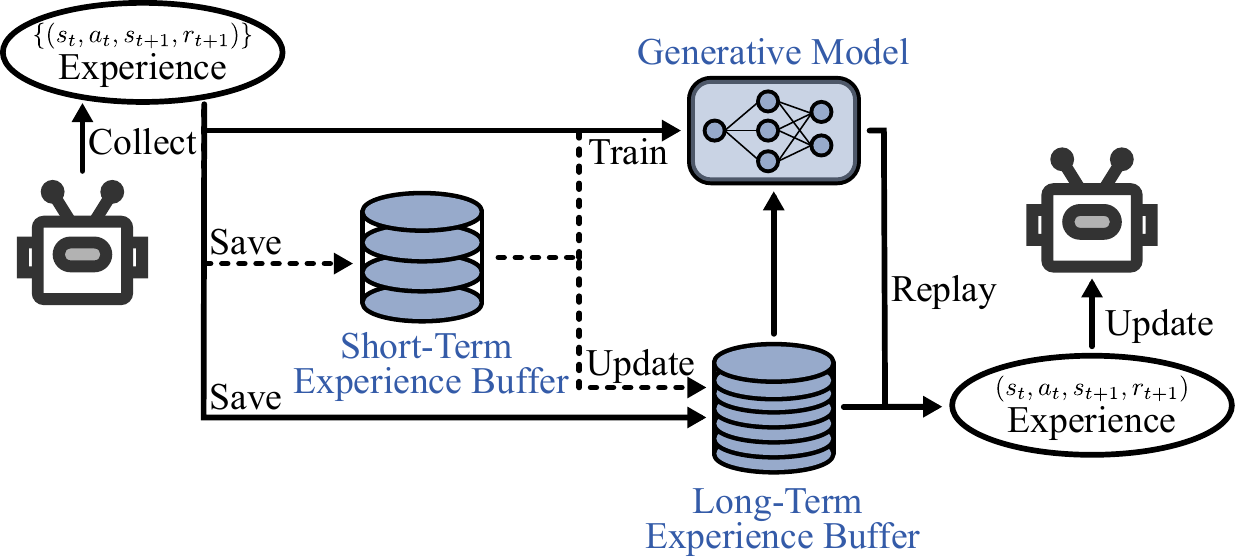}}
    \caption{
        The framework of experience-focused methods. 
        Some methods use a complementary learning system, including a cross-task long-term experience buffer and a short-term experience buffer for the current task \cite{isele2018selective,caccia2023task,li2021sler,craig2021pseudo}.
        Typical subcategories include direct replay and generative replay.
        }
    \label{fig: experience-focused}
    \vspace{-0.2cm}
\end{figure}

Experience-focused methods in CRL aim to enhance the agent's ability to store and reuse experiences effectively.
These methods are similar to the experience replay mechanism widely used in DRL, where experiences are stored in a buffer and replayed to stabilize training and break the correlation in data.
In CRL, experience-focused methods leverage replay buffers, memory mechanisms, or experience relabeling to maintain a balance between retaining critical information from previous tasks and integrating new knowledge.
Experience-focused methods are particularly valuable in CRL, as they provide a direct mechanism for revisiting past knowledge without requiring task-specific information, making them versatile for both task-aware and task-agnostic scenarios.
Based on whether the experience is stored or generated, these methods can be further divided into direct replay and generative replay.
Replay can also be strengthened by explicitly enhancing which transitions are kept or replayed, so that limited buffers retain the most informative continual signal \cite{zhang2023ReplayenhancedContinual}.

\subsubsection{Direct Replay}\label{sec: direct_replay}
A major focus is direct replay, explicitly storing and reusing buffered experiences.
\textit{Selective experience replay} \cite{isele2018selective} prioritizes long-term storage by importance (e.g., distribution matching or state-space coverage), ensuring diverse revisits to mitigate forgetting.
Similarly, \textit{Continual Learning with Experience And Replay} (CLEAR) \cite{rolnick2019experience} merges on-policy learning with off-policy replay, utilizing behavior cloning and V-Trace corrections to stabilize multi-task learning without task boundaries.

Recent methods integrate replay with other techniques.
For example, CoMPS \cite{berseth2022comps} buffers high-reward experiences for meta-policy search, accelerating adaptation.
\textit{Replay-based Recurrent RL} (3RL) \cite{caccia2023task} uses \textit{Recurrent Neural Networks} (RNNs) to encode history, facilitating task-agnostic adaptation.
Additionally, incorporating relabeling, weighting, and task inference improves sample efficiency in robotics and natural language processing \cite{xie2022lifelong,hafezbehavior2021,zhang2024cppo}.

Despite their success, explicit storage raises memory, scalability, and privacy concerns.
Future research could investigate dynamic memory allocation or selective forgetting to alleviate these bottlenecks.

\subsubsection{Generative Replay}\label{sec: generative_experience}
Instead of explicitly storing experiences, generative replay methods leverage generative models to recreate or simulate previous experiences, enabling the agent to revisit and learn from past knowledge without requiring a large memory footprint. 
This makes generative replay particularly suited for scenarios with limited memory resources or strict privacy constraints.
By synthesizing experiences on demand, these methods provide a flexible and efficient mechanism for continual learning.

Most generative replay methods rely on models like \textit{Variational Auto-Encoders} (VAEs) or \textit{Generative Adversarial Networks} (GANs).
For example, \textit{Reinforcement-Pseudo-Rehearsal} (RePR) \cite{craig2021pseudo} uses a GAN-based dual memory system to generate and rehearse representative states from previous tasks alongside new data, succeeding in Atari environments. 
Similarly, \textit{Self-generated Long-term Experience Replay} (SLER) \cite{li2021sler} pairs short-term replay with an \textit{Experience Replay Model} (ERM) that simulates past experiences based on minimal retained task information, maximizing memory efficiency in complex domains like StarCraft II.

Beyond state-level generation, trajectory-level synthesis generates task-relevant rollouts, preserving long-horizon credit assignment \cite{yue2025TDGRTrajectoryBased}.

Other works introduce unique triggers and management systems. 
\textit{Self-TRIGgered GEnerative Replay} (S-TRIGGER) \cite{caselles2021strigger} employs statistical tests to detect environmental shifts, prompting a VAE to generate prior samples adaptively. 
Alternatively, a model-free approach \cite{daniels2022modelfree} adopts a wake-sleep cycle, consolidating memory by alternating between task learning and VAE-based replay, highlighting hidden replay as a state-of-the-art technique.

While synthesizing experiences balances memory efficiency and knowledge retention, challenges remain in preserving sample fidelity and diversity. 
Generative models are prone to feature drift and inaccuracies that destabilize long-term learning \cite{li2021sler,daniels2022modelfree,caselles2021strigger}. 
Future research may explore robust models and hybrid systems that selectively store critical raw experiences.

\textbf{Scalability considerations.} The main bottleneck of experience-focused methods is the burden of maintaining a replay buffer or a generator that remains useful throughout long task streams. Their \textit{memory growth} is dominated by stored trajectories, latent summaries, or auxiliary generator states, while \textit{compute growth} rises with replay sampling, relabeling, and synthetic data generation. Their \textit{task-scaling} is usually strongest under recurring non-stationarity, where revisiting prior experience directly stabilizes learning, especially in task-aware or task-agnostic settings with repeated contexts. At the same time, these methods are tightly constrained by memory and privacy requirements, so their practical scalability depends on how aggressively past experience can be compressed, filtered, or regenerated.

\begin{figure}[htbp]
    \centerline{\includegraphics[width=0.8\linewidth]{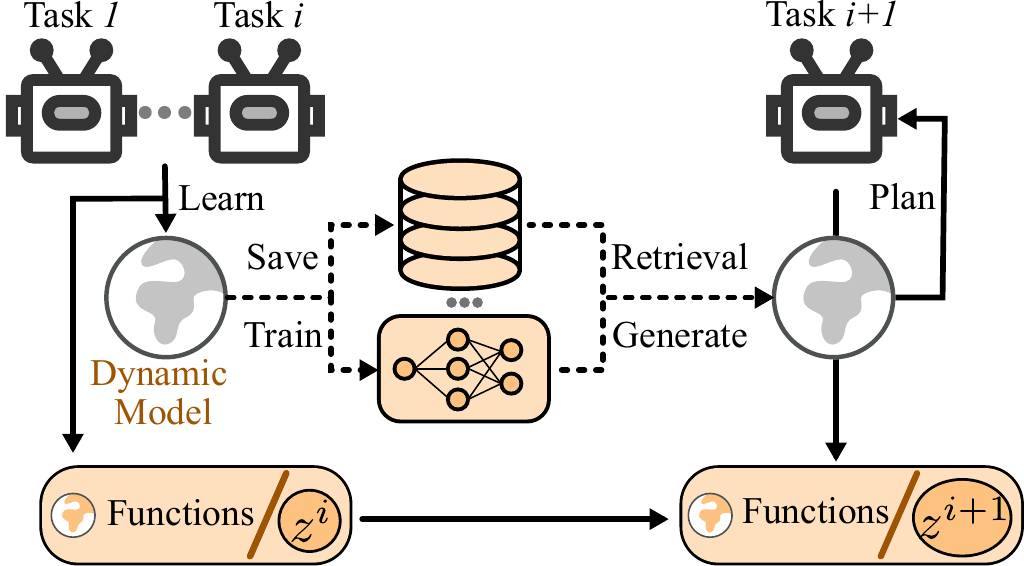}}
    \caption{
        The framework of dynamic-focused methods. 
        Direct modeling (above) and indirect modeling (below) are two main categories.
        }
    \label{fig: dynamic-focused}
    \vspace{-0.5cm}
\end{figure}

\subsection{Dynamic-Focused Methods}\label{sec: dynamic-focused}
Dynamic-focused methods in CRL are closely related to \textit{Model-Based Reinforcement Learning} (MBRL), where the core idea is to learn a model of the environment's dynamics to predict future states and rewards.
In CRL, dynamic-focused methods extend this concept to tackle non-stationary environments.
They enhance the agent's ability to adapt to changing environments and tasks by modeling the environment's dynamics ($T(s'|s, a)$).
As shown in Fig. \ref{fig: dynamic-focused}, dynamic-focused methods can be divided into two categories: direct modeling and indirect modeling.

\subsubsection{Direct Modeling}\label{sec: direct_modeling}
Direct modeling explicitly learns the dynamics of the environment (\eg the transition function) based on observed state-action pairs.
These methods aim to capture the underlying structure of the environment, allowing the agent to predict future states and adapt its behavior accordingly. 
By maintaining an explicit model of the environment, direct modeling approaches are well-suited for tasks requiring long-term planning and reasoning in changing conditions.

Direct modeling often uses mixture models and probabilistic frameworks to manage forgetting and task adaptation. 
Approaches like \textit{Meta-learning for Online Learning} (MOLe) \cite{nagabandi2018deep} and \textit{LifeLong Incremental Reinforcement Learning} (LLIRL) \cite{wang2022lifelong} employ the \textit{Chinese Restaurant Process} (CRP) to dynamically expand a library of dynamics models $\{d_{\eta_t}^{(l)}\}_{l=1}^L$. 
Assigning a new observation $(s_t, a_t, s_{t+1})$ to a model is dictated by a probabilistic prior, such as the CRP:
\begin{equation}
    P(x_t = l) = 
    \begin{cases} 
    \frac{n^{(l)}}{t - 1 + \zeta}, & \text{if } l \leq L, \\
    \frac{\zeta}{t - 1 + \zeta}, & \text{if } l = L + 1,
    \end{cases}
    \label{eq:crp}
\end{equation}
where $x_t$ assigns the observation, $n^{(l)}$ denotes counts for model $l$, and $\zeta$ scales new model creation. 
Leveraging probabilistic mixtures (e.g., infinite Gaussian processes \cite{xu2020task}) allows agents to efficiently reuse familiar models or create new ones for unseen shifts.

To boost scalability, \textit{Continual Reinforcement Learning via Hypernetworks} (HyperCRL) \cite{huang2021continual} avoids maintaining multiple models by using a fixed-capacity hypernetwork to generate task-specific dynamics models:
\begin{equation}
    \hat{s}_{t+1} = f_{\eta_k}(s_t, a_t), \eta_k = \mathcal{H}_{\Theta_k}(e_t),
    \label{eq:hypernetwork}
\end{equation}
where $\mathcal{H}_{\Theta_k}$ processes task embedding $e_k$ ($e_t$). Regularization ensures sustained predictive accuracy. 
Additionally, Losse-FTL \cite{liu2024locality} applies locality-sensitive sparse encoding and a \textit{Follow-The-Leader} (FTL) objective, enabling efficient incremental updates in high-dimensional environments. 

\subsubsection{Indirect Modeling}\label{sec: indirect_modeling}
Indirect modeling methods do not directly model the environment's dynamics but instead use alternative representations or abstractions (\eg latent variables) to infer or adapt to the dynamics.
They allow the agent to generalize across tasks without requiring a detailed model of the environment's transitions.

A prominent example is \textit{LIfelong Latent Actor-Critic} (LILAC) \cite{xie2020DeepReinforcement}, utilizing latent variables for non-stationary environments.
LILAC frames a \textit{Dynamic Parameter Markov Decision Process} (DP-MDP) where a latent variable ${z}$ tracks parameters shifting stochastically across episodes. 
It maximizes expected returns while keeping a compact dynamics representation via:
\begin{equation}
    \begin{aligned}
    P( o_{1:H}=1| &\tau^{1:i-1}) \geq \\ \mathbb{E}_{P({z}^{i}|\tau^{1:i-1})}&\left[\sum_{t=1}^{T}R(s_{t},{a}_{t};{z}^{i})-\log\pi({a}_{t}|{s}_{t},{z}^{i})\right],
    \end{aligned}
    \label{eq:lilac}
\end{equation}
where $\tau^{1:i-1}$ is the past trajectory, and ${z}^i$ the current task embedding. 
This optimizes both return and policy entropy for robust adaptation.
Similarly, task-agnostic methods like 3RL \cite{caccia2023task} and Continual-Dreamer \cite{kessler2023effectiveness} use abstract representations to handle non-stationarity. 
3RL relies on recurrent memory and replay, whereas Continual-Dreamer merges world models with reservoir sampling to consolidate knowledge.

Additionally, intrinsic rewards often guide exploration in these models.
\textit{Lifelong Skill Planning} (LiSP) \cite{lu2021resetfree} and Continual-Dreamer both formulate intrinsic rewards from prediction uncertainty, promoting exploration in unfamiliar regions. 
These models prioritize generalized adaptability over exact environmental modeling, often incorporating variational inference \cite{xie2020DeepReinforcement} or ensembles \cite{kessler2023effectiveness} for scalability. 
However, managing unobserved or continuous dynamic shifts remains challenging \cite{xie2020DeepReinforcement}, motivating future work toward relaxing episodic boundaries for real-time adaptation.

\textbf{Scalability considerations.} The main bottleneck of dynamic-focused methods is the overhead of maintaining world models, model libraries, or latent predictors together with any downstream planning machinery. Their \textit{memory growth} comes from storing model components, task-conditioned embeddings, or uncertainty estimates, and their \textit{compute growth} is amplified by model fitting, inference over mixtures, and planning through learned dynamics. Their \textit{task-scaling} is strongest when tasks share dynamics structure or when transitions evolve more systematically than rewards, because the same model family can be reused across multiple tasks. These methods are therefore especially well matched to non-stationarity learning and related settings where compact dynamics reuse can amortize modeling cost over long horizons.

\begin{figure}[htbp]
    \centerline{\includegraphics[width=0.8\linewidth]{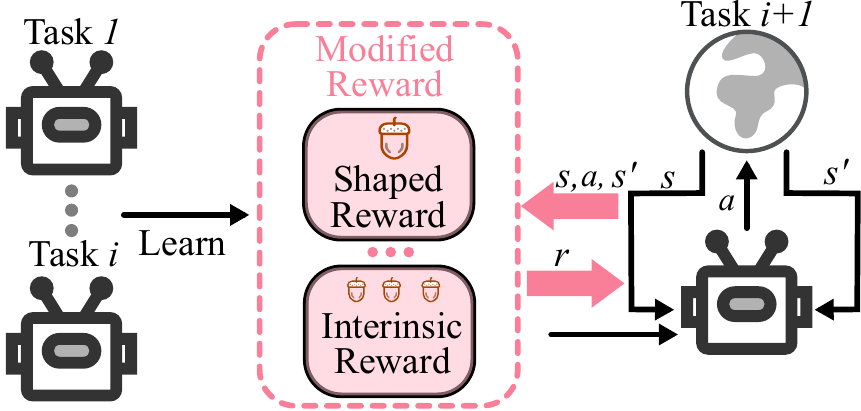}}
    \caption{
        The framework of reward-focused methods.
        }
    \label{fig: reward-focused}
    \vspace{-0.5cm}
\end{figure}

\subsection{Reward-Focused Methods}\label{sec: reward-focused}
Reward-focused methods in CRL concentrate on managing and leveraging reward signals to facilitate efficient learning and adaptation to new tasks, which is similar to reward shaping in transfer reinforcement learning.
These methods are particularly significant in CRL because rewards directly influence the agent's policy optimization and learning trajectory. 
By restructuring or reshaping reward distributions, reward-focused approaches address key challenges such as sparse or delayed rewards, knowledge transfer across tasks, and maintaining consistent learning performance over time.
In general, these methods modify the reward function $r_t$ by incorporating shaping functions or intrinsic components, which can be expressed as:
\begin{equation}
    R_t^{\text{M}} = R_t + h(s, a, s') + \alpha R_t^{\text{I}},
    \label{eq:reward-focused}
\end{equation}
where $h(s, a, s') $ is a shaping function derived from external knowledge or task-specific information, and $ R_t^{\text{I}} $ is an intrinsic reward component that encourages exploration or other desirable behaviors. 
The weighting factor $\alpha$ balances the contribution of intrinsic rewards to the overall reward signal.

Several approaches exemplify these reward-focused methods. 
\textit{Shaping Rewards for LifeLong RL} (SR-LLRL) \cite{kun2021accelerating} employs a \textit{Lifetime Reward Shaping} (LRS) function based on cumulative visit counts from prior tasks:
\begin{equation}
    h_i(s, a, s') = (1 - \gamma) V_{\text{max}} \frac{c_{i-1}(s, a, s')}{c_{i-1}(s)},
    \label{eq:lrs}
\end{equation}
where $c_{i-1}(s, a, s')$ and $c_{i-1}(s)$ count visits from past optimal trajectories. 
This mitigates sparse rewards and accelerates learning. 
Jiang \etal \cite{jiang2021temporal} apply temporal-logic-based shaping, deriving $h(s, a, s')$ from a potential function $\Phi$ and the optimal $Q^{\star}$:
\begin{equation}
    h(s, a, s') = \Phi\left(s', \arg\max_{a'} Q^{\star}(s', a')\right) - \Phi(s, a).
    \label{eq:temporal-logic}
\end{equation}
This robustly guides exploration using domain logic, even with imperfect advice.

Intrinsic rewards also critically encourage exploration and curiosity. 
Combined with extrinsic rewards, they are formalized as:
\begin{equation}
    R_t = R_t^{\text{E}} + \alpha R_t^{\text{I}},
    \label{eq:intrinsic-reward}
\end{equation}
where $R_t^{\text{E}}$ and $R_t^{\text{I}}$ denote extrinsic and intrinsic components weighted by $\alpha$. 
For example, \textit{Intrinsically Motivated Lifelong exploration} (IML) \cite{bougie2021intrinsically} merges short- and long-term intrinsic rewards:
\begin{equation}
    R_t^{\text{I}} = \max(R^{\text{D}}_t, 1.0) \cdot R^{\text{L}}_t,
    \label{eq:iml}
\end{equation}
where local bonus $R^{\text{L}}_t$ and deep bonus $R^{\text{D}}_t$ jointly stimulate exploration of novel states.
Similarly, \textit{Reactive Exploration} \cite{steinparz2022reactive} uses prediction errors to generate intrinsic signals:
\begin{equation}
    R_{t+1} = \alpha R_{t+1}^{\text{S}} + \beta R_{t+1}^{\text{R}} + \lambda R_{t+1}^{\text{E}},
    \label{eq:reactive-exploration}
\end{equation}
with $R_{t+1}^{\text{S}}$ and $R_{t+1}^{\text{R}}$ derived from observation and reward models, and $\lambda = 1 - \alpha - \beta$. 
This fosters dynamic adaptation by exploring significantly changed state regions. 
Additionally, \textit{Efficient Lifelong Imitation Reinforcement Learning} (ELIRL) \cite{mendez2018lifelong} extends this to inverse RL, rebuilding task-specific reward functions via shared latent components.

Overall, these methods overcome traditional RL reward limitations, mitigating sparse feedback and improving cross-task transfer. 
However, scaling to high-dimensional state-action spaces remains challenging \cite{kun2021accelerating,bougie2021intrinsically}. 
Furthermore, predefined structures in logic-based shaping \cite{jiang2021temporal} or reward machines \cite{zheng2022lifelong} may restrict fully autonomous deployment. 
Future work could develop automated reward-shaping mechanisms and integrate them seamlessly with policy- or experience-focused CRL paradigms. 

\textbf{Scalability considerations.} The main bottleneck of reward-focused methods is the overhead of maintaining reward models, shaping rules, or intrinsic-reward estimators that stay aligned with a changing task stream. Their \textit{memory growth} is often modest compared with replay-heavy methods, but it still increases with stored reward abstractions, logic specifications, or auxiliary novelty estimators, while \textit{compute growth} comes from evaluating shaping terms and updating intrinsic signals online. Their \textit{task-scaling} is strongest when reward semantics drift more than transition dynamics, because reused shaping structure can accelerate adaptation without rebuilding the full policy or world model. These methods therefore fit best in settings with recurrent reward redesign, sparse feedback, or semantic task variation, but they weaken when reward engineering itself becomes the dominant maintenance cost.

%% file: sections/future.tex
In this section, we present some open challenges and future directions in CRL, based on both retrospectives of the discussed methods and outlooks to the emerging trends in AI.
More future directions are discussed in Appendix~F, including evaluation and benchmark, interpretable knowledge, embodied agents, multi-agent CRL, offline learning, imitation learning and adjacent directions.

\vspace{-0.2cm}
\subsection{Task-Free CRL}\label{sec:task-free}
Most CRL methods assume that tasks are given in advance, the boundaries between tasks are clear, and the environment is stationary within a task.
However, the environment may continuously change over time. 
Moreover, in real-world scenarios, agents are expected to learn in a non-stationary environment.
Task-free CRL is a challenging problem that requires agents to learn from the environment without any explicit tasks. 
It is closely related to online learning and has been preliminarily explored in some works on task boundary detection and task-agnostic CRL.
Recent work has also studied statistical context detection to infer task changes online without task supervision \cite{dick2024StatisticalContext}.
We believe this direction is the necessary path for RL to move towards general AI.

\vspace{-0.2cm}
\subsection{Large-Scale Pre-Trained Model}\label{sec:ptms}
Recently, unprecedented breakthroughs have been achieved in learning large-scale \textit{Pre-Trained Models} (PTMs), such as \textit{Large Language Models} (LLMs), built on massive computation resources and diverse data.
This area presents several challenges relevant to the continual learning community.
We briefly point out two directions that are worth exploring:
\begin{enumerate}
    \item \textbf{CRL for large-scale PTMs}: One important method for aligning LLMs with human preferences is RLHF \cite{nisan2022learning}.
    CPPO \cite{zhang2024cppo} has been proposed to enhance RLHF with CL, allowing LLMs to adapt to human preferences continuously without extensive retraining.
    Recently, methods such as RL with verifiable rewards have been proposed to fine-tune LLMs to achieve better reasoning ability \cite{shao2024deepseek,guo2025deepseek,hu2025openreasonerzero}.
    Integrating CRL with these methods is expected to further improve the performance of LLMs in terms of reasoning and decision-making.
    \item \textbf{Large-scale PTMs for CRL}: PTMs can be used as a knowledge base for CRL, which can be transferred to the target task to improve sample efficiency and generalization ability.
    Existing works have shown that PTMs can be used to improve the performance of CL and RL in some specific scenarios \cite{zhang2023bootstrap,chen2024llm,bai2024digirl}.
    MT-Core first integrates an LLM into the CRL paradigm, enabling agents to perform multi-granularity knowledge transfer across diverse tasks \cite{pan2025multigranularity}.
    How to effectively leverage PTMs in CRL is also an open question, and we expect more research to be conducted in this direction.
\end{enumerate}

\subsection{Adjacent Directions}\label{sec:adjacent}
Several adjacent research areas offer complementary perspectives and techniques that can significantly advance CRL. These include \textbf{Unsupervised RL}, which fosters task-agnostic adaptation through reward-free exploration and self-supervision \cite{frans2024unsupervised,peac2024ying,self2025zhao}, \textbf{Meta-RL}, which targets fast lifelong adaptation and forward transfer via meta-learned structures \cite{meta2023bing,meta2024xu}, and studies on \textbf{Transfer and Negative Transfer}, which dissect the boundaries of knowledge reuse to balance plasticity and stability \cite{garces2025adaptive,ahn2025prevalence}. 
Additionally, \textbf{Representation-Based Continual Learning} emphasizes learning disentangled and discrete representations to moderate memory growth and mitigate interference \cite{botteghi2024unsupervised,meyer2023harnessing,chua2024successor}, while \textbf{Sim-to-Real and Continual Domain Randomization} provide vital levers for scalable domain adaptation in real-world embodied deployments \cite{josifovski2024continual}. Finally, the emergence of \textbf{In-Context RL} introduces prompt-driven, low-cost adaptation mechanisms that naturally synergize with large-scale pre-trained models \cite{moeini2025survey,wang2025provableicrl,retrieval2024schmied,unlock2024jiashun}.

%% file: sections/conclusion.tex
Continual reinforcement learning represents a pivotal paradigm for advancing autonomous decision-making, transitioning from isolated task optimization toward the realization of persistent agents capable of lifelong adaptation in non-stationary environments. 
In this survey, we have systematically navigated the evolving CRL landscape, comprehensive examinated its challenges, scenario settings, taxonomy, and open challenges.
Our analysis underscores that the core of CRL lies in navigating the triangular balance among plasticity, stability, and scalability. 
By introducing a novel taxonomy based on knowledge storage and transfer, we have established a structured framework for categorizing diverse methods and their respective strategies. 
Despite substantial progress, the field faces many challenges, particularly in terms of scalability and the fragmentation of benchmarks.
Ultimately, addressing these challenges will be critical for bridging the gap between theoretical research and the deployment of robust, general-purpose AI in dynamic, real-world applications.

%% file: sections/appendix.tex
\section*{Appendix A\\ The Popularity of CRL}\label{sec: popularity}
\begin{figure}[htbp]
    \centering
    \includegraphics[width=0.99\linewidth]{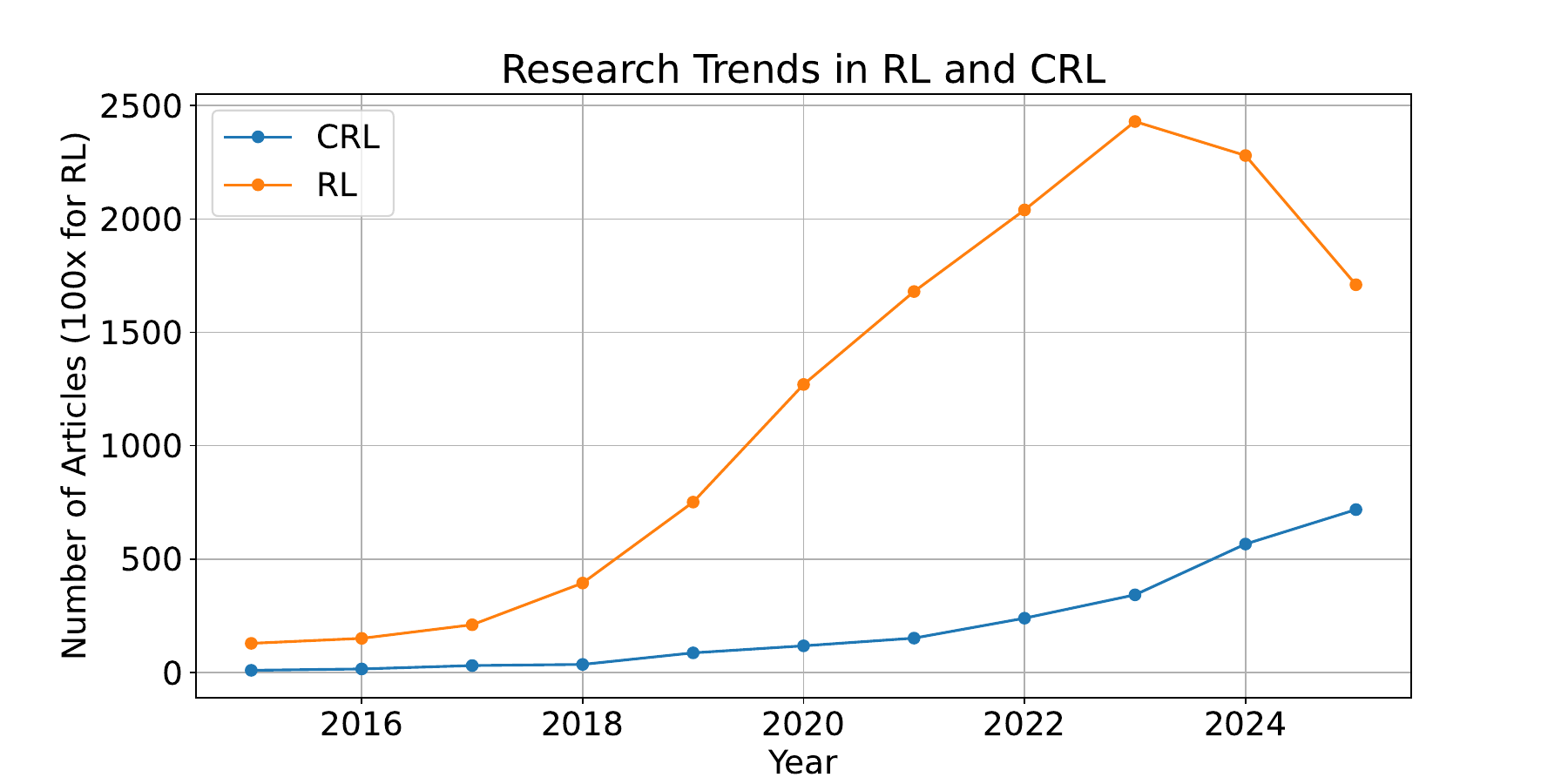}
    \caption{The popularity of CRL and RL from 2015 to 2025.}
    \label{fig: popularity}
  \end{figure}

Figure \ref{fig: popularity} shows the number of published articles in CRL and RL over the past ten years (from 2015-2025) according to Google Scholar.
For the CRL articles, we included those that have either the ``continual reinforcement learning'' or ``lifelong reinforcement learning'' keywords.
The results show that the number of articles in CRL has been increasing rapidly since 2018, indicating a growing interest in this field.

\begin{table}
  \centering
  \caption{Mathematical Notations.
  }
  \label{tab: notations}
  \begin{tabular}{m{1cm}m{6cm}}
  \hline
  Notation & Meaning \\
  \hline
  $M$ & MDP (also represents the task in CRL)\\
  $\mathcal{S}, \mathcal{A}, \mathcal{O}$ & State space, action space and observation space \\
  $T,R,\Omega$ & Transition function, reward function and observation function \\
  $\rho_0$ & Initial state distribution\\
  $t$ & Time step\\
  $H$ & Episode horizon for episodic finite-horizon MDPs \\
  $\gamma$ & Discount factor for continuing (infinite-horizon) MDPs \\
  $s, a, r$ & State, action and reward \\
  $\pi , \theta(\bm{\theta})$ & Policy function and its parameters \\
  $V,Q$ & State value function and state-action value function \\
  $G$ & Return \\
  \hline
  $\mathcal{M}$ & Task set \\
  $\mathcal{K}$ & The space of all tasks\\
  $N$ & Number of tasks\\
  $D$ & Data \\
  $g$ & Goal state \\
  $i,j,k$ & Task indices\\
  $p_{i,j}$ & Normalized performance of task $j$ after training on task $i$\\
  $A_i$ & Average performance after training sequentially through tasks $1$ to $i$\\
  $A_N$ & Final average performance after training through all $N$ tasks\\
  $FG_i$ & Forgetting on task $i$ (performance drop from learning time to final)\\
  $FG$ & Average forgetting over tasks\\
  $p_i(t)$ & Normalized performance on task $i$ at training step $t$\\
  $\Delta$ & Per-task training budget (in steps) used for AUC normalization\\
  $\mathrm{AUC}_i$ & Normalized area under the learning curve for task $i$ over its training interval\\
  $\mathrm{AUC}_i^{b}$ & AUC of a single-task baseline reference run for task $i$\\
  $FT_i$ & Per-task forward transfer based on AUC improvement over baseline\\
  $FT$ & Average forward transfer over tasks\\
  $BWT$ & Backward transfer (final performance change on past tasks)\\
  \hline
  $\bm{L}, \bm{s}$ & Policy latent basis and the task-specific coefficients \\
  $\mathcal{G}, $ & Task group\\
  $\bm{D}$ & Task descriptors latent basis\\
  $\mathcal{D}$ & Task domain \\
  $\bm{q}$ & Environmental coefficients\\
  $\delta$ & TD error\\
  $\alpha, \beta, \lambda$ & Learning rates or weighting factors\\
  $m, \bm{M}$ & module and module set\\
  $f$ & solution function \\
  $u$ & activation score \\
  $\eta$ & parameters of dynamic models\\
  $\bm{F}$ & Fisher information matrix\\
  $l, L$ & Dynamic model and capacity of dynamic models libraries\\
  $\zeta$ & concentration parameter\\
  $x$ & assignment \\
  $H, \Theta$ & Hyperparameter and its parameters\\
  $e$ & Task embedding\\
  $\tau$ & trajectory\\
  $z$ & Task latent variable\\
  $h$ & Shaping function\\
  $c$ & Cumulative visit counts\\
  $\Phi$ & Potential function\\
  \hline
  \end{tabular}
\end{table}

\section*{Appendix B\\ Mathematical Notations}\label{sec: notations}
Table \ref{tab: notations} summarizes the mathematical notations used in this survey.

\section*{Appendix C \\ Continual Reinforcement Learning Tasks}\label{sec: tasks}
\subsection{Tasks}\label{sec:tasks}
In the domain of CRL, most agents are tasked with objectives at each step of a task sequence that aligns with the goals of RL tasks. 
This positions tasks as the foundational units of CRL. 
This section aims to introduce existing tasks within CRL and provide a succinct analysis of them.

\textbf{Navigation tasks} are one of the most commonly employed scenarios in CRL, often utilizing two-dimensional state spaces and a discrete set of actions. 
In these tasks, agents must explore unknown environments via continuous movement to reach a designated goal. 
Researchers frequently design task sequences based on grid-world environments \cite{sutton2018reinforcement}, where rewards or environmental dynamics vary to assess CRL algorithms \cite{abel2018policy}. 
These tasks are relatively simple to learn and provide a lower difficulty for computationally intensive CRL algorithms. 
Furthermore, navigation tasks lend themselves well to environment procedural generation, which is essential for CRL. 
MiniGrid \cite{MinigridMiniworld23}\footnote{\url{https://minigrid.farama.org}} is the most widely used environment library, offering a variety of map sizes and layouts for task generation. 
It provides preset environments like Doorkeyenv, Fourroomsenv, and Memoryenv for constructing diverse CRL task sequences. 
Additionally, JBW offers a testbed for lifelong learning by generating non-stationary environments within a 2D grid world. 
For more realistic evaluations, 3D navigation tasks, such as those based on DeepMind Lab \cite{beattie2016deepmind}, have been used to further assess CRL algorithms \cite{schwarz2018progress}.

\textbf{Control tasks} are another prevalent CRL task type, typically involving three-dimensional state spaces and a discrete action set. 
Classic examples include the mountain car, inverted pendulum, and double pendulum tasks \footnote{\url{https://www.gymlibrary.dev/environments/classic_control}}, where agents must reach specific target states (e.g., the peak of the mountain, an upright position, or a target height) using simple control commands (e.g., forward, backward, left turn, right turn) \cite{towersgymnasium2023}. 
Task sequences in control tasks are often formed by altering the objectives \cite{kessler2022same} or by switching between different tasks \cite{kaplanis2018continual}, enabling the evaluation of CRL algorithms. 
In more complex tasks, agents are required to control robotic devices, such as robotic arms and legs \cite{todorov2012mujoco}. 
These tasks involve physical properties, presenting significant challenges while also offering practical applications.
Researchers typically modify parameters such as limb length, mass, environmental friction, and gravity within control tasks to create diverse task sequences for CRL evaluation \cite{fu2022modelbased, gaya2022building, kaplanis2019policy}.

\textbf{Video games} present challenging reinforcement learning tasks, where the state space typically consists of images, and the actions are discrete. 
Within these environments, agents must perform complex controls to achieve specific goals, making video games an ideal testbed for evaluating the scalability of CRL algorithms in challenging scenarios \cite{espeholt2018impala, anand2023prediction}. 
The Atari 2600 \footnote{\url{https://www.gymlibrary.dev/environments/atari}}, with its collection of games that share consistent state and action spaces, is one of the most frequently used sets in DRL experiments \cite{bellemare2013arcade, mnih2015human}. 
Researchers evaluate CRL algorithms by combining different games into task sequences \cite{rusu2016progressive, james2017overcoming, schwarz2018progress}. 
For more complex tasks involving long-horizon strategy games with rich observations, some works have explored environments like MineCraft and StarCraft II \cite{oriol2017startcraft, zheng2022lifelong, lu2021resetfree, daniels2022modelfree}. 
However, these tasks are computationally expensive due to their large state spaces and complex task structures, requiring extended training durations \cite{abbas2023loss}.

Although most tasks are in simulated environments, some studies have also applied CRL algorithms to real-world robotic tasks. 
These include 2D navigation tasks for mobile robots \cite{liu2019lifelong, traore2019discorl}, robotic arm control tasks \cite{liu2023continual}, and home robot tasks \cite{powers2023evaluating}. 
They present additional challenges, such as sensor noise, mechanical constraints, and the need for robust online learning, making them a practical direction for the further development of CRL techniques.

\section*{Appendix D \\ Applications}\label{sec: applications}
Although the research on CRL is shown to be promising, the application of CRL in real-world scenarios is still in its infancy.
In this section, we summarize recent applications that are closely related to CRL, including robotics, autonomous driving, and game playing.

\subsection{Robotics Learning}\label{sec: robotics} 
Robotics is a prominent application domain of CRL, where lifelong robots are required to adapt to new tasks or environments without forgetting previously learned tasks.
For example, robots must continuously learn new skills as they encounter various tasks, appliances, and user preferences in home environments \cite{powers2023evaluating}.
Traditional robot learning methods typically rely on a large amount of independent and identically distributed data, which is impractical in dynamic settings. 
Recent studies have focused on addressing this challenge by using offline reinforcement learning, skill learning, and distillation that enable robots to learn efficiently from limited demonstrations and adapt to new tasks while mitigating catastrophic forgetting \cite{zhou2022forgetting,xie2022lifelong,powers2023evaluating,zhao2024ExperienceConsistency}.
Fine-tuning pre-trained policies is also an effective strategy for continual learning in robotics.
By adapting policies to new variations with minimal offline data, robots can improve their performance in dynamic environments \cite{julian2021NeverStop}.

The navigation task of mobile robots has been widely studied in CRL due to its extensive application and relative simplicity.
CRL has been applied to enable mobile robots to learn sequentially in unknown environments through techniques such as policy distillation, task decomposition, and behavior self-organization \cite{traore2019ContinualReinforcement,wang2020LearningNavigate,hafezbehavior2021}.
Additionally, \textit{Lifelong Federated Reinforcement Learning} (LFRL) leverages cloud-based systems to enhance navigation capabilities by fusing experiences from multiple robots, thereby improving generalization across different environments\cite{liu2019lifelong}.

Recent developments have also introduced benchmarks specifically designed to evaluate the performance of CRL methods in robotic tasks.
Continual World provides a structured sequence of robotic manipulation tasks that emphasize forward transfer, challenging existing algorithms to balance forgetting and transfer \cite{maciej2021advances}.
Furthermore, CORA provides a more comprehensive benchmark, in which sequences based on household robot tasks test agents in a more realistic visual domain and evaluate their sample efficiency \cite{sam2022cora}.
Despite these advancements, some researchers have pointed out that they are too challenging for current CRL agents, and have proposed simpler benchmarks to facilitate the development of more effective methods \cite{yang2022evaluations,xie2022lifelong}. 

\subsection{Game Playing}
The game is a common testbed for RL algorithms. 
It has evolved from classical benchmarks such as GridWorld games to more complex settings such as video games with multimodal inputs \cite{Justesen2020DeepLearning}.
These environments provide a controlled yet challenging setting for assessing the continual learning capabilities of CRL agents.
Simple video games, such as Procgen \cite{cobbe2020leveraging} and Atari \cite{bellemare2013arcade}, have been widely utilized as benchmarks for evaluating CRL methods \cite{schwarz2018progress,craig2021pseudo,abbas2023loss,muppidi2024fast}.
These games offer a diverse range of tasks (different environments and play modes) that test an agent's ability to generalize and retain knowledge, making them ideal for studying the effects of catastrophic forgetting and the efficacy of knowledge transfer mechanisms.
Based on these games, CORA introduces controlled variations and benchmarks that further challenge CRL agents, emphasizing the need for robust generalization and sample-efficient learning \cite{sam2022cora}. 
They provide structured sequences of tasks that highlight the strengths and limitations of current CRL methods, revealing that while some algorithms excel in preserving knowledge, they often struggle with adapting to new, visually complex tasks.
Furthermore, HackAtari introduces controlled novelty into traditional game environments \cite{delfosse2024hackatari}.
By modifying game dynamics and reward structures, it facilitates the evaluation of agents' ability to generalize and adapt to new conditions.

As CRL research progresses, more complex games such as online strategy games or 3D video games have become prominent testbeds for advanced CRL methods. 
These games present a higher level of complexity due to their high-dimensional state spaces and more unstable dynamic environments. 
In Minecraft, hierarchical approaches have been proposed to efficiently transfer and reuse skills across tasks, addressing the sample efficiency challenge posed by the game's vast and varied environment \cite{tessler2017DeepHierarchical,lu2021resetfree}. 
In StarCraft 2, model-free generative replay frameworks and wake-sleep mechanisms have been developed to improve continual learning efficiency \cite{daniels2022modelfree,sur2023SystemDesign}. 
Additionally, COOM provides an image-based CRL benchmark based on ViZDoom for evaluating agents with embodied perception \cite{tristan2023coom}. 
These frameworks leverage advanced modeling techniques to maintain performance across complex tasks, demonstrating significant improvements in both forward transfer and retention of previously acquired skills. 

\subsection{Others}
Recently, CRL has been explored in various other fields beyond the above, showcasing its versatility and potential for broad application. 
In \textbf{natural language processing}, CRL has been applied to dialogue systems by integrating a transformer, enabling them to integrate new knowledge dynamically to adapt to new topics and tasks without forgetting \cite{geishauser2022DynamicDialogue}.
Similarly, in controlled text generation, CRL frameworks have been utilized to allow large language models to adaptively generate text that aligns with specified attributes, such as topic or sentiment, in real-time \cite{shulev2024continual}.
Additionally, \textit{Continual Proximal Policy Optimization} (CPPO) has been proposed to enhance \textit{Reinforcement Learning from Human Feedback} (RLHF) \cite{nisan2022learning} by balancing policy learning and knowledge retention, allowing language models to advise human preferences without extensive retraining \cite{zhang2024cppo}.

In the field of \textbf{medical imaging}, CRL addresses the challenge of catastrophic forgetting by employing selective experience replay with corset compression \cite{zheng2024SelectiveExperience}.
In \textbf{finance}, CRL has been utilized for continual portfolio selection, allowing trading agents to adapt to dynamic market conditions by incrementally updating their strategies based on new data, thereby improving returns and reducing risks \cite{liu2023ContinualPortfolio}. 
In the domain of \textbf{autonomous driving}, CRL has been employed to improve the adaptability of self-driving cars in partially observable environments \cite{md2024NovelApproach}.
Additionally, CRL has been applied in the field of resource allocation within \textbf{industrial internet of things networks} \cite{wang2023MultigranularityFusion}. 
Here, the CRL method enables efficient resource management by continuously learning and adapting to the dynamic network conditions, thereby optimizing data transmission and energy consumption.
Furthermore, CRL has been applied to \textbf{data center cooling control}, where it enhances energy efficiency by enabling systems to adapt quickly and safely to changing thermal conditions \cite{wang2023PhyllisPhysicsinformed}.

\section*{Appendix E \\ Beyond Traditional CRL}\label{sec: beyond}
There are several emerging research directions in CRL that extend beyond the traditional methods discussed in Section \ref{sec:review}.
These methods encompass novel techniques and pose unique challenges in CRL, including out-of-distribution detection, imitation learning, and multi-agent coordination.
This section provides an overview of them and their potential impact on the field of CRL.

\subsection{Task Detection}\label{sec: task detection}
As described in Section \ref{sec: scenarios}, the task-agnostic scenario is a common setting in CRL, where the agent is not informed of the task identities or boundaries.
However, many CRL methods rely on task-specific information to guide learning and adaptation \cite{jacobson2022task,dick2024StatisticalContext}.
Task detection methods bridge this gap by enabling agents to identify task identities or detect environmental changes without explicit supervision. 
These methods are particularly useful for approaches that associate specific structures or policies with individual tasks.
Naturally, they can be divided into two categories: task identity detection and environment change detection.

Task identity detection methods aim to identify task labels by leveraging patterns in observations, latent representations, or task-specific features.
Therefore, unsupervised or semi-supervised learning techniques can be naturally applied to this problem.
For example, Jacobson \etal \cite{jacobson2022task} proposed the use of \textit{Familiarity AutoEncoders} (FAEs) for discovering task labels.
FAEs reconstruct input data for specific tasks, assigning task labels based on the autoencoder with the highest reconstruction performance.
This method avoids catastrophic forgetting of the task detector by maintaining separate models for each task.
Additionally, the exploration of variational and adversarial autoencoders as FAE variants highlights the adaptability of the approach to noisy environments.
Instead of using explicit task labels, \textit{Behavior-Guided Policy Optimization} (BGPO) \cite{hafezbehavior2021} uses behavior embeddings as task identities.
It employs a self-organizing network to incrementally learn a behavior embedding space from demonstrations. 
By matching new behaviors to the nearest embedding, the system efficiently infers tasks without requiring predefined task structures. 
This approach is further enhanced by a behavior-matching intrinsic reward, which aligns generated trajectories with demonstrated behaviors. 
The dynamic expansion of the embedding space ensures scalability to novel tasks, making this method suited for continual robot learning.
Additionally, OWL formulates task identity detection as a multi-armed bandit problem, allowing for adaptive policy selection during test time.
By combining a multi-head network with shared representations, this method effectively mitigates interference between tasks. 

In contrast, environment change detection methods focus on detecting environment dynamics rather than explicit task identities.
Environment changes in RL can be categorized as changes in the input distribution, changes in the transition function, or changes in the reward function.
Methods to detect changes in input distributions have been developed in the field of novelty and out-of-distribution detection with applications in CL \cite{aljundi22continual,liu2025wasserstein}.
A commonality is the use of statistical and probabilistic techniques to detect changes in the environment.
For instance, S-TRIGGER \cite{caselles2021strigger} employs a self-triggered generative replay mechanism, utilizing statistical analysis of reconstruction errors from VAEs to detect significant environmental changes.
Similarly, LLIRL \cite{wang2022lifelong} leverages an infinite mixture model with online Bayesian inference to adapt to dynamic environments. 
By using the Chinese restaurant process for environment clustering, LLIRL can detect changes without external signals, showcasing the importance of probabilistic frameworks in managing environment dynamics.

Recently, another statistical method, \textit{Sliced Wasserstein Online Kolmogorov-Smirnov} (SWOKS) \cite{dick2024StatisticalContext}, combines optimal transport methods with the Sliced Wasserstein distance and the Kolmogorov-Smirnov test to measure distances between experience distributions.
This method's ability to operate online without predefined task labels highlights its suitability for real-time adaptation in complex environments.
The use of distance metrics to detect task changes aligns with the statistical approaches of S-TRIGGER and LLIRL, yet SWOKS uniquely incorporates optimal transport methods to enhance detection accuracy.
In contrast to these statistical approaches, \textit{Reactive Exploration} \cite{steinparz2022reactive} focuses on modifying reward structures to include intrinsic rewards based on prediction errors. 
This method utilizes the \textit{Intrinsic Curiosity Module} (ICM) to detect changes and encourage exploration in altered regions of the state space.
This reward-focused strategy provides an alternative perspective, emphasizing the role of intrinsic motivation in detecting environmental changes.

\subsection{Offline Reinforcement Learning and Imitation Learning in CRL}\label{sec: offline and imitation}
\textit{Offline Reinforcement Learning} (Offline RL) and \textit{Imitation Learning} (IL) represent compelling extensions to CRL frameworks, particularly for leveraging static datasets or expert demonstrations to address the challenges of lifelong learning. 
Offline RL focuses on learning policies from pre-collected datasets without requiring direct interaction with the environment, which is advantageous in scenarios where real-world interactions are costly or infeasible. 
Currently, there is still very little research on the integration of offline RL with CRL, but some initial studies have shown promising results.
LiSP is an early step in this direction, which uses offline data to discover skills in reset-free lifelong RL, enabling long-horizon planning in an abstract skill space \cite{lu2021resetfree}.
Its effectiveness is demonstrated in a variety of settings, including offline interactions.

A recent work, \textit{Offline Experience Replay} (OER), formulates the \textit{Continual Offline Reinforcement Learning} (CORL), where an agent learns a sequence of offline reinforcement learning tasks and pursues good performance on all learned tasks with a small replay buffer without exploring any of the environments of all the sequential tasks \cite{gai2023oer}.
OER addresses the distribution shift problem in CORL by introducing a \textit{Model-Based Experience Selection} (MBES) scheme. 
This approach filters offline data to build a replay buffer that closely aligns with the learned model, mitigating catastrophic forgetting while maintaining performance on new tasks. 
Additionally, offline RL has been applied in some CRL methods \cite{mendez2022modular,zhou2022forgetting}, highlighting the potential of this technique to enhance CRL by integrating prior knowledge and reducing the reliance on extensive real-world interactions.

Imitation learning, on the other hand, provides a complementary approach to RL by utilizing expert demonstrations to guide the agent's learning. 
ELIRL \cite{mendez2018lifelong} and \textit{Fast Lifelong Adaptive Inverse Reinforcement learning} (FLAIR) \cite{chen2023fast} exemplify how IL can be adapted to CRL settings. 
ELIRL introduces a shared latent reward structure that facilitates knowledge transfer across sequential tasks while addressing catastrophic forgetting. 
FLAIR builds upon this by incorporating policy mixtures for fast adaptation to heterogeneous demonstrations, ensuring scalability and personalization in lifelong learning scenarios. 
Both approaches emphasize the importance of efficient knowledge-sharing and task-specific adaptation.
In addition, behavioral cloning, as a simple imitation learning method, has been applied in many CRL methods with experience replay \cite{rolnick2019experience,kobayashi2020ReinforcementLearning,wolczyk2022disentangling}. 

\subsection{Multi-Agent CRL}
Many real deployments require multiple agents to coordinate while both their partners and environments drift over time, which raises additional non-stationarity and credit-assignment challenges beyond the single-agent CRL setting. 
Recent work studies continual multi-agent coordination by modeling general relation patterns \cite{yao2025GeneralRelation} and by formalizing multi-agent continual learning setups and benchmarks \cite{yuan2025MultiagentContinual}.
However, the application of CRL to multi-agent systems is still in its early stages, and there are many open questions regarding how to effectively manage the complexities of multi-agent interactions while addressing the challenges of lifelong learning.
Furthermore, the benchmarks for multi-agent CRL are still limited.
Although a preprinted work have proposed multi-agent continual learning benchmarks based on Overcook \cite{tomilin2025MEALBenchmark}, more diverse and realistic benchmarks are needed to further advance research in this area.

\section*{Appendix F \\ More Future Works}\label{sec: more future works}

\subsection{Evaluation and Benchmark}\label{sec:eval}
Variant evaluation metrics have been proposed to measure CRL from different but complementary perspectives, although no single metric can summarize the efficacy of a CRL approach. 
In addition, most metrics are drawn from the CSL field, which may not be suitable for CRL.
Designing a set of generalized, novel metrics is beneficial for the development of CRL.
Moreover, with the effervescent development of large-scale models, it is crucial to standardize evaluation from the perspectives of scalability and privacy. 
The appropriateness of the stored/transferred knowledge and the security of the model should also be quantified as metrics.

\subsection{Interpretable Knowledge}\label{sec:interpretable}
Ranging from exterior experiences to policy function approximators, black-box knowledge is more accessible and predominant than interpretable and well-articulated knowledge.
However, the latter is beneficial for evaluating and explaining the process of CRL.
Moreover, building an interpretable knowledge base can help to transfer knowledge across various tasks and even alleviate catastrophic decision-making for high-stakes tasks such as autonomous driving.

\subsection{Embodied Agents}\label{sec:embodied}
Embodied agents are agents that interact with the environment through sensors and actuators, such as virtual agents and robots \cite{fan2022minedojo,mendez2023EmbodiedLifelong}.
The development of embodied agents with CL capabilities has gained increasing attention, particularly with the advent of LLMs \cite{tziafas2024lifelong}. 
Recent works have demonstrated how LLMs can enhance embodied agents by enabling efficient skill acquisition, interpretable knowledge storage, and adaptive reuse of learned behaviors \cite{wang2024voyager, meng2025preserving}. 
Such capabilities align closely with the goals of CRL, particularly in addressing catastrophic forgetting and improving generalization across tasks.
While some continual embodied agents often rely on imitation learning to acquire new behaviors and adapt to novel \cite{kim2024online,wang2024voyager,zeng2025continual}, recent advancements have begun to explore the integration of CRL into this domain \cite{meng2025preserving}. 
This shift is motivated by the need for agents to autonomously learn from dynamic, interactive environments and operate without predefined task boundaries. 
Despite the application of CRL to embodied agents being relatively nascent, it offers a compelling direction for CRL research, serving as a platform to study lifelong learning in complex, real-world scenarios.

\subsection{Adjacent Directions}\label{sec:appendix_adjacent}
\textbf{Unsupervised RL.} Learning without task labels or reward engineering keeps agents plastic when curricula shift unexpectedly. Works in unsupervised RL from pixels and functional reward encodings \cite{rajeswar2023mastering,frans2024unsupervised,peac2024ying,self2025zhao} show that agents can bootstrap diverse behaviors and handle out-of-distribution interactions, which reinforces CRL's task-agnostic and non-stationarity scenarios and reduces dependence on curated rewards. Related directions that blend reward-free adaptation and self-supervision further underline the relevance to continual deployment settings \cite{unsupervised2023ada,scrl2023fang}.

\textbf{Meta-RL.} Meta-reinforcement learning directly targets the lifelong adaptation and forward-transfer goals by training meta-parameters that admit fast adaptation to new tasks. The tutorial by Beck \etal \cite{beck2023tutorial} surveys recipes that can be mapped onto CRL taxonomies to shrink sample complexity while keeping stability high through shared meta-structures. Concrete lines of work include meta-learned exploration / policy-reuse viewpoints \cite{garcia2019advances}, as well as model-based and online adaptation variants that cope with shifting dynamics \cite{nagabandi2018deep,clavera2019learning}. Recent surveys and methods further systematize meta-RL under non-stationarity and distribution shift \cite{meta2023bing,meta2024xu}.

\textbf{Transfer and Negative Transfer.} Transfer-focused research reminds CRL that reuse can hurt, and recent studies \cite{garces2025adaptive,ahn2025prevalence} dissect when successor features compositions succeed and when negative transfer surfaces as a stability threat. Embedding such diagnostics into CRL benchmarks sharpens the plasticity-stability trade-off, especially for task-incremental and non-stationary settings where prior knowledge must be composed carefully.

\textbf{Representation-Based Continual Learning.} Rich, disentangled representations can isolate transferable structure from spurious details, improving scalability across long task lists. Surveys on unsupervised representation learning in RL \cite{botteghi2024unsupervised} highlight how latent dynamics models and contrastive features sustain memory-efficient knowledge retention, aligning with dynamic- and experience-focused taxonomies that aim to moderate memory and compute growth. Discrete representations have been shown to mitigate interference in CRL \cite{meyer2023harnessing}, while successor features offer compositional transfer across changing reward functions \cite{chua2024successor,sherstan2018accelerating}.

\textbf{Sim-to-Real and Continual Domain Randomization.} Bridging simulation and real-world deployments intensifies non-stationarity along transition and observation spaces, so continual domain randomization becomes an adjacent lever to evaluate generalization beyond synthetic benchmarks \cite{josifovski2024continual}. This direction stresses scalability by exposing agents to parameter shifts, echoing non-stationarity learning scenarios while demanding plastic yet stable domain adaptation.

For robotics and more general embodied agents, sim-to-real is often the practical bottleneck: policies must cope with sensor noise, actuator limits, contacts, and morphology changes that are rarely captured by a single simulator setting. These embodied deployment shifts make continual adaptation central rather than optional, and they connect naturally to the embodied-agent perspective summarized in Section~\ref{sec:embodied}.

\textbf{In-Context RL.} In-context learning empowers PTMs to solve new goals by conditioning on prompts rather than gradients, complementing CRL's need for low-cost adaptation. Surveys and theoretical work on in-context RL \cite{moeini2025survey,raparthy2023generalization,wang2025provableicrl} suggest that prompt-driven behavior could act as an implicit memory buffer, linking back to large-scale PTM aspirations and offering plug-and-play policies for lifelong agents, with retrieval-augmented external memory and related unlocking approaches as representative developments \cite{retrieval2024schmied,unlock2024jiashun}.